%% file: preprint.tex
\documentclass[10pt]{article}
\usepackage{cua-preprint}
\usepackage{amsmath,amssymb,booktabs,graphicx,array,longtable}
\usepackage{xcolor}
\usepackage[normalem]{ulem}
\usepackage{needspace}
\usepackage{float}
\usepackage{placeins}
\usepackage{wrapfig}
\usepackage{hyperref}
\usepackage{url}

\definecolor{coauthorYellow}{RGB}{179,143,0}

\hypersetup{colorlinks=true,linkcolor=CUANavy,citecolor=CUANavy,urlcolor=CUAAccent,pdftitle={CUA-SWE: When Computer-Use Agents Meet Visual Software Engineering},pdfauthor={Prince Zizhuang Wang, Chenhao Liang, Zelong Xu, Aojie Yuan, Xiaolin Zhou, Haiyue Zhang, Yue Zhao, Xiyang Hu, Shuli Jiang}}
\begin{document}
\thispagestyle{cuacover}
{\LARGE\scshape\raggedright
\mbox{CUA-SWE\,\raisebox{-0.25em}{\includegraphics[height=1.5em,trim=78bp 19bp 46bp 15bp,clip]{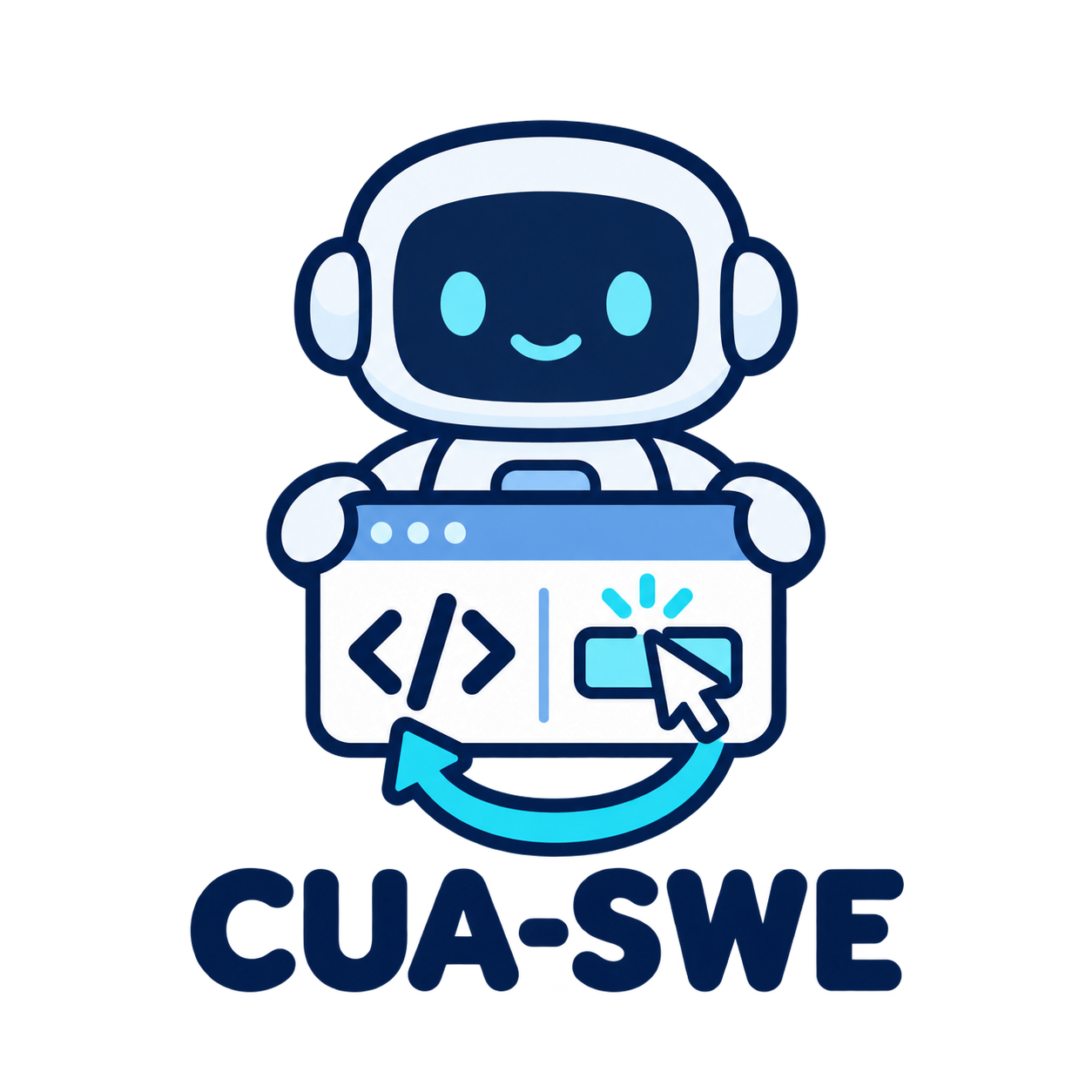}}}: When Computer-Use Agents Meet Visual Software Engineering\par}
\vspace{10pt}
\begin{center}
{\normalsize
Prince Zizhuang Wang\textsuperscript{1,*,$\dagger$},
Chenhao Liang\textsuperscript{2,*},
Zelong Xu\textsuperscript{3,*},
Aojie Yuan\textsuperscript{2,*},
Xiaolin Zhou\textsuperscript{4,*}\\[3pt]
Haiyue Zhang\textsuperscript{2},
Yue Zhao\textsuperscript{2},
Xiyang Hu\textsuperscript{4},
Shuli Jiang\textsuperscript{5}\par}
\vspace{4pt}
{\footnotesize Corresponding author: \href{mailto:princewang@cmu.edu}{\texttt{princewang@cmu.edu}}\par}
{\footnotesize *Equal contribution.\quad $\dagger$Project Lead.\par}
\vspace{4pt}
{\small
\textsuperscript{1}Carnegie Mellon University\quad
\textsuperscript{2}University of Southern California\\
\textsuperscript{3}University of Wisconsin--Madison\quad
\textsuperscript{4}Arizona State University\quad
\textsuperscript{5}AWS Agentic AI\par}
\end{center}
\begin{abstract}
Software development requires more than editing code: developers repeatedly run software, interact with its interfaces, visually inspect its behavior, and use these observations to decide what to change next and whether a change works. Existing coding agents and computer-use agents are largely studied in isolation, leaving this integrated development process underexplored.
Diagnosing a runtime interaction failure requires agents to connect visual observations with the responsible code, then use the application again to verify the repair.
We introduce CUA-SWE, a benchmark, environment, and evaluation pipeline for software engineering with computer use. Beyond studying how GUI feedback supports diagnosis and repair, we ask whether agents can complete software engineering tasks when required specification or operational information is available only through the running application's visual interface.
CUA-SWE spans four software engineering domains and requires agents to modify code and configuration, execute commands, interact with running software, and inspect visual feedback within the same task. Each task includes deterministic, task-specific tests that verify whether the resulting software satisfies the requirements and preserves specified behavior. Our evaluation characterizes how frontier agents combine source-level execution with application screenshots and graphical interaction to produce verified software changes. We examine performance across domains and task information requirements, alongside the development behaviors associated with successful repairs. CUA-SWE provides a unified testbed for studying how agents use visual feedback and interaction to guide software engineering, with executable correctness criteria for the resulting software.

\par\vspace{5pt}
\begin{center}
\href{https://github.com/kingofspace0wzz/cua-swe}{\raisebox{-2.5pt}{\includegraphics[height=15pt]{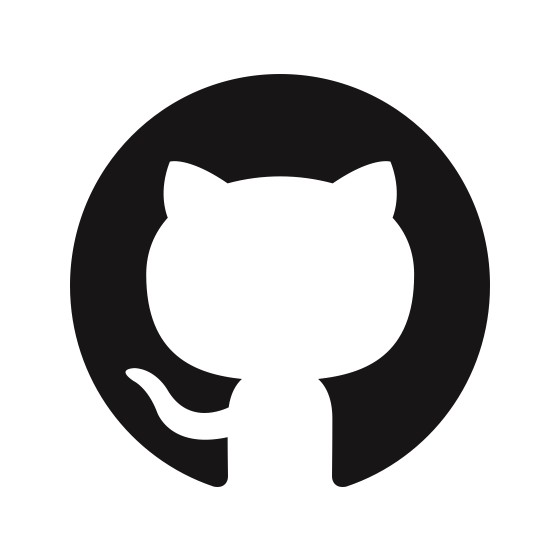}}\,GitHub}\qquad
\href{https://zelongxueric.github.io/cua-swe-website/}{\raisebox{-2pt}{\includegraphics[height=11pt]{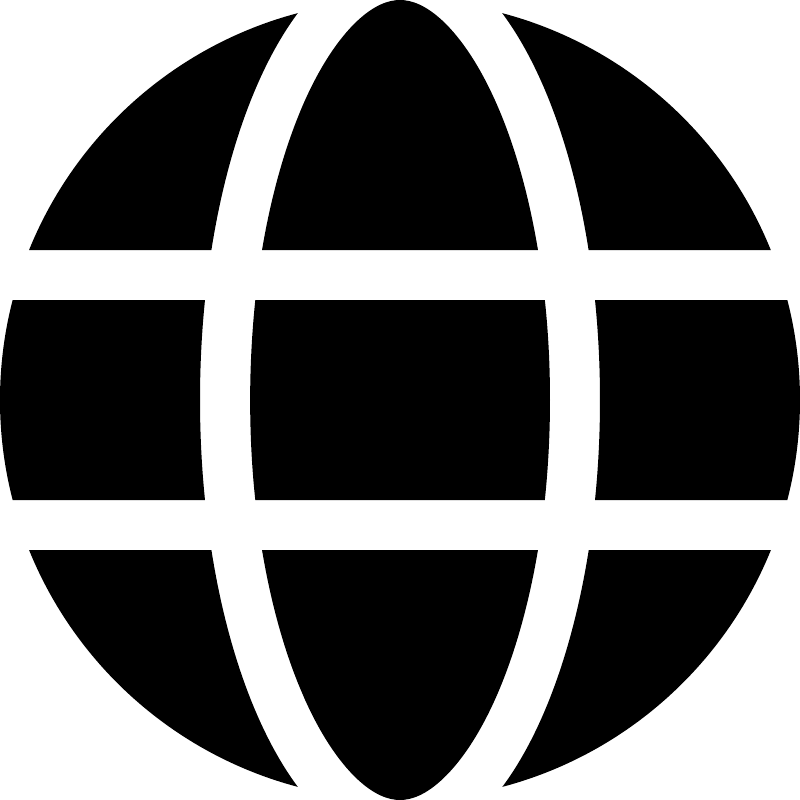}}\,Website}\qquad
\href{https://huggingface.co/datasets/kingofspace0wzz/CUA-SWE}{\raisebox{-2.5pt}{\includegraphics[height=13pt]{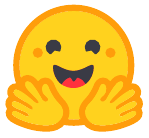}}\,Hugging Face}
\end{center}
\end{abstract}
\begin{figure}[H]
\centering
\begin{minipage}[c]{0.60\linewidth}
\centering
\includegraphics[width=\linewidth]{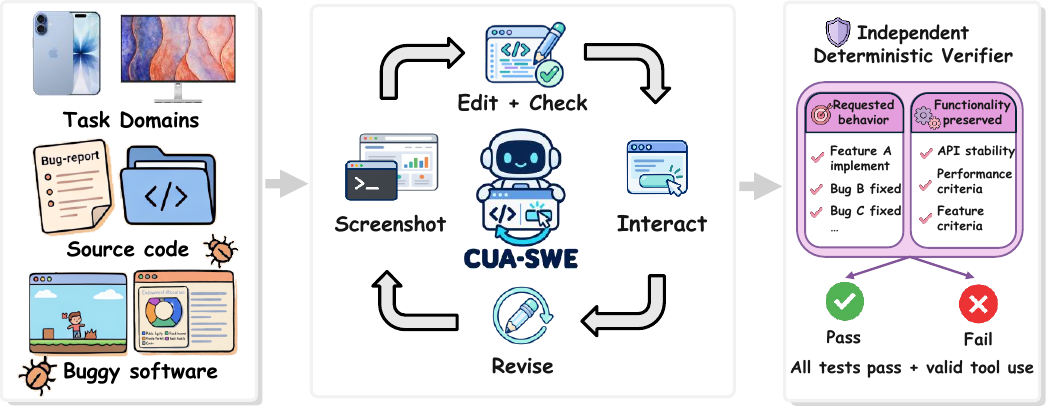}
\end{minipage}\hfill
\begin{minipage}[c]{0.38\linewidth}
\centering
\includegraphics[width=\linewidth]{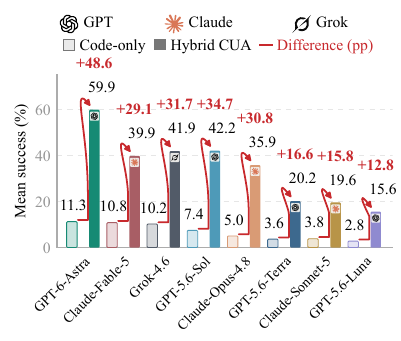}
\end{minipage}
\caption{\textbf{CUA-SWE benchmark overview and Frontier agents performance.} \textbf{Left:} Agents inspect screenshots, interact with applications, and edit code; independent tests check their repairs. \textbf{Right:} Mean task success of Code-only and Hybrid CUA agents across four equally weighted domains. Red labels show the Hybrid CUA gain in percentage points.}
\label{fig:teaser}\label{fig:computer-use-gain}
\end{figure}
\clearpage
\input{sections_preprint/01-introduction}
\input{sections_preprint/02-environment}
\input{sections_preprint/03-benchmark}
\input{sections_preprint/04-experiments}

\FloatBarrier
\input{sections_preprint/05-related-work}
\input{sections_preprint/06-limitations}
\input{sections_preprint/07-conclusion}
\label{maintext:end}
\section*{AI use statement}
AI tools assisted with writing, literature searches, code inspection,
conceptual and experimental design, result interpretation, and figure and
table preparation. An image-generation model assisted with figure layout
studies. Application screenshots and code excerpts in the figures come from
the tasks and recorded runs. The authors are responsible for the manuscript's
claims, citations, and experimental evidence.
\section*{Reproducibility statement}
Sections~\ref{sec:environment} and~\ref{sec:benchmark} describe the environment,
tasks, and tests. Section~\ref{sec:pomdp} specifies the problem formulation.
Section~\ref{sec:post-training} reports supervised post-training;
Appendix~\ref{app:sft-data} describes SFT data construction,
Appendix~\ref{app:sft} gives the training configuration, and
Appendices~\ref{app:rl-objective}--\ref{app:rl-implementation} describe the RLVR extension. Appendix~\ref{app:implementation} documents the implementation
and example tests, and Appendix~\ref{app:inventory} lists all tasks.
Section~\ref{sec:experimental-setup} defines the experimental conditions and metrics;
Appendix~\ref{app:results} describes the evaluation protocol, condition
comparisons, and workload accounting. Appendix~\ref{app:repeated-attempts}
specifies repeated-attempt metrics and statistical resampling.
Figure~\ref{fig:sft-training} reports the SFT loss history and
training-task verifier outcomes for the base and trained checkpoints.
\section*{Ethics statement}
The tasks use locally hosted applications and games, including original
examples and code adapted from existing projects. Distributing this material
requires preserving source and asset attribution. Evaluation tests are kept
separate from the agent's editable files to prevent changes to the scoring
procedure. The study does not measure human performance. Passing the benchmark
tests does not guarantee that a change is safe to deploy.
\clearpage
\bibliography{references}
\bibliographystyle{iclr2027_conference}
\clearpage
\appendix
\input{sections_preprint/08-appendix-benchmark}

\input{sections_preprint/09-appendix-rl}
\clearpage
\input{sections_preprint/10-appendix-results}

\clearpage
\input{sections_preprint/11-appendix-related-work}
\input{sections_preprint/12-appendix-trajectories}

\end{document}

%% file: sections_preprint/01-introduction.tex
\section{Introduction}

Developing interactive software requires an agent to use the software it
writes. Source code and command output do not fully reveal what users see
or what happens when they interact with an application~\citep{yang2024multimodal}. An agent needs
to run the software, operate its interface, and use visual clues to diagnose
problems and guide code changes~\citep{aggarwal2025pixels}. It must then check how those changes
affect the running software~\citep{lu2025webgenbench}. In games, this requires
observing behavior over time and checking responses to player
input~\citep{chi2026gamedevbench,luo2026gamecraft}.
Consider a platform game in which the character falls through a platform
after jumping onto it. The game builds and starts, and a static screenshot
may look correct. To diagnose the failure, an agent must play the game,
observe the jump and landing, and connect that behavior to the movement or
collision code. After editing the code, it must replay the jump and check
that the character lands correctly while normal movement still works.

Computer-use agents and coding agents approach software from complementary
directions. Computer-use agents operate interfaces through visual
observations and mouse and keyboard actions~\citep{agashe2025agents2,qin2025uitars,wang2025opencua},
while coding agents inspect repositories, edit files, and execute
commands and tests~\citep{yang2024sweagent,wang2024openhands}.
Their benchmarks typically assess either changes to a
codebase~\citep{jimenez2024swebench,miserendino2025swelancer} or the completion
of user tasks in existing applications~\citep{xie2024osworld,koh2024visualwebarena}.
Visual software benchmarks incorporate runtime feedback, but concentrate
on particular development domains, such as user-facing JavaScript
software and games~\citep{yang2024multimodal,chi2026gamedevbench}.
Broader environments emphasize access to development tools through a
visual IDE~\citep{aggarwal2025pixels}, or the orchestration of graphical
and command-line interfaces across general workflows assessed by an
agentic judge~\citep{li2026weavebench}.
Together, these capabilities motivate two central questions:
\textbf{(1) How do agents use GUI feedback to diagnose, repair, and verify software?}
\textbf{(2) Can agents complete software engineering tasks when required
specification or operational information is available only through the running
application's visual interface?} Visual observations can thus supply both
feedback on an implementation and requirements that the agent must recover
and translate into working code. Addressing these questions across software
engineering domains
calls for an environment that connects software use to source-code
modification, together with a benchmark that verifies the resulting
behavior. Deterministic tests provide reproducible checks of whether a
change satisfies the task and preserves specified functionality.

To address this challenge, we introduce \textbf{CUA-SWE}, a benchmark,
interactive development environment, and evaluation pipeline for visual software engineering with computer-use
agents, spanning web development, game development, mobile app development,
and DevOps. Tasks focus on diagnosing, repairing, and verifying software
behavior. Each task provides an editable project and running software
that the agent can operate and visually inspect throughout development. To reflect how developers work, the environment
lets agents choose when to inspect source code, execute commands, interact
with the software, examine screenshots, and revise their implementation.
Coding and computer use are available within the same development episode:
visual observations can guide code changes, and subsequent interaction
lets the agent inspect their effects.

CUA-SWE provides deterministic evaluation with verifiable task outcomes.
Each task specifies the behavior to implement or repair and the existing
functionality that must be preserved. Task-specific tests execute the
modified software in a controlled setting, using interaction sequences
and assertions on the rendered interface or runtime state to verify the
requested behavior and detect regressions. The environment also supports
a \emph{code-only} condition in which agents inspect and edit
source, execute permitted builds, tests, and custom scripts, and read their
outputs. Hybrid CUA adds screenshots and graphical interaction with the
running application. The same task requirements and protected tests apply
to both conditions, allowing us to measure development with source-level
execution and with additional application access through its GUI.

\Needspace{7\baselineskip}
Our contributions are:\nopagebreak[4]
\begin{itemize}
\interlinepenalty=10000
\item \textbf{An environment for software development through coordinated coding and computer use.}
Agents work with editable projects, development commands, mouse and
keyboard actions, and screenshot observations. Visual runtime evidence
connects rendered behavior to code changes, and subsequent interaction lets agents check
the effects of their edits.
\item \textbf{A benchmark with deterministic evaluation and verifiable outcomes.}
Across four software engineering domains, tasks connect visually observable
application behavior to an implementation or repair requirement and define
the existing functionality to preserve. Independent tests
verify the resulting software through execution and interaction, providing
consistent criteria for task success.
\item \textbf{An evaluation of frontier models and coding agents.}
We evaluate a broad range of frontier models and coding agents on CUA-SWE
and characterize task success across domains and application families,
including tasks that require recovering specification or operational information
from the running application's visual interface and translating it into working software.
The code-only reference includes nonvisual execution and
command feedback; hybrid CUA additionally provides application screenshots
and graphical actions. We examine this comparison by task information
requirements and use trajectories to study how agents connect execution
feedback and visual observations to implementation and verification.
\end{itemize}

%% file: sections_preprint/02-environment.tex
\section{When CUA Meets Visual Software Engineering}\label{sec:environment}\label{sec:pomdp}
\begin{figure}[t]
\centering
\includegraphics[width=\linewidth]{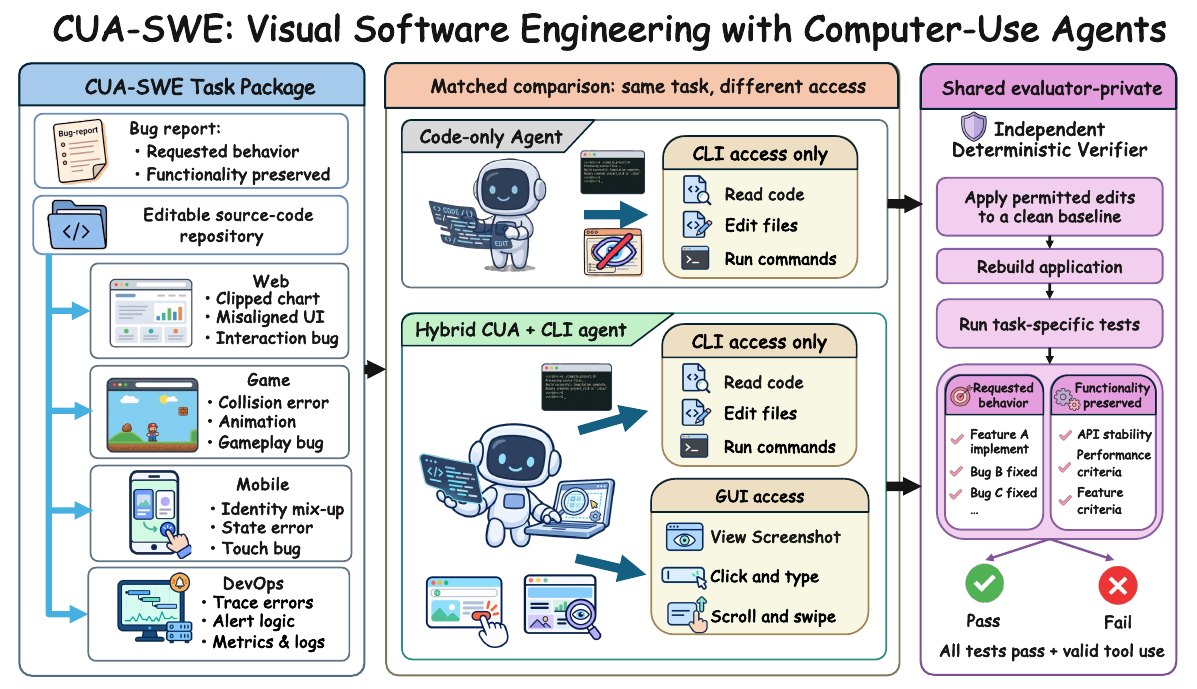}
\caption{\textbf{The CUA-SWE development and evaluation pipeline.}}
\label{fig:environment}
\end{figure}

The computer-use channel exposes the rendered application through
screenshots and graphical actions, while coding tools provide direct
source access. This observation boundary makes interpreting the visible
interface and choosing actions part of the measured capability. Structured
DOM, HTTP, and application-state queries would supply a different observation
channel; the benchmark's visual condition measures development through the
interface presented to the agent.

\paragraph{Problem Formulation.}
For task $u$, the agent receives an instruction $I_u$, an editable
codebase $c_u^0$, and access to its running application through screenshots
and graphical actions. The task also
defines a deterministic verifier $V_u$ that checks whether a submitted
codebase satisfies the requested functionality, preserves designated
existing behavior, and respects constraints on permitted changes.
The agent interacts with the development environment within a finite
action budget and produces a final codebase $c_T$, where $T$ is the number
of actions taken before submission or budget exhaustion. We denote its
changes relative to $c_u^0$ by $\Delta c$.
We model the interaction as a finite-horizon partially observable
Markov decision process
$\mathcal{M}_u=(\mathcal{S}_u,\mathcal{A},\mathcal{O},P_u,\Omega_u,R_u)$,
where $\mathcal{S}_u$, $\mathcal{A}$, and $\mathcal{O}$ are the state,
action, and observation spaces; $P_u$ and $\Omega_u$ are the transition and
observation models; and $R_u$ is the terminal repair reward determined by $V_u$.

\paragraph{State and observation.}
At step $t$, the environment state is
$s_t=(c_t,x_t,b_t)\in\mathcal{S}_u$, where $c_t$ is the current codebase
(with $c_0=c_u^0$), $x_t$ is the complete runtime state of the development
environment and running application, and $b_t$ is the remaining action
budget. The agent does not observe $s_t$ directly. Instead, it receives
an observation $o_t\in\mathcal{O}$ containing requested file contents,
command output, or screenshots of the running application. Screenshots
provide visual runtime evidence about the rendered interface and its
response to interaction. These observations give a partial view of execution: the visible
effect of a code change can depend on prior inputs and whether the
application has loaded the edit. After action $a_t\in\mathcal{A}$,
the environment evolves according to
$s_{t+1}\sim P_u(\cdot\mid s_t,a_t)$ and returns
$o_{t+1}\sim\Omega_u(\cdot\mid s_{t+1},a_t)$.

\paragraph{Action space, verification, and reward.}
The agent can modify the project, operate the software, or submit its work.
We define the agent's action space as follows:
$\mathcal{A}=\mathcal{A}_{\mathrm{coding}}\cup
\mathcal{A}_{\mathrm{computer}}\cup\{\mathrm{finish}\}$.
$\mathcal{A}_{\mathrm{coding}}$ contains commands for inspecting and editing
code and running development tools. $\mathcal{A}_{\mathrm{computer}}$
contains screenshot observation, mouse and keyboard input, and application
controls such as navigation and resizing. The $\mathrm{finish}$ action
terminates the episode and submits the final codebase $c_T$ for verification.
The complete interface is given in Appendix~\ref{app:actions}.
For a candidate codebase $c$, patch correctness combines three binary checks,
$V_u(c)=V_u^{\mathrm{task}}(c)\land V_u^{\mathrm{reg}}(c)
\land V_u^{\mathrm{perm}}(c)$.
Here $V_u^{\mathrm{task}}$ checks the requested implementation or repair,
$V_u^{\mathrm{reg}}$ checks designated existing functionality for regressions,
and $V_u^{\mathrm{perm}}$ checks that the modifications to $c_u^0$ respect
the task's restrictions on permitted changes. Behavioral checks use
controlled execution, interaction sequences, and assertions over runtime
or rendered application state. For evaluation episode $i$ on task $u$,
we write $P_i=V_u(c_{T_i})$ for patch correctness. A separate trajectory
check $U_i$ evaluates the episode's tool access and observation evidence
under its assigned condition. The reported task outcome is $S_i=P_iU_i$;
Section~\ref{sec:experimental-setup} specifies its aggregation.
The terminal repair reward is $R_u(s_T)=V_u(c_T)\in\{0,1\}$, which supplies
the correctness signal for post-training (Section~\ref{sec:post-training}).

\paragraph{Policy and objective.}
Let $h_t=(o_0,a_0,\ldots,o_{t-1},a_{t-1},o_t)$ denote the interaction
history. A vision-language policy with parameters $\theta$ selects actions
conditioned on the instruction and the interaction history,
$a_t\sim\pi_\theta(\cdot\mid I_u,h_t)$.
The same policy selects coding actions from $\mathcal{A}_{\mathrm{coding}}$
and computer-use actions from $\mathcal{A}_{\mathrm{computer}}$, allowing
visual runtime evidence from application interaction to guide subsequent edits.
The objective is to maximize the expected terminal reward
$\mathbb{E}_{\pi_\theta,P_u,\Omega_u}[R_u(s_T)]$, where the expectation is
over the policy's actions and the resulting environment interactions.
Because the reward is binary, this objective maximizes the probability
that the final codebase passes verification.

%% file: sections_preprint/03-benchmark.tex
\Needspace{10\baselineskip}
\section{The CUA-SWE benchmark}\label{sec:benchmark}\label{sec:construction}
\begin{figure}[htbp]
\centering
\includegraphics[width=\linewidth]{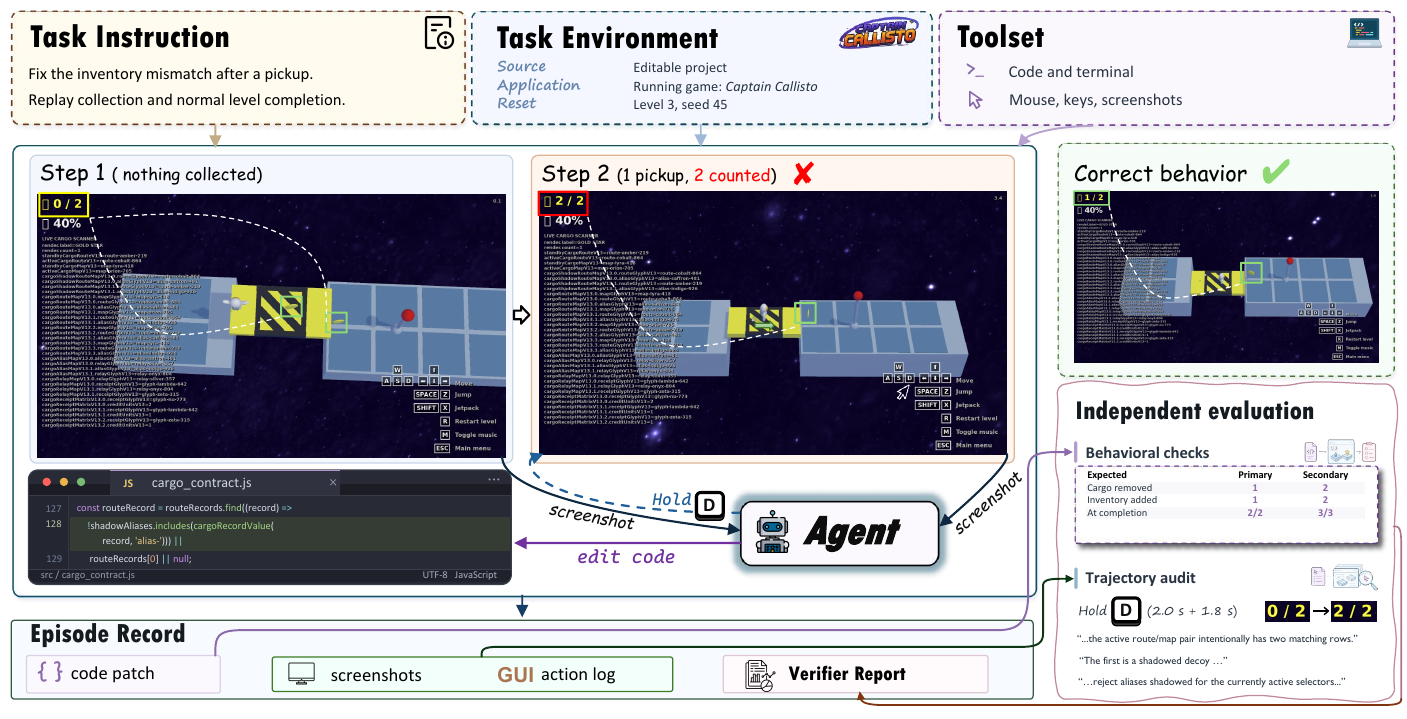}
\caption{\textbf{A concrete benchmark task: repairing inventory updates in Captain Callisto.}
Agents combine coding tools with interaction and screenshot observations.
In Captain Callisto, one pickup incorrectly changes the inventory
from 0/2 to 2/2.
The code excerpt shows the agent's final edit, while ``Correct behavior''
shows a separate reference replay.
Independent evaluation checks both the protected behavioral tests and
the validity of tool use.
The table lists expected outcomes for two replay scenarios.}
\label{fig:benchmark-example}
\end{figure}

\begin{figure}[t]
\centering
\includegraphics[width=\linewidth]{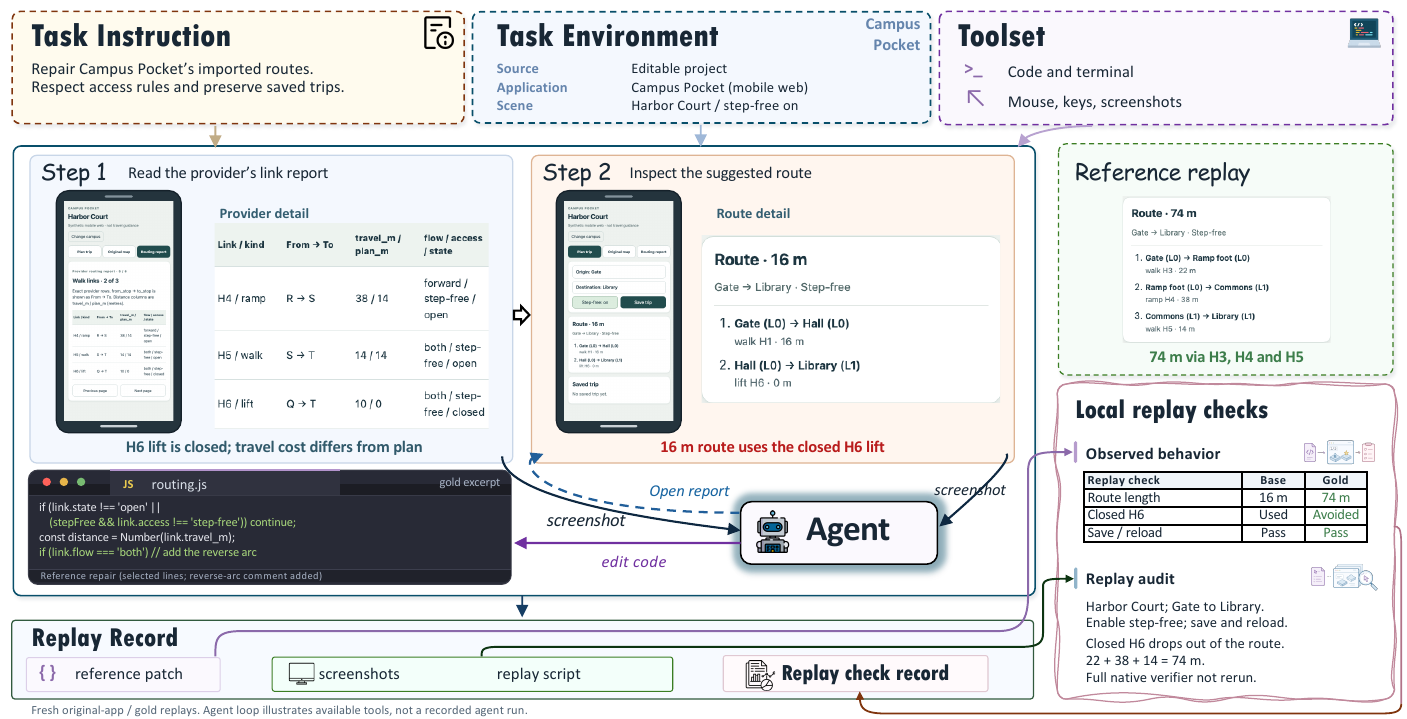}
\caption{\textbf{Mobile example: repairing imported routes in Campus Pocket.}
The provider report identifies a closed lift and the travel costs used to
construct a step-free route. The faulty route uses the closed H6 lift;
the reference repair follows H3, H4, and H5 for a total of 74 m.
The figure shows original-application and reference replays, with local
route and persistence checks. Its agent loop illustrates the available tools.}
\label{fig:mobile-task-example}
\end{figure}

\begin{figure}[t]
\centering
\includegraphics[width=\linewidth]{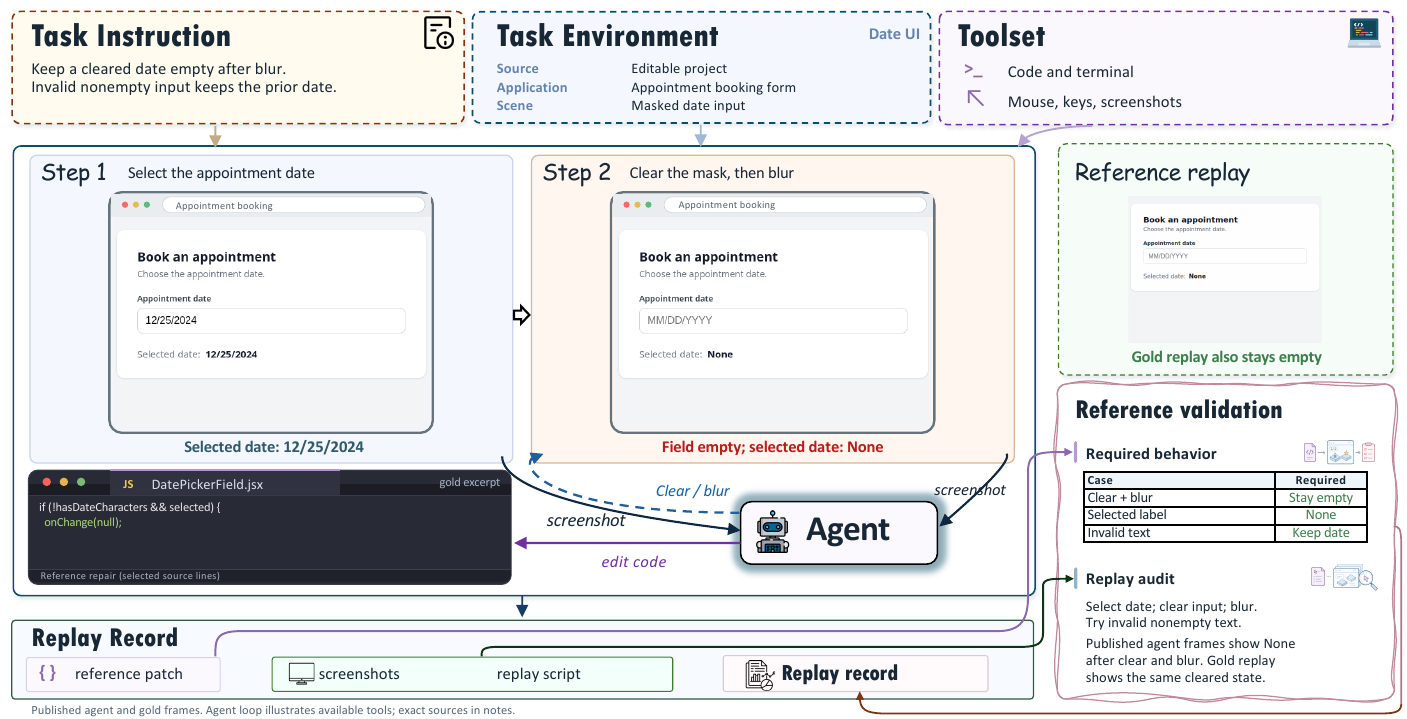}
\caption{\textbf{Web example: preserving distinct masked-date input behaviors.}
Clearing the appointment date and leaving the field should remove the
selection, while invalid nonempty input should preserve the previous date.
The published agent frames and reference replay show the cleared state.
The source excerpt is from the reference repair; the agent loop illustrates
the coding and interaction interface.}
\label{fig:web-task-example}
\end{figure}

We construct the benchmark around three
principles. First, tasks should expose visual runtime evidence that is relevant to
software development through the computer-use interface. Second, repair correctness
should be determined independently of the agent's trajectory by executing
$V_u$ on the submitted codebase $c_T$. Third, the construction process
should support controlled variations in task demands so that the benchmark
can probe where hybrid coding and computer-use agents succeed or fail.
We use LLM-assisted task authoring to scale this process, but admit a task
only after human review and executable validation of its requirement,
runtime behavior, reference solution, and verifier. We then use reviewed
construction-time agent trials to identify useful variations in difficulty.

\subsection{Tasks and domain coverage}

CUA-SWE contains 105 tasks spanning four software engineering domains:
36 Web, 29 Game, 20 DevOps, and 20 Mobile tasks. Each task instantiates
the same interface described in Section~\ref{sec:environment}: the agent
receives $I_u$ and $c_u^0$, can modify the project through
$\mathcal{A}_{\mathrm{coding}}$, and, in the computer-use setting, can
observe and operate the running software through
$\mathcal{A}_{\mathrm{computer}}$.
We select these domains because they expose different relationships between
runtime observations and implementation changes. Web tasks emphasize visual
layout, geometry, and interaction state; Game tasks involve motion, timing,
and state transitions; DevOps tasks connect service-level observations to
configuration and processing logic; and Mobile tasks couple displayed
interface behavior with application state. Tasks are adapted from open-source
applications and documented software behaviors or built as purpose-designed
applications when controlled behavior is needed. Appendix~\ref{app:task-sources}
describes the sources and adaptations, and Appendix~\ref{app:inventory}
lists the tasks.

\paragraph{Specification basis.}
Tasks also differ in where their requirements are specified.
In \emph{source-specified} (S) tasks, the instruction and readable project code
define the target behavior. \emph{Application-material-dependent} (M) tasks
require interpreting additional application materials, such as imported
drawings, service contracts, or graphical reference cards. These labels record
specification provenance; diagnosis, implementation, and verification are
required in both groups. For example, the gear task's drawing specifies
component relationships, while the submitted repair must realize those
relationships in the moving preview and preserve phase editing, undo, and
reload behavior.
Appendix~\ref{sec:runtime-information} defines the categories, and
Appendix~\ref{app:inventory} records each task's specification basis.

\subsection{Task construction and validation}\label{sec:development-verification}
\paragraph{From runtime behavior to a task specification.}
We begin with a concrete software behavior that can be reproduced in the
running application and whose implementation can be modified in the
codebase. From this seed, an LLM author constructs the instruction $I_u$, initial
codebase $c_u^0$, reproducible runtime setup, gold reference repair,
plausible negative repairs, and protected behavioral tests implementing
$V_u$. The author also specifies an interaction sequence that exposes
behavior relevant to the task through the agent's available observations.
Human review checks that the instruction describes an externally meaningful
development requirement, that the initial software exhibits the intended
problem, and that relevant runtime evidence is observable through the
benchmark interface. This step connects the task specification to the
partial observations $o_t$ available to the agent rather than relying on
information that is visible only to the benchmark author.
For example, in the Captain Callisto task in
Figure~\ref{fig:benchmark-example}, collecting one item changes the visible
inventory counter from (0/2) directly to (2/2). The task instruction asks
the agent to repair this behavior while preserving normal item collection
and level completion. The agent can reproduce the fault through game
controls and observe its effect in screenshots before modifying the code.

\paragraph{Constructing an independent verifier.}
For each task, we validate the reference repair $c_u^\star$ and the
deterministic verifier $V_u$ defined in Section~\ref{sec:environment}.
At minimum, we require $V_u(c_u^0)=0$ and $V_u(c_u^\star)=1$ 
so that the verifier distinguishes the original faulty program from a known
correct implementation.
The verifier combines checks of the requested behavior
$V_u^{\mathrm{task}}$, preserved functionality
$V_u^{\mathrm{reg}}$, and permitted modifications
$V_u^{\mathrm{perm}}$. Behavioral checks execute the submitted program
under controlled interaction sequences and may assert properties of the
rendered interface, runtime behavior, or internal application state.
Correctness therefore depends on the behavior of $c_T$, rather than on
matching the reference patch.
We further test the verifier with plausible incomplete or incorrect repairs.
These \emph{negative repairs} target common ways of satisfying only part of
the requirement and must remain rejected by $V_u$. For example, clearing a masked date field should remove the selected
date, whereas entering an invalid date should preserve the current selection.
A repair that clears the selection in both cases should fail verification. Gold and negative repairs
therefore serve different roles: the former establishes that a correct repair
exists, while the latter tests whether the verifier discriminates the
intended behavior from plausible partial fixes.

\paragraph{Validating the interactive development path.}
A valid verifier alone does not establish
that a task is suitable for computer-use software engineering. We therefore replay the relevant runtime interactions and check that the behavior needed to understand the task can
be exposed through the benchmark interface. We verify that the application
can be driven to the intended state, that relevant visual observations are
available, that code changes can take effect during development, and that
verification remains reproducible from a clean project state.
This validation deliberately separates information available during
development from information used for evaluation. The agent reasons from
its instruction $I_u$ and interaction history $h_t$, including source, command output, and
screenshots, whereas the evaluator may use privileged application state when
executing $V_u(c_T)$. For example, a game task may present only the rendered
board to the agent while the verifier directly inspects the underlying board
state. This allows evaluation to remain deterministic without exposing
privileged information to the agent. Appendix~\ref{app:examples} gives further verifier examples.

\subsection{Capability-oriented task expansion}

\noindent\textbf{Construction-time trials.}
After a task passes the preceding validation stages, we use construction-time
agent trials to calibrate its difficulty and identify related task variants.
These trials do not determine correctness (all candidate solutions are still
judged by the fixed verifier $V_u$), but they reveal whether a task is
trivial, beyond the tested agent's effective operating range, or informative
about a particular failure mode. Appendix~\ref{app:construction-trials} details this construction
and expansion loop (see Figure~\ref{fig:task-construction}).
\textbf{Controlled variants.}
When a task admits controlled variation, we construct related tasks by
changing a specific development demand while retaining the underlying
application and repair lineage. Examples include increasing the geometric
complexity of a selection problem, extending the temporal dependency of a
runtime state, or requiring an additional interaction condition to be
preserved. Each variant receives its own instruction $I_u$, initial
codebase $c_u^0$, and verifier $V_u$, and must independently pass the
same human review and executable validation described above.
\textbf{Task families.}
This process produces task families with related semantics but different
development demands. Rather than assigning a scalar difficulty label, the
families allow us to examine where agent performance changes as a particular
requirement becomes more demanding. They also provide a mechanism for
extending CUA-SWE as hybrid computer-use agents improve. Construction trials
are used only during benchmark development; the released task instructions,
initial codebases, runtime conditions, and protected verifiers are fixed
before the evaluations reported in Section~\ref{sec:experiments}.
Additional construction details and task-family definitions appear in
Appendices~\ref{app:construction-trials} and~\ref{app:task-families}, respectively.

%% file: sections_preprint/04-experiments.tex
\section{Experiments}\label{sec:experiments}
We study how agents use GUI feedback during software repair and whether they
can complete development tasks whose required specification or operational
information is available only through the running application's visual interface.
Task success measures the resulting implementation; information annotations
and trajectory analyses distinguish the demands of acquiring requirements,
diagnosing behavior, and implementing a repair.

\subsection{Experimental setup}\label{sec:experimental-setup}
\paragraph{Tasks and agents.}
We evaluate frontier computer-use agents and harness systems such as Codex and Claude Code on our CUA-SWE benchmark. Full evaluated model list can be found in~\ref{app:domain-performance}. 
Table~\ref{tab:four-domain-results-main} reports the
single-attempt comparison with code-only and hybrid CUA agents.
Appendix~\ref{app:repeated-attempts}, Table~\ref{tab:hybrid-pass3}, reports a three-attempt hybrid CUA evaluation
of GPT-6-Astra, GPT-5.6-Sol, and Claude-Fable-5 on the same task collections.

\paragraph{Coding and Hyrbird CUA.}
Both conditions allow agents to inspect and edit project
source, execute permitted builds, available tests, and custom scripts, and
inspect their outputs. The \emph{code-only} condition thus includes
nonvisual execution feedback. \emph{Hybrid CUA} additionally provides
screenshots and graphical interaction with the running application,
including its displayed state and materials. Agents choose when to execute
code, observe the interface, interact, and check their changes. Both
conditions use the same public task instructions and protected patch
tests. Appendix~\ref{app:access} details the access rules and gives a
source-level execution example. Codex and Claude Code are evaluated in
the CUA condition.

\paragraph{Metrics.}
For episode $i$ on task $u$, patch correctness is
$P_i=V_u(c_{T_i})$, combining the functionality, regression, and
permitted-change checks defined in Section~\ref{sec:environment}.
The trajectory indicator $U_i$ checks tool access and observation evidence
under the assigned condition. The task outcome is $S_i=P_iU_i$.
Within each domain, all models follow the same condition-specific scoring
rules (Appendix~\ref{app:success}). For evaluation set $\mathcal{E}$,
we report task success as $\mathrm{TSR}=100|\mathcal{E}|^{-1}\sum_{i\in\mathcal{E}}P_iU_i$.
Appendix~\ref{app:score-decomposition} reports patch
correctness, task success, and the proportion of passing patches with
$U_i=0$ for each model and access condition.
The four-domain mean gives Web, Game, DevOps, and Mobile equal weights.
Table~\ref{tab:four-domain-results-main} uses one attempt per task.
In Table~\ref{tab:hybrid-pass3}, pass@1 uses these same first-attempt
outcomes, while pass@3 counts tasks solved in the first or either of two
additional attempts. Their difference measures the task coverage added
by those attempts.
Appendix~\ref{app:workload-accounting} defines development-time
and workload measures.

\begingroup
\subsection{Frontier performance and computer-use access}\label{sec:results}
\input{tables/four-domain-results-main}
\paragraph{Broad competence across domains.}
GPT-6-Astra leads the four-domain comparison (Figure~\ref{fig:computer-use-gain}),
with 59.9\% mean task success, 17.7 percentage points above GPT-5.6-Sol.
It also achieves the highest observed hybrid success within every domain.
Its advantage spans graphical editing, game state, service behavior, and
mobile application integration. The task-level results locate this lead
more precisely: Canvas interactions after geometry changes and the recovery
of exact gear and sprite relationships distinguish it from other strong
agents. Table~\ref{tab:four-domain-results-main} summarizes these models;
Appendix~\ref{app:domain-performance} provides the complete results, and
Appendix~\ref{app:task-families} shows their distribution across task families.

\Needspace{6\baselineskip}
\paragraph{Code-only versus hybrid CUA.}\label{sec:computer-use-gains}
Hybrid CUA combines source-level execution with application screenshots
and GUI interaction (Figure~\ref{fig:computer-use-gain}). Every frontier
model in the four-domain comparison achieves higher aggregate task success
with hybrid access; GPT-6-Astra gains 48.6 percentage points. The gains
concentrate on tasks whose requirements include application materials;
source-specified subsets have near-zero domain-average changes and substantial
variation across models (Appendix~\ref{app:task-outcomes}). This comparison
measures completing visual software engineering tasks through the provided
access conditions. In material-dependent tasks, agents must recover requirements
from drawings or displayed behavior, implement them, and preserve the surrounding
workflow. Source-level execution tests the implementation; application
observations supply requirements and feedback that those executions alone
may not expose. The following analysis examines how agents turn this evidence
into verified repairs under hybrid access.

\subsection{Development behavior and reliability}\label{sec:analysis}
\paragraph{Following the consequences of a code change.}
The nested-selection task separates fixing a visible defect from preserving
the interaction that follows it. GPT-6-Astra first repairs resizing, then
drags the object beyond its containing frame. When selection is lost, it
extends hit testing to selectable descendants and repeats the interaction.
GPT-5.6-Sol and Claude-Opus-5 repair the resize behavior but leave the
selection defect in their submitted patches. The paired outcomes therefore
identify a concrete advantage: using feedback after an edit to uncover and
repair a second, connected behavior. Appendix~\ref{app:repair-cases}
traces the observations, changes, and verifier outcomes.

\paragraph{Recovering relations and translating them into behavior.}
On the Mobile gear task, GPT-6-Astra reconstructs all 23 driving relations
and implements their angle, phase, and editing behavior. GPT-5.6-Sol
instead chooses the wrong driving rim at two compound assemblies, causing
the implemented motion to depart from the drawing. A similar distinction
appears in sprite animation: GPT-6-Astra recovers the complete pose sequence,
whereas GPT-5.6-Terra's near-complete reconstruction retains incorrect pose
assignments. In DevOps, GPT-6-Astra uniquely solves the gauge-calibration
task by probing endpoint behavior and updating the inferred mode when the
deployment changes. These cases connect its lead to precise recovery of
relationships and operating rules, followed by implementation of their
consequences across application interactions.

\begin{figure}[H]
\setlength{\abovecaptionskip}{4pt}
\centering
\begin{minipage}[t]{0.485\linewidth}
\vspace{0pt}\centering
\vspace{-10pt}
\includegraphics[width=\linewidth]{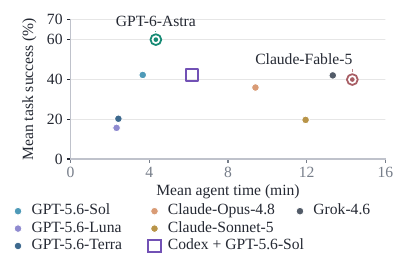}
\par\vspace{1pt}
{\small\textbf{(a)} Four-domain success and time}
\end{minipage}\hfill
\begin{minipage}[t]{0.485\linewidth}
\vspace{0pt}\centering
\vspace{-10pt}
\includegraphics[width=\linewidth]{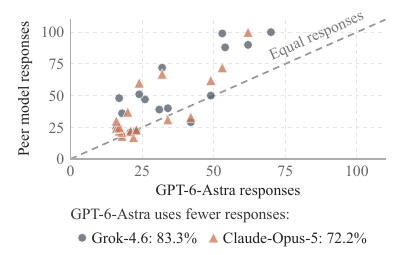}
\par\vspace{1pt}
{\small\textbf{(b)} Interaction on shared successes}
\end{minipage}
\caption{\textbf{Success, time, and interaction effort.}
\textbf{(a)} Task success rates and mean episode times for nine models and
systems, averaged equally across the four domains (circles: frontier models;
open square: Codex). Appendix~\ref{app:workload-accounting} specifies resource
coverage and aggregation.
\textbf{(b)} Completed responses on the same Web tasks solved by both
GPT-6-Astra and each peer (18 tasks per comparison); above the diagonal,
GPT-6-Astra uses fewer responses.}
\vspace{-10pt}
\label{fig:success-effort}\label{fig:shared-main}
\end{figure}
\begin{samepage}
\paragraph{Success and development effort.}\label{sec:efficiency}
Figure~\ref{fig:success-effort}(a) places task coverage alongside mean
development time. GPT-6-Astra's higher success than GPT-5.6-Sol accompanies
a mean episode time of 4.3 rather than 3.7 minutes, combining broader
coverage with a modest increase in time per attempt.
Its broader coverage is accompanied by economical interaction
on tasks that its peers also solve. It uses fewer responses on 83.3\%
of shared Web successes with Grok-4.6 and 72.2\% with Claude-Opus-5
(Figure~\ref{fig:success-effort}(b)). Each comparison holds the completed task
set fixed. Broader task coverage thus coexists with fewer interaction
rounds on shared successes. The repair traces show how individual
observations inform subsequent edits and checks. Appendix~\ref{app:shared-success}
retains the task-level effort pairs, and Appendix~\ref{app:resource-profiles}
reports the broader workload profiles.
\par\end{samepage}

\begin{wrapfigure}{r}{0.45\textwidth}
\centering
\includegraphics[width=\linewidth]{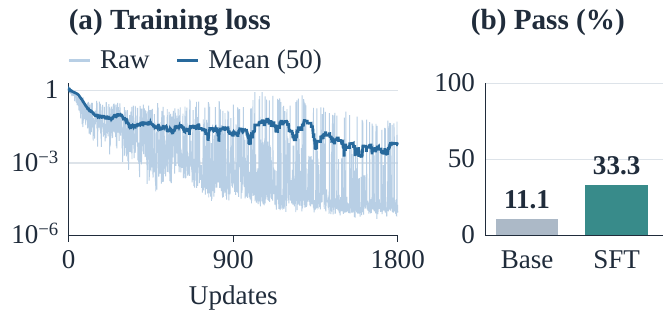}
\caption{\textbf{SFT performance on Web domain.}
(a) Batch loss and its 50-update mean (log scale).
(b) Native-verifier pass rate, one greedy attempt per task.}
\vspace{-5pt}
\label{fig:sft-training}
\end{wrapfigure}

\textbf{Capability boundaries and reliability.}\label{sec:repeated-attempts}
Reliable repairs require agents to coordinate coupled state and recover
relations that remain valid across subsequent interactions. Core Ball
exposes event-ordering demands, while Mobile schedule tasks require accurate
reconstruction of dense dependencies; local corrections can leave connected
behaviors unresolved (Appendix~\ref{app:capability-comparisons}). Repeated
attempts distinguish stable competence from additional task coverage.
In the three-attempt Game study, GPT-6-Astra solves 34.5\% of tasks in all
three attempts, compared with 3.4\% for GPT-5.6-Sol. On Mobile, later
attempts add a sprite repair for GPT-5.6-Sol and a schedule repair for
GPT-6-Astra. Thus retries broaden coverage, while consistency across attempts
captures how reliably agents complete these interacting requirements.
Appendix~\ref{app:repeated-attempts} gives the full repeated-attempt results
and uncertainty estimates.

\subsection{Supervised post-training}\label{sec:post-training}\label{sec:rl-experiments}

\paragraph{Learning from verified development trajectories.}
CUA-SWE also supplies supervision for hybrid development. We fine-tune
Qwen3.8-27B on action prefixes from verified reference replays with a
27/9 train/test split, and evaluate the base and final checkpoints on the
held-out tasks, with
one greedy attempt per task and matched interaction settings.
The native-verifier pass rate rises from 11.1\% to 33.3\%
after SFT (Figure~\ref{fig:sft-training}). SFT on gold-patch replays
trains the model to reproduce reference repairs, providing early positive
indications of a learning signal. This measure is patch correctness $P_i$;
benchmark task success additionally checks
trajectory validity $U_i$. Appendix~\ref{app:sft} preserves the full
training configuration, loss summaries, and evaluation details.

\paragraph{Extension to RLVR.}
Our immediate future work is to establish whether CUA-SWE provides a
verifiable learning signal and to develop scalable RLVR training in this
environment. The executable verifiers can score newly sampled development
episodes, supplying rewards for sequences of coding and computer use,
while parallel application workspaces support rollout collection.
Appendix~\ref{app:rl-objective} specifies the RLVR objective and rollout
interface.
\par\endgroup

%% file: tables/four-domain-results-main.tex
\begin{table}[htbp]
\centering\small
\caption{\textbf{Single-attempt task success (\%) across four domains.} Code denotes code-only access; hybrid CUA adds screenshots and graphical actions. Teal marks column maxima; an em dash indicates no reported result. Full results: Appendix~\ref{app:domain-performance}, Table~\ref{tab:four-domain-results}.}
\label{tab:four-domain-results-main}
\setlength{\tabcolsep}{3pt}
\resizebox{0.72\linewidth}{!}{%
\begin{tabular}{@{}lrrrrrrrr@{}}
\toprule
& \multicolumn{2}{c}{Web} & \multicolumn{2}{c}{Game} & \multicolumn{2}{c}{DevOps} & \multicolumn{2}{c}{Mobile} \\
\cmidrule(lr){2-3}\cmidrule(lr){4-5}\cmidrule(lr){6-7}\cmidrule(lr){8-9}
Model & Code & \shortstack{Hybrid\\CUA} & Code & \shortstack{Hybrid\\CUA} & Code & \shortstack{Hybrid\\CUA} & Code & \shortstack{Hybrid\\CUA} \\
\midrule
GPT-6-Astra & 27.8 & \textcolor[HTML]{167D8D}{\textbf{66.7}} & \textcolor[HTML]{167D8D}{\textbf{17.2}} & \textcolor[HTML]{167D8D}{\textbf{37.9}} & 0.0 & \textcolor[HTML]{167D8D}{\textbf{80.0}} & 0.0 & \textcolor[HTML]{167D8D}{\textbf{55.0}} \\
GPT-5.6-Sol & 19.4 & 58.3 & 10.3 & 10.3 & 0.0 & 55.0 & 0.0 & 45.0 \\
Claude-Opus-5 & 22.2 & 55.6 & 13.8 & 20.7 & 0.0 & 60.0 & \textemdash & \textemdash \\
Claude-Opus-4.8 & 16.7 & 27.8 & 3.4 & 20.7 & 0.0 & 55.0 & 0.0 & 40.0 \\
Claude-Sonnet-5 & 8.3 & 0.0 & 6.9 & 3.4 & 0.0 & 40.0 & 0.0 & 35.0 \\
Claude-Fable-5 & 19.4 & 47.2 & 13.8 & 17.2 & \textcolor[HTML]{167D8D}{\textbf{5.0}} & 55.0 & \textcolor[HTML]{167D8D}{\textbf{5.0}} & 40.0 \\
Grok-4.6 & \textcolor[HTML]{167D8D}{\textbf{30.6}} & 55.6 & 10.3 & 17.2 & 0.0 & 65.0 & 0.0 & 30.0 \\
\bottomrule
\end{tabular}}
\end{table}

%% file: sections_preprint/05-related-work.tex
\section{Related work}\label{sec:related-work}
\noindent\textbf{Software engineering and coding benchmarks.}
HumanEval, LiveCodeBench, and BigCodeBench test program synthesis and
execution~\citep{chen2021code,jain2024livecodebench,zhuo2025bigcodebench};
SWE-bench, SWE-bench Pro, and SWE-Lancer assess repository changes and
professional development tasks~\citep{jimenez2024swebench,deng2025swebenchpro,miserendino2025swelancer}.
Terminal-Bench extends evaluation to complex command-line
workflows~\citep{merrill2026terminalbench}. CUA-SWE connects software changes
to runtime observations and graphical interaction, with deterministic tests
of the resulting behavior.
\textbf{Computer-use benchmarks and agents.}
Mind2Web, WebArena, and VisualWebArena study web
interaction~\citep{deng2023mind2web,zhou2023webarena,koh2024visualwebarena};
OSWorld, Windows Agent Arena, and AndroidWorld cover desktop and mobile
workflows~\citep{xie2024osworld,bonatti2024windowsarena,rawles2024androidworld}.
Agent S2, UI-TARS, and OpenCUA develop planning, grounding, and learned
policies for computer use~\citep{agashe2025agents2,qin2025uitars,wang2025opencua}.
CUA-SWE studies these capabilities within development, where agents use
software to inform and check their own code changes.
\textbf{Connecting coding and computer use.}
Programming with Pixels combines visual IDE interaction, coding tools, and
programmatic verification~\citep{aggarwal2025pixels}. WeaveBench evaluates
hybrid GUI, CLI, and coding workflows through a trajectory-aware agentic
judge~\citep{li2026weavebench}. CUA-SWE focuses on software engineering across
four domains, providing deterministic correctness criteria for evaluation
and verifiable feedback for post-training. Appendix~\ref{app:extended-related-work}
extends these comparisons to visual software development, agent interfaces,
and training data.

%% file: sections_preprint/06-limitations.tex
\section{Limitations}\label{sec:limitations}
The initial SFT study offers early positive indications of a learning signal
from gold-patch replays (Section~\ref{sec:post-training}). Establishing whether
the environment provides a verifiable learning signal and developing
scalable RLVR training remain immediate future work. This requires efficient parallel
rollout collection, stable optimization over long multimodal trajectories,
and effective credit assignment across code edits, observations, and
interactions. Appendix~\ref{app:rl-objective} specifies an objective based on
executable task feedback.
The current benchmark spans Web, Game, DevOps, and Mobile. Expanding to
more domains, applications, and tasks will broaden its coverage of visual
software engineering. Future extensions can include richer development
workflows and longer tasks that require agents to coordinate coding and
computer use across multiple tools and applications.

%% file: sections_preprint/07-conclusion.tex
\section{Conclusion}
CUA-SWE is a benchmark, environment, and evaluation pipeline for visual
software engineering across Web, Game, DevOps, and Mobile. It connects code
editing and execution with screenshots, GUI actions, and deterministic
verification. Comparing code-only and hybrid CUA agents, our evaluation
examines how agents use visual runtime evidence to diagnose and repair
software and recover requirements supplied through application interfaces.
CUA-SWE supports measuring and expanding agents' ability to translate this
evidence into verified software changes.

%% file: sections_preprint/08-appendix-benchmark.tex
\section{Evaluation protocol and measurement}\label{app:results}
\subsection{Patch correctness and trajectory validation}\label{app:success}
\paragraph{Patch correctness.}
The verifier applies the submitted changes to a clean task
workspace and checks the requested functionality, designated existing
behavior, and restrictions on permitted changes. These are the three
components of $P_i=V_u(c_{T_i})$ in Section~\ref{sec:environment}.
The same patch checks apply to code-only and hybrid CUA executions.

\paragraph{Trajectory validation.}
The indicator $U_i$ checks that the execution uses the
interfaces allowed by its assigned condition and satisfies the domain's
observation-evidence requirements. Protected evaluation materials remain
outside the agent's access. Web and Mobile require screenshot observations
for CUA success. DevOps verifies screenshot use through its execution
records. Game validates the assigned CUA capability and records screenshot
use as a trajectory measure. Task success is $S_i=P_iU_i$.

\paragraph{Screenshot observations.}
Screenshot records link the rendered application to the
agent's visual observations. The Web API interface associates each
screenshot returned by the application launcher with its image-view call.
Interfaces with direct image attachments record those observations in the
model request; the Mobile audit links captures to the corresponding
request and response.

\subsection{Agent capabilities and access rules}\label{app:access}
Both conditions receive the public task instruction and
editable project source. Agents can read code and configuration, modify
permitted files, run nonvisual builds and available project tests, and
execute custom scripts against project modules. Command output, errors,
and test results are returned to the agent. For example, the evaluated
2048 code-only trajectory in Appendix~\ref{app:code-execution-example}
loads game modules in Node, simulates key events and timers, and checks
restart behavior. Protected evaluation tests and reference patches remain
outside the agent's workspace.

Code-only provides this source-level execution interface.
It excludes application screenshots, graphical actions, and access to the
served application through a development server, direct HTTP or sockets,
or browser automation. Hybrid CUA adds the designated screenshot-and-action
interface: the evaluator starts the application, and the agent observes
pixels and issues supported mouse, keyboard, and navigation actions.
Direct DOM reads, accessibility trees, script evaluation in the browser,
and direct HTTP extraction are outside this observation interface.

The same task instruction and protected tests apply to both
conditions. In Mobile, both receive the ordinary client source and build
inputs; the prepared app also contains imported product materials such as
drawings and issuer records. These materials can be inspected through the
hybrid condition's application interface. Code-only retains its nonvisual
execution and command feedback on the supplied client code.

\subsection{What visual runtime evidence contributes}\label{sec:runtime-information}
\paragraph{Information and feedback.}
Visual runtime evidence consists of the rendered application and its
visible responses to graphical actions. Screenshots capture layout,
spatial relationships, displayed text and values, and changes across
interaction states. These observations supply task-specific information
and support diagnosis and verification. Information absent from the
editable source can include an imported drawing, a service's displayed
response, or a target shown in the application. The Mobile gear task, for example, requires reconstructing
connections from received assembly drawings before implementing the
corresponding motion. Its state report describes the current client model;
the received drawing specifies the intended component relations. The permitted
code-only tests and scripts exercise the supplied client code, but do not expose
the imported drawing. Agreement with that drawing requires reconstructing
its relations and implementing their consequences for motion, phase, editing,
and persistence. During development, an agent can also inspect a failure,
edit the code, and exercise the modified behavior, as in checking a resize
gesture or a restart. These activities connect visual requirement recovery
to implementation and verification within the same task.

\paragraph{Task information requirements.}
We annotate each task by the location of its specification.
\emph{Source-specified} (S) tasks define the target behavior in the public
instruction and readable program text; executing and observing the
application supports diagnosis and checking. \emph{Application-material
dependent} (M) tasks specify part of the target behavior through application
materials, such as an imported drawing, a displayed service contract, or
a graphical reference card. Tasks requiring both material interpretation
and runtime diagnosis receive M. The labels describe specification
provenance; they do not partition tasks into information acquisition versus
debugging. Both groups require implementing the requested behavior and
verifying the resulting software. M includes runtime-supplied materials
and bundled graphical references, as detailed below.

\paragraph{Annotation basis.}
We inspect task instructions, readable source, reference
materials, and the boundary between the coding workspace and the prepared
application. Model outcomes are not a classification criterion. A runtime
fixture alone does not imply M: 2048 specifies restart invariants in its
instruction, Fabric exposes its Help in readable HTML, and Flappy Bird
includes target impulses and arcs in its source. These tasks receive S.
The received gear drawings and live service contracts receive M because
they supply the required interpretation. Five Boxel tasks also receive M:
their restart rule appears on a graphical card bundled with the project,
which the code-only interface does not display. Thus M covers 58 tasks
with runtime-supplied materials and five with bundled graphical references.
Appendix~\ref{app:inventory} gives every task's label and specification basis.

\subsection{Condition comparisons}
Code-only and CUA comparisons pair the same tasks within each model
and domain.

The four-domain mean weights each domain's task success rate equally.
CUA success is computed over the full task inventory for the nine entries
evaluated in all four domains.
Figure~\ref{fig:computer-use-gain} uses matched tasks for eight frontier models,
forming 32 model--domain cohorts and 840 paired model--task observations.
Differences are computed from the unrounded means.

For the information-requirement analysis, we partition
these same tasks by the labels in Appendix~\ref{sec:runtime-information}.
Within each domain, category, and model, code-only and hybrid rates use
identical task membership and the original selected attempts. We report
the number of tasks, both success rates, and their percentage-point
difference. Summary rates average the same eight frontier models with both
conditions evaluated in all four domains; full per-model results also
include Claude-Opus-5 where evaluated. A category absent from a domain is not
applicable. Appendix~\ref{app:task-outcomes} reports the results.

Related tasks share applications and interaction patterns: Mobile's
20 tasks cover 12 families, including nine Gantt variants, and Game's
29 tasks cover 12 games. Task-weighted scores measure coverage of the
specified requirements; application-family analyses characterize breadth.

\subsection{Single- and repeated-attempt evaluation}\label{app:cohort-reconciliation}
\paragraph{Evaluation design.}
Table~\ref{tab:four-domain-results} reports one attempt per task.
For GPT-6-Astra, GPT-5.6-Sol, and Claude-Fable-5, we combine these
first-attempt outcomes with two further executions per task, each starting
from a fresh workspace. Pass@1 uses the first attempt, and pass@3 counts
tasks solved in any of the three attempts. Pass@1 matches
Table~\ref{tab:four-domain-results}. Appendix~\ref{app:repeated-attempts}
reports cumulative coverage, success frequencies, and uncertainty estimates.

\paragraph{Game evaluation.}
The three rounds cover the same 29 task definitions and protected verifiers.
Game success follows the native verifier under the assigned-access policy;
browser use and screenshot grounding are recorded as trajectory measures.
For each model and task, pass@1 uses the first execution, as in
Table~\ref{tab:four-domain-results}. Cumulative pass@$k$ counts success
in at least one of the first $k$ executions.
All repeated-attempt statistics are computed from these task-level triples.

\subsection{Execution and workload accounting}\label{app:workload-accounting}
\paragraph{Execution settings.}
Model, provider, interface, and execution settings define each evaluated
system. Mobile uses a budget of 60 model responses and 2,700 agent seconds,
a maximum of 16,384 output tokens per request, high reasoning, and a
600-second request timeout. Its client makes one HTTP attempt per request.
The Web Anthropic client permits up to ten HTTP attempts with a 300-second
timeout per attempt and exponential backoff capped at 30 seconds.
HTTP retries occur within a task trial.

\paragraph{Resource measures.}
Agent wall time measures the development episode; setup and final
verification have separate timers where available. DevOps uses the
episode timer in its budget attestation. For frontier models, steps count assistant
responses, Codex steps count completed items, and Claude Code steps use
the larger of assistant records and terminal turns. Native system step
counts are reported in their respective units. Screenshot captures and
GUI actions measure interaction with the application. Token accounting
includes input, cached-input, and output tokens; Web has final token
records for 718 of its 720 single-attempt executions. Response length is
output tokens divided by completed responses in an episode.

\paragraph{Aggregation.}
Workload summaries use recorded episode resource
measurements. Four-domain summaries average episode measurements within each
domain and weight the domains equally. Agent minutes per success is the
ratio of this mean episode time to the full-inventory success probability.
Domain-level profiles report medians, with workload intervals spanning
the 25th--75th percentiles. Shared-success analyses pair tasks completed
successfully by both models with resource measurements available for
both executions, then summarize their per-task effort ratios.

\subsection{Task sources and adaptations}\label{app:task-sources}
\paragraph{Provenance and reproducibility.}
Task sources include open-source applications, documented software
behaviors, and purpose-built applications with controlled faults.
The date-input task adapts a reported clear-on-blur defect in React
Datepicker; the DevOps gauge task implements update semantics from StatsD;
and Mobile tasks use synthesized applications that import visual materials.
Adaptations preserve the relevant behavior while simplifying setup for
reproducible execution. Construction records identify the source revision
and changes made, together with the initial conditions, triggering
interactions, and observable outcomes. Seeds, input materials, and service
configurations reproduce the conditions needed to investigate the task.

\paragraph{Domain coverage and task families.}
Web tasks cover interface geometry, selection, and application contracts.
Game tasks cover temporal behavior, physics, and state transitions across
12 games. DevOps tasks connect observations from service interfaces to
configuration and data-processing semantics. Mobile tasks cover visually
specified relationships and editing behavior across 12 mobile-web
application families. Tasks retain their source application, failure
mechanism, and shared implementation in their lineage; the nine Mobile
wall-chart scheduling tasks, for example, belong to one family.
New applications broaden behavioral coverage, while controlled variants
extend difficulty within a family. Appendix~\ref{app:task-families}
reports family-level performance.

\clearpage
\subsection{Seeded task construction and difficulty expansion}\label{app:construction-trials}
\begin{figure}[htbp]
\centering
\includegraphics[width=\linewidth]{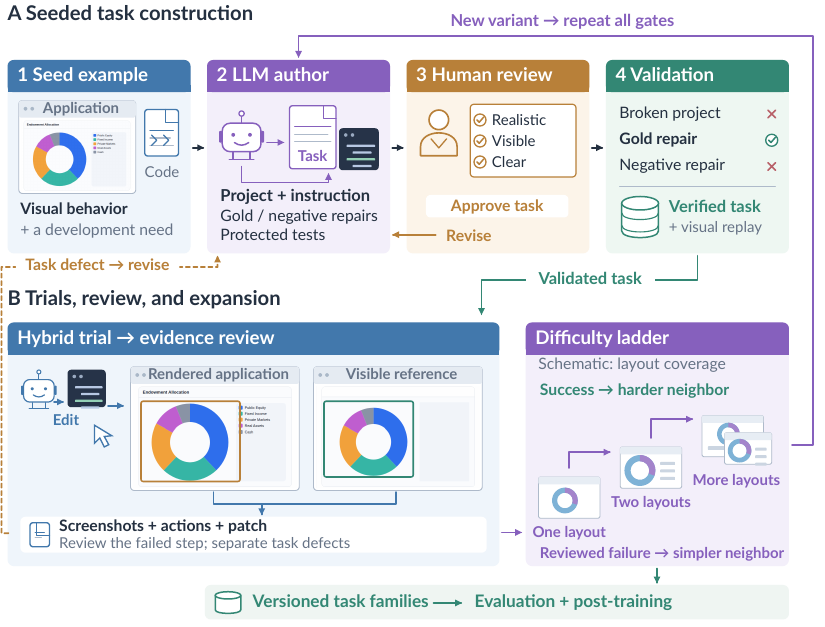}
\caption{\textbf{From a seed example to a verified difficulty ladder.}
An LLM author turns a seed into a runnable task. Human review and executable
checks establish task validity. Reviewed agent trials drive a separate
expansion loop, with every new variant returning through both gates.
The chart panels show the rendered application and its visible reference
from a recorded observation. The ladder illustrates increasing layout
coverage; trials assess the difficulty of proposed variants.
Validated task versions support benchmark evaluation and post-training.}
\label{fig:task-construction}
\end{figure}

Construction begins with a seed example and produces a
family of executable visual software engineering tasks.
Figure~\ref{fig:task-construction} shows the process: an LLM author
creates the task, human review checks its requirements and visual
evidence, and executable validation establishes a correct repair.
Reviewed agent trials then guide the generation of neighboring tasks
that extend the family's difficulty range.

\paragraph{Seed examples and task authoring.}
A seed specifies an application, a concrete development need,
and an interaction that exposes the relevant visual behavior. Seeds come
from the application sources described in Appendix~\ref{app:task-sources}.
The LLM author uses this context to construct a runnable project, a task
instruction, a gold reference patch, plausible incomplete repairs, and
protected behavioral tests. In the Captain Callisto inventory example
(Figure~\ref{fig:benchmark-example}), the seed links collecting an item to
an incorrect counter update. Authoring turns that example into a requirement
to correct inventory updates while preserving cargo removal and level
completion, with reproducible scenarios that check these behaviors.

\paragraph{Human review and executable validation.}
Human reviewers inspect the candidate requirement, the
visual evidence available through interaction, and the expected outcome.
They check that the development need is meaningful, the instruction is
unambiguous, and the screenshots expose the behavior needed to investigate
it. Review feedback returns to the LLM author to revise the candidate.
Executable checks then require the broken project and incomplete repairs
to fail, and the gold patch to pass the protected tests. Interaction replay
checks the corresponding visible behavior. The gold patch establishes an
executable solution; negative repairs test whether the verifier rejects
partial fixes. A candidate with inconsistent behavior, an inaccessible
trigger, or an inadequate verifier returns for correction and revalidation.

\paragraph{Agent trials and reviewed failures.}
Hybrid agents attempt validated tasks by observing the
application, diagnosing its behavior, editing code, and checking the result.
A successful trajectory supplies a concrete solution under the tested
interface and budget. For a failed attempt, review connects the outcome to
the screenshots, actions, and code changes, identifying the step at which
the agent lost the relevant evidence or implemented an incorrect repair.
Task ambiguity, runtime faults, and verifier defects trigger corrections
to the task. Failures attributed to observation, reasoning, or implementation
instead provide a target for expanding the task family.

\paragraph{Difficulty ladders and iterative expansion.}
The LLM author generates neighboring variants by changing
one declared demand while retaining the underlying application and
behavioral requirement. A reviewed failure motivates a simpler neighbor
that isolates the difficult step; a successful repair motivates a harder
neighbor that extends it. For example, a selection task can vary the degree
of nesting, while a temporal task can vary the history needed to interpret
a displayed state. Every variant passes through the same human review and
executable checks before another trial. Adjacent successes and reviewed
failures locate a capability boundary for the tested agent and budget;
trial outcomes determine whether the proposed ordering forms a useful
ladder. Repeating this loop extends the benchmark toward more demanding
visual SWE problems as agents improve. Task versions retain their seed
and family lineage, preserving the distinction between new applications
and difficulty variants when reporting benchmark coverage and performance.

\section{Execution protocol and verifier examples}\label{app:implementation}
\subsection{Execution and evaluation isolation}
Each episode starts from a working copy of the task's source and a running
application. Browser actions are validated for required arguments and
permitted values before execution. Visual observations contain screenshots
without DOM text or accessibility metadata.

The protected evaluation procedure keeps tests separate from editable files
and checks the integrity of evaluation files. Patch-only verification
reconstructs the application and any task-specific services, applies the
candidate repair, and executes the hidden tests without invoking an agent.
This separates the behavior of the submitted software from the agent's own
claims or development checks.

Task setup specifies dependency installation, application startup and reset,
and the viewport or device. Task-specific inputs, such as game seeds, are
fixed where applicable. Each attempt receives a fresh project copy. The run
protocol limits agent decisions and elapsed time, with an additional
generated-token cap where configured. Interaction ends when the agent
selects $\mathrm{finish}$ or reaches an applicable limit; the candidate
change consists of the source and configuration edits accumulated by then.

\subsection{Action interface}\label{app:actions}
Each action specifies an operation and its admissible arguments.
Shell actions take a command string and a timeout in seconds, capped by
the remaining episode time. The computer interface is listed below;
operations without arguments use an empty argument tuple.
\begin{center}
\small
\begin{tabular}{@{}p{0.25\linewidth}p{0.70\linewidth}@{}}
\toprule
Action category & Operations and arguments\\
\midrule
Startup and observation & $\mathrm{start}$, $\mathrm{observe}$\\
Mouse input & $\mathrm{click}(p)$, $\mathrm{scroll}(p,\Delta p)$,
 $\mathrm{drag}(p,p',n)$\\
Drag continuation & $\mathrm{drag\_start}(p)$,
 $\mathrm{drag\_move}(p,n)$, $\mathrm{drag\_end}(p,n)$\\
Keyboard input & $\mathrm{type}(\xi)$, $\mathrm{press}(k)$,
 $\mathrm{key\_hold}(k,d)$\\
Browser navigation & $\mathrm{back}$, $\mathrm{forward}$\\
Viewport and timing & $\mathrm{resize}(w,h)$, $\mathrm{wait}(d)$\\
\bottomrule
\end{tabular}
\end{center}
Here $p,p'$ are viewport pixel coordinates, $\Delta p$ is a scroll
displacement, $\xi$ is text, $k$ is a key or key combination, $n$ is an
interpolation-step count, and $(w,h)$ gives viewport dimensions.
Computer-action durations $d$ use milliseconds. Optional arguments use
the tool's defaults. The separate $\mathrm{finish}$ action submits the
current workspace for verification.

\subsection{Verifier examples}\label{app:examples}
The support-appointment and 2048 restart tasks illustrate how tests
distinguish complete repairs from changes that fix only part of a requirement.
For both tasks, the original buggy version and an intentionally incomplete
repair fail the tests, while the reference repair passes.

For the support-window task, the buggy version failed both service-response
tests. The incomplete repair displayed the first appointment correctly but
reused that value for the second appointment. The reference repair passed
both tests while preserving the schedule details.

For 2048, the reference repair passed tests with seeds 42, 1337, and 3735928559.
The buggy version and incomplete repair both allowed held input to change the
new board after a restart. The tests compare board contents, score, occupied
cells, maximum tile, and seed. They record a fresh board and a sequence of
moves, hold a direction key through each restart method, and check that the
new board and subsequent moves match the originals. A separate check confirms
that holding a key still changes the board during normal play.

\begin{figure}[htbp]
\centering
\begin{minipage}{0.49\linewidth}
\centering\includegraphics[width=\linewidth]{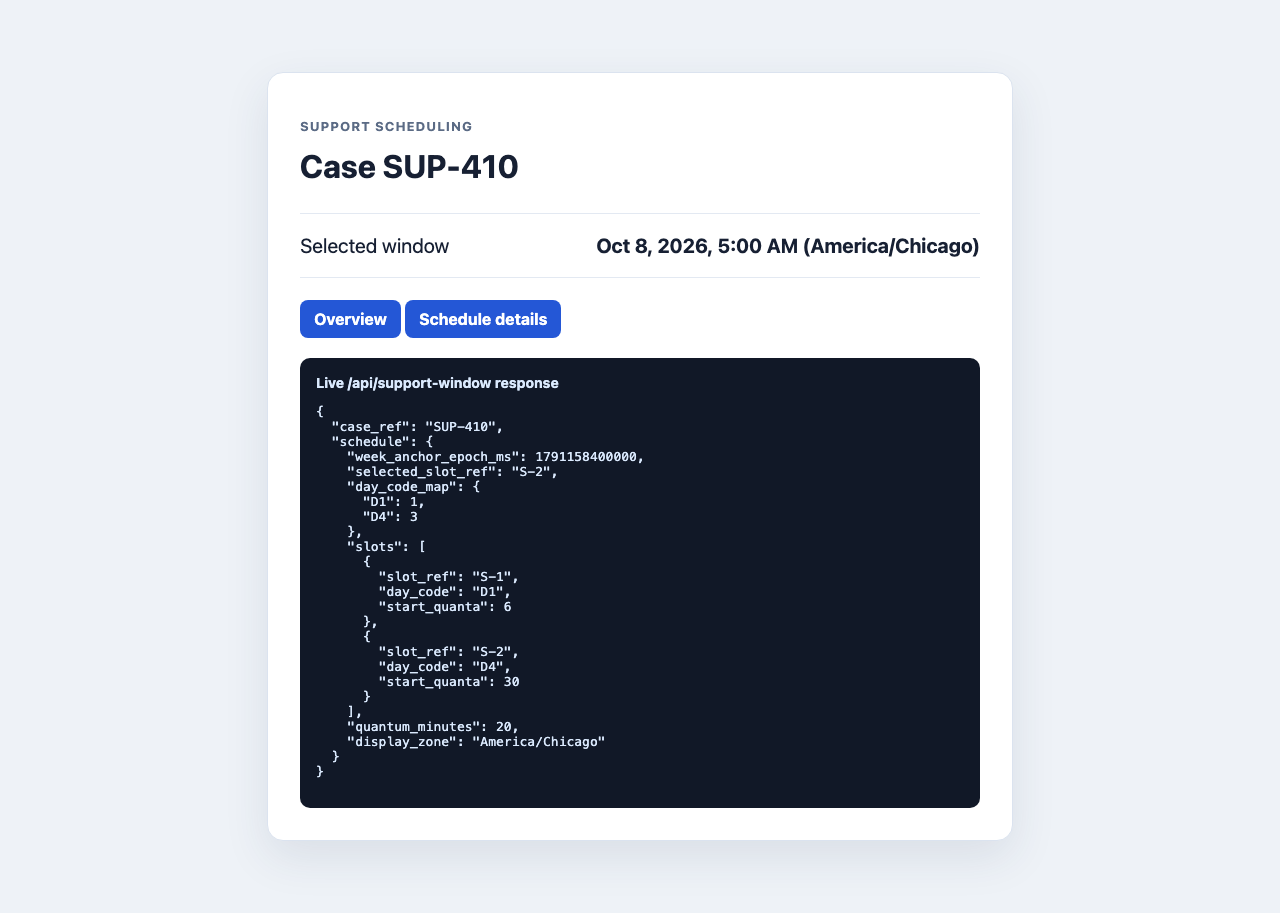}
\small (a) Support schedule
\end{minipage}\hfill
\begin{minipage}{0.49\linewidth}
\centering\includegraphics[width=\linewidth]{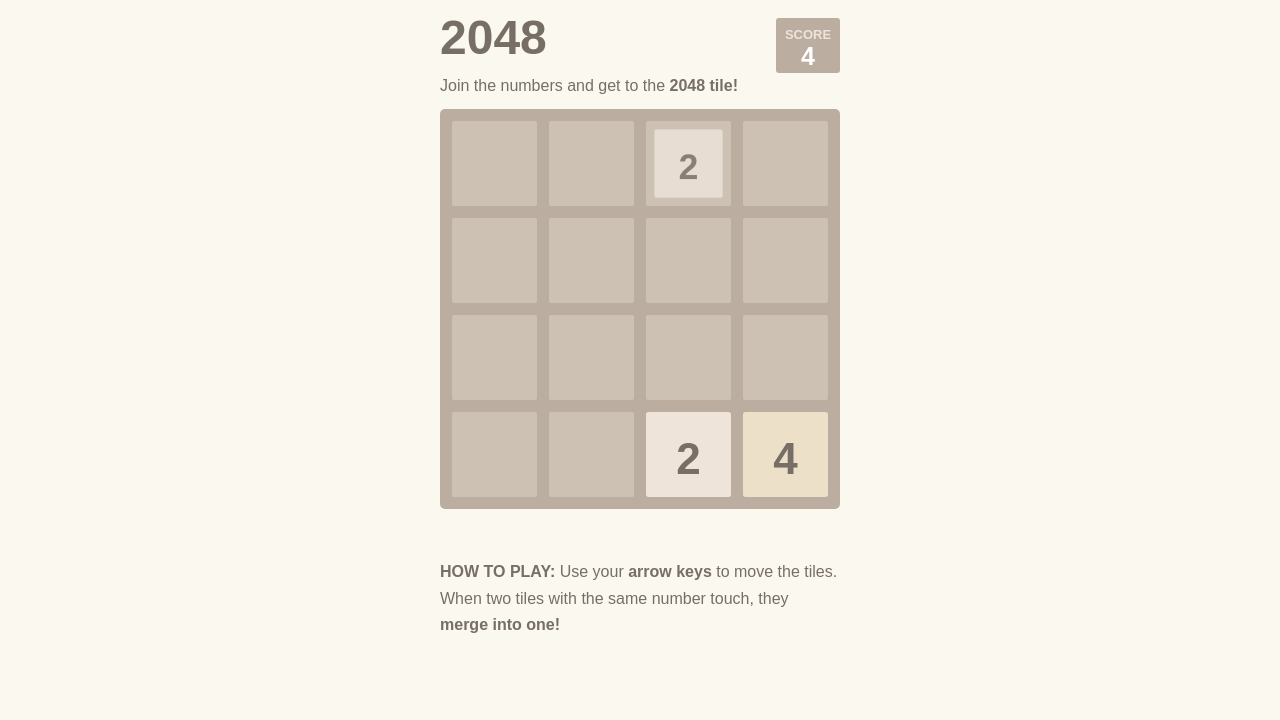}
\small (b) 2048
\end{minipage}
\caption{The support-schedule and 2048 applications with reference repairs
applied, illustrating the reference solutions.}
\label{fig:examples}
\end{figure}

\section{Development examples}\label{app:trajectories}
The first example is a code-only trajectory from the
Game evaluation. The remaining development episodes and reference replays
were collected separately and illustrate how visual observations guide
diagnosis, code changes, and behavioral checks. The Web examples combine
a view of the bug, a reference patch, and an application replay; the Game
examples trace interactions and verifier feedback.

\paragraph{2048: checking restart behavior through code execution.}
\label{app:code-execution-example}
In its successful code-only attempt, Claude-Fable-5 loads the
keyboard, grid, tile, game-manager, and game-API modules into a Node test
harness. It substitutes event handlers and timers, simulates held keys and
restart inputs, and reads assertion output for seeded board restoration,
queued movement after restart, and ordinary held-key movement. This
execution checks program behavior without launching or observing the
browser application. The submitted repair passes the task verifier.

\paragraph{Allocation Ring Studio: sizing a chart within its container.}
The chart is clipped in the Stacked layout. The reference repair measures the
plot area instead of the enclosing card and removes an extra gutter deduction.
A replay with the same board and layout shows the ring fitting its card.
The verifier checks plot-relative radius and containment across four layout
profiles and three boards.

\paragraph{Threadloop: revealing updates without losing reading position.}
The reference repair chooses between revealing a new paragraph and preserving
the reader's position based on who added the text and where that collaborator
is reading. The replay shows the paragraph becoming visible. Tests check the
update, the reveal-or-preserve behavior, and use of the Quill runtime.

\paragraph{Captain Callisto: persistent corner contact.}
After a 900-ms \texttt{W+Shift} hold and a later 700-ms \texttt{D+W} hold, the
screen clock advances from 7.4 to 14.0 seconds while the astronaut remains at
the corner. The agent's patch resets the collision-contact mask each frame
and removes a condition that zeros horizontal velocity. The agent later
reports movement after reloading, but does not complete the route and the
full task fails.

\paragraph{Flappy Bird: flap strength and contact during motion.}
A short flap produces an impulse of 120 against a target of 220, while a
held flap reaches its target of 250. A contact drill also shows an overlap
without a registered collision. The agent changes the impulse calculation,
update ordering, and contact handling. After reloading, the short-flap drill
still shows 120 against the target of 220. The overall attempt fails.

\paragraph{Core Ball: a completed replay can miss remaining defects.}
The agent traces a queued pin's incorrect attachment to an angle predicted
before launch. It changes the impact-angle calculation and promotes queued
shots after the previous attachment updates the core's motion. Its custom
replay completes four attachments and a restart, and the build passes.
The independent tests still reject the patch: the impact is resolved at
the next simulation step rather than its exact arrival time, and a rejected
collision still adds a pin to the occupied positions. The verifier checks
both exact event timing and state preservation after rejected collisions.

\paragraph{Letter Arc: replay reveals an incomplete animation repair.}
A completed round initially shows the correct letters and colors. The agent
identifies a conflict between the finish animation and the tile's reveal
transform, then edits the CSS. Replaying the finish exposes a new problem:
the revealed colors disappear in the paused frame. Further edits separate
the animation layers and add styles preserving the revealed status during
celebration. The recorded interactions do not show a successful replay of
the final patch. Independently, the evaluator records a passing build and
24 passing browser checkpoints, covering visual continuity, repeated replay,
Continue behavior, restart, daily/practice modes, and reduced-motion completion.

\paragraph{Recipient selection: code checks and a stale interface.}
Interface guidance indicates that Enter should append the highlighted
recipient. The agent updates the selection policy and builds the application.
Direct transition checks confirm that Enter appends Iceland to Norway and
Sweden, while a keep operation preserves the existing selection. Browser
interactions still show cleared recipients, which the agent attributes to a
stale application bundle. Thus, the source-level checks pass while the
recorded browser replay still shows the failure.

\paragraph{Vector Relay: transferring a scheduled event.}
The agent launches an orb and observes a lost return: the relay stays armed,
the score remains zero, and the failed-return indicator turns red. It traces
the failure to a scheduled event whose owner changes when the orb is replaced.
The patch cancels the event on manual recall but transfers it on replacement
or receiver movement. Source-level checks confirm cancellation, settlement
after two transfers, and rejection of repeated contact. A subsequent browser
replay shows ``Circuit complete'' and a score of 500; the independent evaluator
also records a pass.

\paragraph{Hextris: continuing after a failed check.}
\begin{figure}[htbp]
\centering
\includegraphics[width=\linewidth]{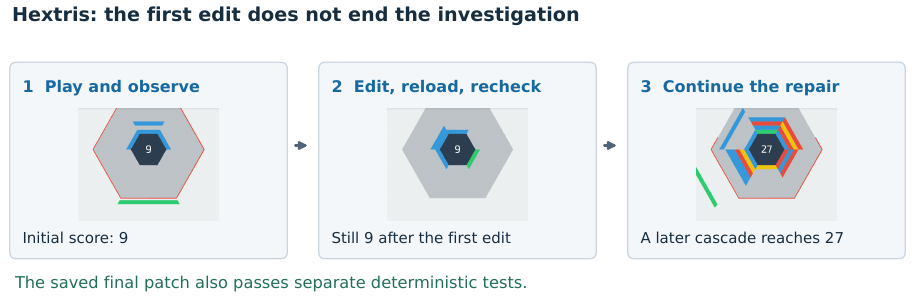}
\caption{\textbf{A replay can show that more work is needed.} Three frames
from one Hextris run show the initial score of nine, an unsuccessful recheck
after the first edit, and a later cascade reaching 27. The intervening work
includes further inspection, temporary diagnostics, and reloads. Screenshots
are cropped around the board; the task's implementation hint is above the
crop. The final source change is checked separately by the evaluator.}
\label{fig:iteration}
\end{figure}
The task provides a visible implementation hint. After observing a score of
nine, the agent queues blocks that settle again and delays each check until
blocks of that color stop moving. The first replay still shows nine. Further
inspection, temporary diagnostics, and reloads produce a cascade reaching 27.
The agent then removes the diagnostics and revises the queue handling. A
local test confirms that the score remains nine while a same-color block
moves and reaches 27 after settlement. The final patch preserves the
three-block match threshold. The independent evaluator records a pass,
including two negative controls; its separate test fixture ends at 41.
Figure~\ref{fig:iteration} illustrates several rounds of revision and checking.

\section{Complete task list}\label{app:inventory}
Table~\ref{tab:four-domain-inventory} lists the 36 Web,
29 Game, 20 DevOps, and 20 Mobile tasks, with information-requirement
labels and their specification basis. M denotes application-material
dependent and S denotes source-specified, as defined in
Appendix~\ref{sec:runtime-information}.
\input{tables/inventory-four-domain}

%% file: tables/inventory-four-domain.tex
\begingroup\footnotesize
\setlength{\tabcolsep}{3pt}\setlength{\LTcapwidth}{\linewidth}
\begin{longtable}{@{}p{0.09\linewidth}>{\raggedright\arraybackslash}p{0.29\linewidth}c>{\raggedright\arraybackslash}p{0.49\linewidth}@{}}
\caption{Task information requirements. M denotes application-material dependent; S denotes source-specified (Appendix~\ref{sec:runtime-information}). Each row gives the specification basis. Names recur across distinct scenarios; tasks follow the released manifest order within each domain.}\label{tab:four-domain-inventory}\\
\toprule Domain & Task & Type & Specification basis \\ \midrule\endfirsthead
\toprule Domain & Task & Type & Specification basis \\ \midrule\endhead
\bottomrule\endfoot
Web & MapLibre reset touch reacquisition & S & The instruction specifies pointer attachment and preservation of navigation. \\
Web & Fabric opposite-edge resize & S & Readable Help in index.html specifies the resize and selection contract. \\
Web & ECharts linked data-zoom toolbar & S & The instruction specifies corresponding chart ranges; source exposes range-mapping strategies. \\
Web & Fabric viewport rotation controls & S & The instruction specifies frame, handle, and zoom agreement under camera transforms. \\
Web & JupyterLab Shift-click cell selection & S & The instruction distinguishes cell-range selection from extension of text selection. \\
Web & Fabric through-edge resize with nested camera & S & Readable Help in index.html specifies the resize and selection contract. \\
Web & Quill paste and scroll round trip & M & The collaboration bar supplies the presence-dependent meaning of the update. \\
Web & DatePicker mask, clear and blur & S & The instruction specifies cleared versus invalid-date behavior. \\
Web & Fabric through-edge resize with nested camera & S & Readable Help in index.html specifies the resize and selection contract. \\
Web & Recharts responsive pie layout & M & The approved reference image specifies how the ring fits the board. \\
Web & Slate soft break and scroll & S & The instruction specifies caret visibility while retaining the preceding lines. \\
Web & MapLibre mouse ridge attachment & S & The instruction specifies pointer attachment and preservation of navigation. \\
Web & Tiptap keyboard focus position & M & The workspace note and rendered draft distinguish the current bookmark from stale anchors. \\
Web & MapLibre reset terrain reacquisition & S & The instruction specifies pointer attachment and preservation of navigation. \\
Web & Fabric Single-to-Nested return & S & Readable Help in index.html specifies the resize and selection contract. \\
Web & Account recovery contract & M & The live identity policy specifies the applicable recovery channel. \\
Web & MapLibre reset integration handoff & S & The instruction specifies pointer attachment and preservation of navigation. \\
Web & Delegated action contract & M & The live authorization policy specifies the delegated action. \\
Web & Handsontable menu scroll anchor & S & The instruction specifies menu anchoring and closure on anchor disappearance. \\
Web & MapLibre terrain pinch anchor & S & The instruction specifies the geographic point to preserve during pinch and pan. \\
Web & ComboBox selection & M & The workspace keyboard guide specifies Enter, Escape, and Tab behavior. \\
Web & ComboBox selection & M & The workspace keyboard guide specifies Enter, Escape, and Tab behavior. \\
Web & Responsive masonry overlap & S & The instruction specifies persistent section visibility and non-overlap on resize. \\
Web & Saleor export of current search & M & The workspace export guide specifies request fields and filter encoding. \\
Web & MapLibre post-wheel reacquisition & S & The instruction specifies pointer attachment and preservation of navigation. \\
Web & MapLibre reset wheel reacquisition & S & The instruction specifies pointer attachment and preservation of navigation. \\
Web & Subscription renewal contract & M & The live lifecycle contract specifies the renewal interpretation. \\
Web & Support window contract & M & The service contract specifies the selected support window. \\
Web & Tax rounding contract & M & The live calculation contract specifies the tax convention. \\
Web & Fabric through-edge resize & S & Readable Help in index.html specifies the resize and selection contract. \\
Web & Vue sticky table columns & M & The table preferences supply the active density geometry token and value. \\
Web & Warehouse cutoff & M & The live dispatch contract specifies cutoff and shipping-day rules. \\
Web & Warehouse cutoff & M & The live dispatch contract specifies cutoff and shipping-day rules. \\
Web & Nested Fabric selection & S & Readable Help in index.html specifies the resize and selection contract. \\
Web & Nested Fabric selection & S & Readable Help in index.html specifies the resize and selection contract. \\
Web & Nested Fabric selection & S & Readable Help in index.html specifies the resize and selection contract. \\
Game & Captain Callisto: jetpack timing & S & The instruction and jetpack implementation define burst growth, release, and restart consistency. \\
Game & Captain Callisto: jetpack timing & S & The instruction and jetpack implementation define burst growth, release, and restart consistency. \\
Game & Captain Callisto: level-completion transition & S & The instruction requires completion to persist after the approach and release sequence. \\
Game & Captain Callisto: cargo collection & M & The live cargo scanner supplies the route-to-credit interpretation. \\
Game & Captain Callisto: platform contact & S & The instruction specifies non-clipping contact and continued passage at the deck corner. \\
Game & Captain Callisto: lift-transition physics & S & The instruction specifies directional continuity across the moving-support handoff. \\
Game & Captain Callisto: level-completion transition & S & The instruction requires completion to persist after the approach and release sequence. \\
Game & 2048: seeded restart consistency & S & The instruction specifies deterministic restart without a queued post-restart move. \\
Game & 2048: seeded restart consistency & S & The instruction specifies deterministic restart without a queued post-restart move. \\
Game & Astray: glass collisions and parallax & S & The instruction specifies the full color sequence and continued movement. \\
Game & Core Ball: pin attachment and collision & S & The instruction specifies agreement between pin approach and attachment. \\
Game & Hextris: stable cascade settlement & S & The instruction specifies cascade completion under the existing match-size rule. \\
Game & Hextris: cascade settlement & S & The instruction specifies cascade completion under the existing match-size rule. \\
Game & Pac-Man: warning-state handoff & S & The instruction specifies agreement among warning, contact, and resumed route. \\
Game & Minesweeper: delayed-loss reset & M & The Episode integrity panel supplies the required delayed-loss timing. \\
Game & Signal Weave: checkpoint and hazard synchronization & M & The route card supplies the lane handoff orientation. \\
Game & Vector Relay: deferred-event ownership & S & The instruction specifies one delayed settlement after natural replacement. \\
Game & Vector Relay: pulse-balance handoff & S & The instruction distinguishes manual recall from natural replacement. \\
Game & Letter Arc: finish animation and replay & S & The instruction specifies an intact completed row throughout finish and replay. \\
Game & Boxel Rebound: restart and camera continuity & M & The application image card specifies restart phase and camera reconstruction. \\
Game & Boxel Rebound: restart and camera continuity & M & The application image card specifies restart phase and camera reconstruction. \\
Game & Boxel Rebound: restart and camera continuity & M & The application image card specifies restart phase and camera reconstruction. \\
Game & Boxel Rebound: restart and camera continuity & M & The application image card specifies restart phase and camera reconstruction. \\
Game & Boxel Rebound: restart and camera continuity & M & The application image card specifies restart phase and camera reconstruction. \\
Game & Flappy Bird: flight timing and contact & S & The source contains the target impulses and arcs used by the Flight Lab. \\
Game & Flappy Bird: flight timing and contact & S & The source contains the target impulses and arcs used by the Flight Lab. \\
Game & Flappy Bird: flight timing and contact & S & The source contains the target impulses and arcs used by the Flight Lab. \\
Game & Flappy Bird: flight timing and contact & S & The source contains the target impulses and arcs used by the Flight Lab. \\
Game & Minesweeper: delayed-loss reset & M & The Episode integrity panel supplies the revised delayed-loss timing. \\
DevOps & Async trace context propagation & M & The compatibility catalog supplies the manager-specific continuation contract. \\
DevOps & Cross-host trace clock correction & M & Operator notes and calibration guidance specify clock authority and correction scope. \\
DevOps & Alert inhibition identity & M & The deployed scope registry and operator guidance specify suppression eligibility. \\
DevOps & Rate-before-aggregation semantics & M & The deployed query definitions and publication lineage identify the correct input binding. \\
DevOps & SLO availability rollups & M & The reporting agreement specifies the required regional weighting and rollup. \\
DevOps & Log severity normalization & M & Receiver records and fleet conventions specify the severity interpretation. \\
DevOps & Tail-sampling decision windows & M & The receiver policy or runbook supplies the contractual decision-time band. \\
DevOps & Alert pending identity and routing & M & The receiver selector and routing policy specify the stable alert identity. \\
DevOps & StatsD gauge update modes & M & The endpoint provisioning or relay announce contract specifies update semantics. \\
DevOps & Rate-before-aggregation semantics & M & The deployed query definitions and publication lineage identify the correct input binding. \\
DevOps & Tail-sampling decision windows & M & The receiver policy or runbook supplies the contractual decision-time band. \\
DevOps & StatsD gauge update modes & M & The endpoint provisioning or relay announce contract specifies update semantics. \\
DevOps & Rate-before-aggregation semantics & M & The deployed query definitions and publication lineage identify the correct input binding. \\
DevOps & Async trace context propagation & M & The compatibility catalog supplies the manager-specific continuation contract. \\
DevOps & SLO availability rollups & M & The reporting agreement specifies the required regional weighting and rollup. \\
DevOps & Alert inhibition identity & M & The deployed scope registry and operator guidance specify suppression eligibility. \\
DevOps & Log severity normalization & M & Receiver records and fleet conventions specify the severity interpretation. \\
DevOps & Cross-host trace clock correction & M & Operator notes and calibration guidance specify clock authority and correction scope. \\
DevOps & Alert pending identity and routing & M & The receiver selector and routing policy specify the stable alert identity. \\
DevOps & Scrape response phase deadlines & M & The policy activation board and authority matrix supply the freshness and abort bounds. \\
Mobile & Brine \& Beam: received wall-chart schedules & M & Received drawings specify job durations, dependencies, and report requirements. \\
Mobile & Brine \& Beam: received wall-chart schedules & M & Received drawings specify job durations, dependencies, and report requirements. \\
Mobile & Brine \& Beam: received wall-chart schedules & M & Received drawings specify job durations, dependencies, and report requirements. \\
Mobile & Brine \& Beam: received wall-chart schedules & M & Received drawings specify job durations, dependencies, and report requirements. \\
Mobile & Brine \& Beam: received wall-chart schedules & M & Received drawings specify job durations, dependencies, and report requirements. \\
Mobile & Brine \& Beam: received wall-chart schedules & M & Received drawings specify job durations, dependencies, and report requirements. \\
Mobile & Brine \& Beam: received wall-chart schedules & M & Received drawings specify job durations, dependencies, and report requirements. \\
Mobile & Brine \& Beam: received wall-chart schedules & M & Received drawings specify job durations, dependencies, and report requirements. \\
Mobile & Brine \& Beam: received wall-chart schedules & M & Received drawings specify job durations, dependencies, and report requirements. \\
Mobile & Repair imported inspection measurements & M & Imported calibration and reports specify annotation and measurement interpretation. \\
Mobile & Marble Post: imported runs no longer agree with office packets & M & Received track packets specify the simulation and edited-plan receipt contract. \\
Mobile & Repair Tideglass's received-plan preview & M & Received assembly drawings specify gear connections and motion relationships. \\
Mobile & Repair Mural Desk's imported projection workflow & M & Imported projection-office packets specify mapping and inverse-edit semantics. \\
Mobile & Northline tour review: hotspots drift from imported panoramas & M & Imported tour reports specify panorama registration and hotspot interpretation. \\
Mobile & Repair the received-plate integration & M & Received finder/plate pairs specify field registrations and target offsets. \\
Mobile & Repair Foldnote proof review & M & Imported artwork and panel reports specify placement and orientation. \\
Mobile & Repair Lantern's delivered show integration & M & Delivered patchboard and timing sheets specify signal and memory behavior. \\
Mobile & Repair the Lamp courier pickup review & M & Delivered pose and exposure boards specify the approved performance. \\
Mobile & Repair the pass wallet & M & Issuer records and receipts specify availability and validation semantics. \\
Mobile & Repair Campus Pocket's imported-campus routing & M & Imported maps and routing reports specify permitted trips and route interpretation. \\
\end{longtable}
\endgroup

%% file: sections_preprint/09-appendix-rl.tex
\section{Post-training implementation and RLVR extension}\label{app:rl}
\subsection{SFT data construction across domains}\label{app:sft-data}
\paragraph{Verified reference replays.}
We use the same SFT data construction pipeline in each domain, instantiated
with that domain's application environments, reference
repairs, and native verifiers. For each task, a fixed reference
policy executes the gold repair in a fresh environment: it starts the
browser and receives a screenshot, applies the gold patch through a shell
action, runs the reference validation command, waits 750\,ms through the
computer tool to obtain an updated screenshot, and finishes the episode.
The task's native verifier then checks the resulting repair. A complete
replay supplies training data only after passing verification. These scripted
replays provide supervision for repair execution and its surrounding tool
interactions, using screenshots returned by the running application.

\paragraph{Action prefixes and sampling weights.}
Each accepted replay yields five cumulative conversation prefixes, one for
each action. A prefix contains the task instruction and the messages and
observations available before its target action, followed by that action
as the supervised response. This preserves the information available at
each decision point. We oversample these prefixes with the fixed weights
in Table~\ref{tab:sft-prefix-weights}, emphasizing the patch-application
action. Thus, $N$ accepted replays produce $5N$ distinct task/action prefix
positions and $44N$ weighted training rows. Repeated rows change the sampling
weight of a prefix; the underlying task and demonstration remain the same.

\begin{table}[htbp]
\centering\small
\caption{\textbf{SFT prefix construction for each domain.}
Each row supervises the indicated action, conditioned on the preceding
conversation. Screenshot counts refer to images available before that action.}
\label{tab:sft-prefix-weights}
\begin{tabular}{@{}lcc@{}}
\toprule
Supervised action & Copies per task & Screenshots in context\\
\midrule
Start browser & 2 & 0\\
Apply gold patch & 32 & 1\\
Validate repair & 4 & 1\\
Wait and observe & 2 & 1\\
Finish episode & 4 & 2\\
\midrule
Total & 44 & ---\\
\bottomrule
\end{tabular}
\end{table}

\paragraph{Multimodal representation and target masking.}
Each example stores the conversation, references to its screenshot inputs,
and metadata linking it to the task version, reference replay, and prefix
position. The loader applies the supervised loss only to the final assistant
action. Earlier assistant actions, task instructions, tool outputs, and
screenshots provide conditioning context and receive no prediction loss.
The browser-start prefix is text-only because its target action precedes the
first screenshot; the finish prefix conditions on both observations. Before
training, token-level checks verify the target mask, context length, and
action-token limit for every weighted example.


\subsection{Supervised fine-tuning}\label{app:sft}
\paragraph{Training configuration.}
The SFT experiment in Section~\ref{sec:post-training} uses Qwen3.8-27B,
the weighted action-prefix examples described in
Appendix~\ref{app:sft-data}, and 1,801 optimizer updates. Training starts
from the base checkpoint with shuffled examples, per-token supervised loss,
and global and rollout batch sizes of four. The vision encoder and
linear-attention parameters are frozen. The learning rate decreases from
$10^{-6}$ to $2\times10^{-7}$ with zero warm-up, and the token budget is
16,384 per GPU. The update budget corresponds to 7,204 nominal sample slots,
approximately 4.96 passes over the weighted dataset.
Figure~\ref{fig:sft-training} includes every recorded batch loss. Its smooth
curve is the trailing arithmetic mean over at most 50 updates, rather than
an epoch-level average.

\paragraph{Evaluation.}
Both checkpoints are evaluated on the same set of tasks, with one
greedy attempt per task and matched interaction settings. The outcome is
the native task-verifier result $P_i$: 11.1\% for the base model
and 33.3\% for the SFT checkpoint. The task-success metric used in the benchmark
comparison also includes
trajectory compliance, $S_i=P_iU_i$; these SFT results report $P_i$.

\subsection{Verifiable rewards and the trajectory-level GRPO objective}\label{app:rl-objective}
\paragraph{Verifiable rewards.}
For trajectory $i$ ending with codebase $c_{T_i}$, the terminal repair
reward is $R_i=P_i=V_u(c_{T_i})$. It is determined
by the verifier, which checks the requested behavior, designated existing
functionality, and restrictions on permitted changes.
Scoring starts from a clean baseline, transfers permitted edits, rebuilds the
application, and runs the task's tests. Changes to protected evaluator or
build files are rejected; candidate dependencies and generated build outputs
are not used for scoring. A compilation failure or failed requirement earns
zero. A successful repair supplies supervision for the entire sequence of coding and computer
use that produced it. Test progress is recorded for analysis; its optional
use in reward shaping is disabled in the default configuration described
below.

\paragraph{Group-relative learning.}
The RLVR extension adapts GRPO~\citep{shao2024deepseekmath} to complete development episodes.
For each task, the behavior policy samples $G$ trajectories from fresh
workspaces. For a group with valid scoring outcomes, define
\begin{equation}
 \bar R=\frac{1}{G}\sum_{i=1}^{G}R_i,\qquad
 s_R=\sqrt{\frac{1}{G-1}\sum_{i=1}^{G}(R_i-\bar R)^2},\qquad
 A_i=\frac{R_i-\bar R}{s_R+\delta}.
 \label{eq:rl-advantage}
\end{equation}
Here $\delta>0$ is a numerical stability constant.
Every generated token in trajectory $i$ receives the same advantage $A_i$.
Groups in which all repairs succeed or all fail are retained with zero
advantage.

An interleaved trajectory contains both model outputs and environment
observations. Let $m_{i,k}=1$ at generated-token positions and zero elsewhere,
$L_i=\sum_k m_{i,k}$, and $H_{i,k}$ denote the multimodal prefix preceding
token $z_{i,k}$. With
$\rho_{i,k}(\theta)=\pi_\theta(z_{i,k}\mid H_{i,k})/
\pi_{\mathrm{old}}(z_{i,k}\mid H_{i,k})$,
the policy objective is
\begin{equation}
 \begin{aligned}
 J(\theta)&=\mathbb{E}_{u,\,\tau_{1:G}\mid F=1}\!\left[
 \frac{1}{G}\sum_{i=1}^{G}\frac{1}{L_i}\sum_k m_{i,k}\ell_{i,k}(\theta)\right],\\
 \ell_{i,k}(\theta)&=\min\!\left(\rho_{i,k}A_i,
 \operatorname{clip}(\rho_{i,k},1-\epsilon_{\mathrm{low}},
 1+\epsilon_{\mathrm{high}})A_i\right).
 \end{aligned}
 \label{eq:rl-objective}
\end{equation}
The hyperparameters $\epsilon_{\mathrm{low}}$ and $\epsilon_{\mathrm{high}}$
control the lower and upper clipping bounds on the probability ratio.
Here $F=1$ means that all trajectories in the group have usable scoring
outcomes. The old-policy probabilities are computed by the training backend.
Both KL regularization and entropy bonuses are
disabled. The mask excludes observations from the prediction loss while
retaining them as conditioning context. Per-trajectory normalization by
$L_i$ prevents longer responses from receiving more weight solely because
they contain more generated tokens.

This adaptation changes the unit of sampling from a standalone answer to a
development episode, replaces answer correctness with executable repair
verification, and restricts the loss to generated tokens across all turns.
We implement the objective using SLIME~\citep{slime2026}, coupling its
policy updates to the coding and computer-use environment.
Worker failures require separate treatment: a failed attempt is retried from
a fresh workspace, and if retries are exhausted, all trajectories sampled
for that task prompt are
excluded and replaced before normalization. Such failures have no model
reward. Evaluation reports a failure status after retry exhaustion.

\paragraph{Training procedure.}
Each iteration samples a batch of task prompts, executes $G$ development episodes
per prompt, computes verifier rewards and relative advantages, and updates
the policy. Updated weights are synchronized to the rollout engine for the
next iteration. Appendix~\ref{app:rl-implementation} describes the rollout and update
interface. This procedure defines the RLVR extension; the completed
post-training experiment in Section~\ref{sec:post-training} uses SFT.

\subsection{Rollout and update interface}\label{app:rl-implementation}
\begin{figure}[htbp]
\centering
\includegraphics[width=\linewidth]{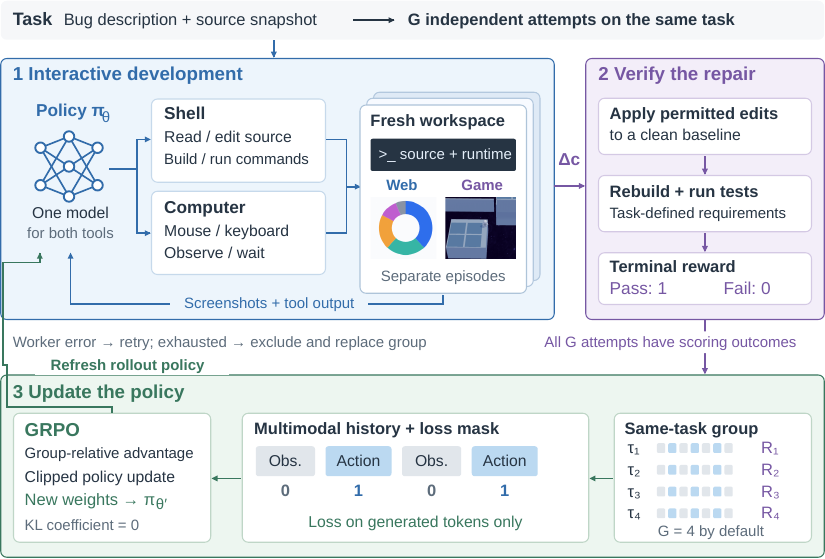}
\caption{\textbf{RLVR extension using verifiable software repairs.}
For the same task, the policy samples several development trajectories in
separate workspaces. Shell commands and computer actions return text and
screenshots. At termination, permitted source changes are transferred to a
clean scoring copy and checked by deterministic tests. Complete, successfully
scored groups provide relative advantages for a policy update; observations
condition the policy but receive no direct prediction loss. Application
thumbnails show example Web and Game applications.}
\label{fig:rl-pipeline}
\end{figure}

\paragraph{Parallel development episodes.}
The implementation connects a persistent browser and editable workspace to
SLIME's multimodal rollout interface. Each attempt receives fresh files and
application ports, allowing independent development episodes to run in
parallel. The policy's generated tokens, their sampling log probabilities,
image inputs, and observation masks are preserved in one interleaved sample.
The policy conditions on the sequence of observations and actions up to its
context limit. SGLang supplies inference and the Megatron backend supplies
policy updates, with weights synchronized through SLIME's training loop.
Figure~\ref{fig:rl-pipeline} illustrates this interface.

\paragraph{Verifiable feedback.}
The scoring copy rejects protected-file and symlink edits and retains
baseline dependencies and build configuration. Baseline fingerprints detect
mutation outside the candidate workspace. Reset, browser, inference, and
verifier-runtime failures are distinguished from unsuccessful repairs;
unresolved infrastructure failures are excluded before computing group
advantages. The default reward is binary repair correctness, with no
intermediate rewards. Generated tokens share the episode-level advantage,
while environment observations condition the policy without receiving
prediction loss.

\paragraph{Scaling RLVR.}
Independent rollouts and executable scoring permit additional workers to
collect verifiable experience without manually labeling each attempted
repair. Scaling this process to long hybrid-CUA trajectories requires
balancing rollout throughput, optimization stability, and credit assignment
across coding and interaction. Scalable RLVR is immediate future work;
the completed experiment reported here is supervised fine-tuning.

%% file: sections_preprint/10-appendix-results.tex
\section{Repeated-attempt evaluation}\label{app:repeated-attempts}
\begingroup
Table~\ref{tab:hybrid-pass3} measures the coverage gained
from two additional attempts relative to the first-attempt results in
Table~\ref{tab:four-domain-results}. Each attempt starts from a fresh workspace
under the domain's fixed task, access, and verification settings.
\paragraph{Additional attempts expand task coverage.}
GPT-6-Astra reaches pass@3 of 75.0\% on Web, 62.1\% on
Game, 90.0\% on DevOps, and 60.0\% on Mobile. On Game, its cumulative
pass@1, pass@2, and pass@3 are 37.9\%, 58.6\%, and 62.1\%, respectively:
11 tasks are solved in the first attempt, 17 in at least one of the first
two attempts, and 18 in at least one of all three attempts.
Appendix~\ref{app:cohort-reconciliation} describes the repeated evaluation's
cohort composition.

\paragraph{Stable competence and additional-attempt coverage.}
Repeated evaluation distinguishes tasks an agent completes consistently
from those reached through another development trajectory. In the
three-attempt Game study, GPT-6-Astra solves 34.5\% of tasks in all three
attempts, compared with 3.4\% for GPT-5.6-Sol. Ten of its 18 solved
tasks succeed in every attempt; the remaining eight succeed in one
or two attempts. On Mobile, an additional attempt lets GPT-5.6-Sol complete
the sprite task and GPT-6-Astra complete one schedule variant; the remaining
schedule variants continue to distinguish this family from the broadly
solved application tasks. Thus additional attempts expand coverage while
the task-level success patterns identify capabilities requiring more reliable
execution. Appendix~\ref{app:repeated-attempts} reports all rates and
uncertainty estimates within the repeated evaluation.

\paragraph{Task coverage and sampling uncertainty.}
We assess task-sampling uncertainty on
Web and Game with 20,000 paired bootstrap resamples, preserving each task's
outcomes across models and attempts. On Game, GPT-6-Astra's pass@3
advantage over GPT-5.6-Sol is 31.0 points, with a 95\% interval of
[13.8, 48.3]. Resampling whole game families gives an interval of
[11.5, 50.0], supporting the same ordering under variation in the
composition of source games. Appendix~\ref{app:repeated-attempts} gives
both model comparisons, the pass@$k$ profiles, and the resampling methods.
\par\endgroup

\input{tables/hybrid-pass3}
\subsection{Metrics and aggregation}
We evaluate GPT-6-Astra, GPT-5.6-Sol, and Claude-Fable-5 with three attempts per
task across Web, Game, DevOps, and Mobile. For task $i$, let
$s_{ij}\in\{0,1\}$ denote success in attempt $j$, in the recorded execution
order. For a domain with $N$ tasks, cumulative success is
$\mathrm{pass@}k=100N^{-1}\sum_i\mathbb{1}[\sum_{j=1}^{k}s_{ij}>0]$,
for $k\in\{1,2,3\}$. A task remains covered once any of its first $k$
attempts succeeds. Pass@1 therefore matches the single-attempt comparison,
and pass@3 counts tasks solved at least once across all three attempts. Table~\ref{tab:repeated-passk} reports
this cumulative profile for Web and Game, where ordered task-level outcomes
are available. Appendix~\ref{app:cohort-reconciliation} describes the
execution protocol and attempt composition.
\label{eq:passk}
Four-domain means average the unrounded domain rates equally.
Web and Game statistics use the recorded ordered task-level outcomes.
DevOps and Mobile pass@1 and pass@3 use the recorded aggregate
outcomes. We compute task-level bootstrap intervals for Web
and Game.

The task-level success profiles and success-frequency distributions below
describe Web and Game. The frequency distributions give the percentage
of tasks solved in zero, one, two, or all three attempts, distinguishing
consistently solved tasks from those reached through additional attempts.
\input{tables/repeated-passk}
\input{tables/repeated-success-frequency}

\clearpage
\subsection{Task-sampling uncertainty}
For the paired model comparisons in Table~\ref{tab:repeated-paired-differences},
we also report the mean success rate across all three attempts,
$\overline{S}=100(3N)^{-1}\sum_i\sum_{j=1}^{3}s_{ij}$.
For statistical intervals on Web and Game, we resample tasks
within each domain, drawing the domain's
original number of tasks with replacement. Each sampled task carries
the outcomes of all three models and all three attempts, preserving the
pairing in model comparisons. We recompute the three-attempt mean $\overline{S}$, cumulative pass@3, and the paired model
differences in each of 20,000 bootstrap replicates, and use the 2.5th and
97.5th percentiles as the 95\% interval. These intervals quantify variation
under resampling of the evaluated tasks.

We also perform a family-level sensitivity analysis, using nine
taxonomy-defined task families on Web and twelve source games on Game.
Each replicate resamples the domain's original number of groups and
includes every task from each sampled group. Rates retain task
weighting, with the total number of sampled tasks providing the replicate's
normalization. All model and attempt outcomes remain paired. This analysis
preserves dependence among related tasks while measuring sensitivity to
the benchmark's composition of task families and source games.

\input{tables/repeated-paired-differences}

\clearpage
\section{Domain-level CUA performance}\label{app:domain-performance}
\begin{figure}[H]
\centering
\includegraphics[width=\linewidth]{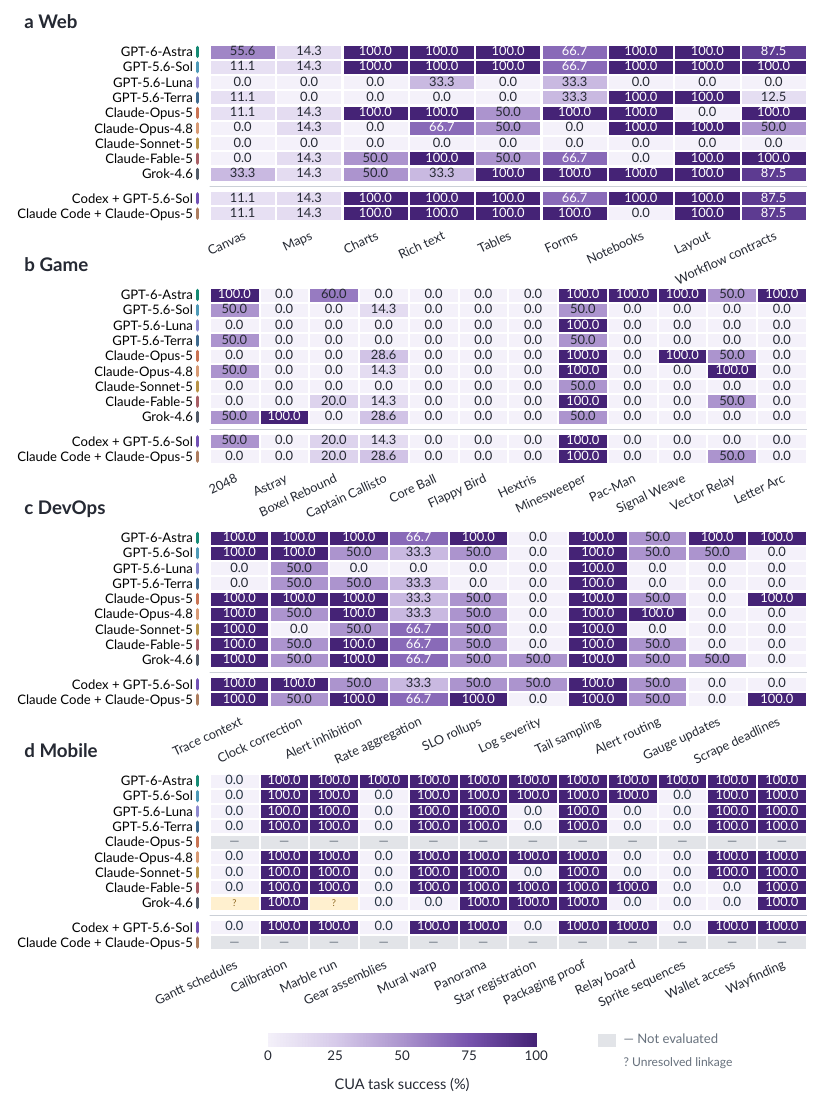}
\caption{\textbf{CUA task success across task families.} Cells report
task success rates within each family, with purple intensity encoding the
percentage. Web families group applications and tools; Game families
retain source-game identity; DevOps and Mobile follow task-contract
categories. Rows below the separator show agent systems. Rates use each
family's full task inventory. Em dashes denote unevaluated
models. Question marks denote Grok-4.6's two Mobile family estimates with
unresolved task outcomes: 0--11.1\% for Gantt schedules and 0--100\% for
Marble run (Appendix~\ref{app:cohort-reconciliation}).}
\label{fig:task-families}
\end{figure}

\begingroup
We summarize CUA performance across Web, Game, DevOps, and Mobile,
giving each domain equal weight. GPT-6-Astra
achieves a mean task success rate of 59.9\%, followed by GPT-5.6-Sol at 42.2\%,
Grok-4.6 and Codex with GPT-5.6-Sol at 41.9\%, and Claude-Fable-5 at 39.9\%.
Table~\ref{tab:four-domain-results} provides the complete domain-level results.

\paragraph{Domain and family strengths.}
GPT-6-Astra achieves the highest observed CUA task success rate in each domain:
66.7\% on Web, 37.9\% on Game, 80.0\% on DevOps, and 55.0\% on Mobile
(Appendix~\ref{app:domain-performance}). Its Web advantage over GPT-5.6-Sol is
concentrated in Canvas interactions after geometry changes, while its
Mobile results distinguish precise gear relationships and sprite pose
assignments (Figure~\ref{fig:task-families}). Across Game, the union of
successful CUA repairs covers 58.6\% of tasks.

\paragraph{Complementary strengths among frontier agents.}
Individual tasks reveal capabilities that the aggregate ranking compresses.
Claude-Opus-4.8 uniquely solves a DevOps alert-routing task by preserving
the labels consumed by receiver selectors. Grok-4.6 and Codex with
GPT-5.6-Sol solve a log-normalization task by combining transport format
with the original producer's severity convention. GPT-6-Astra misses both
tasks. These contrasts expose distinct demands on identifying which
application facts determine a correct repair.

\paragraph{Complete agent systems.}
Codex with GPT-5.6-Sol reaches 55.6\% on Web, 17.2\% on Game, 55.0\% on DevOps,
and 40.0\% on Mobile. Claude Code with Claude-Opus-5 reaches 55.6\%, 20.7\%,
and 65.0\% on Web, Game, and DevOps. Relative to the corresponding frontier models, Codex's Game rate is 6.9 percentage points higher and Claude
Code's DevOps rate is 5.0 points higher. These comparisons characterize
the combination of model, tools, and execution strategy.
\par\endgroup

\input{tables/four-domain-results}
\clearpage
\begin{figure}[H]
\centering
\includegraphics[width=\linewidth]{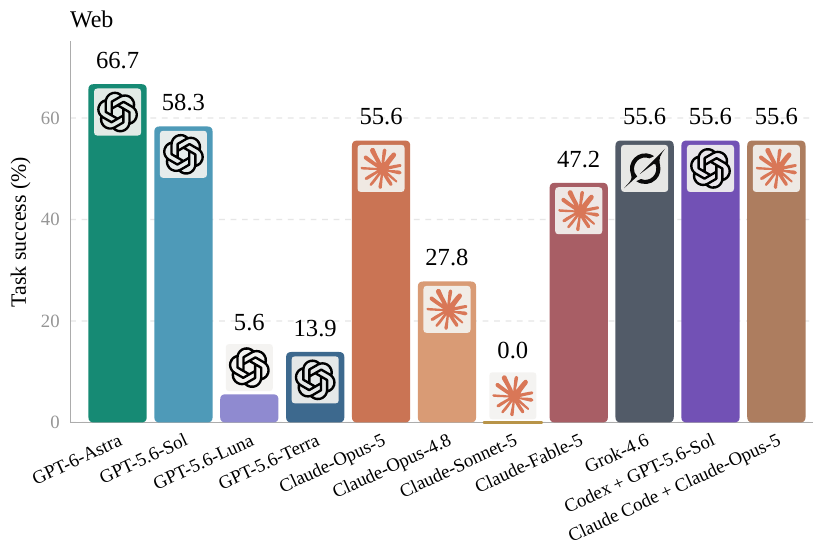}
\par\vspace{3pt}
\includegraphics[width=\linewidth]{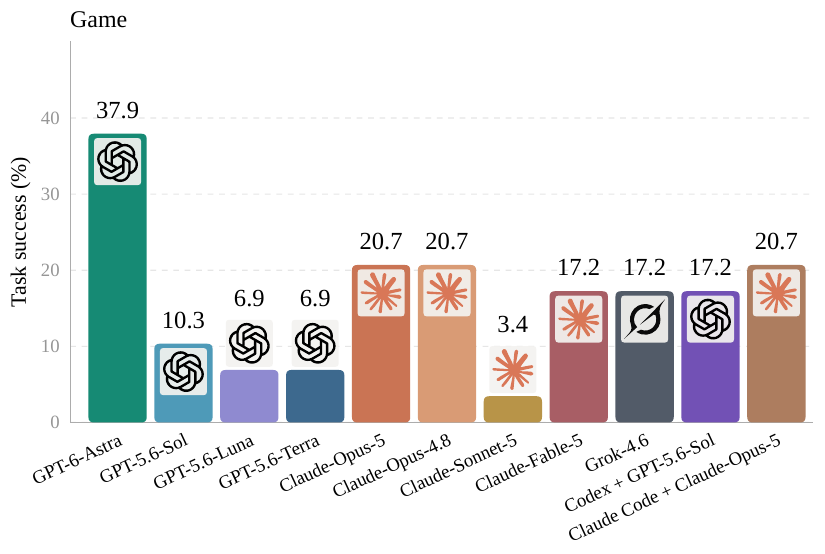}
\caption{\textbf{CUA task success on Web (top) and Game (bottom).}
Bars report success rates for nine frontier models and two agent systems over
the full task inventories: 36 tasks on Web and 29 on Game.}
\label{fig:domain-web}
\label{fig:domain-game}
\end{figure}

\clearpage
\begin{figure}[H]
\centering
\includegraphics[width=\linewidth]{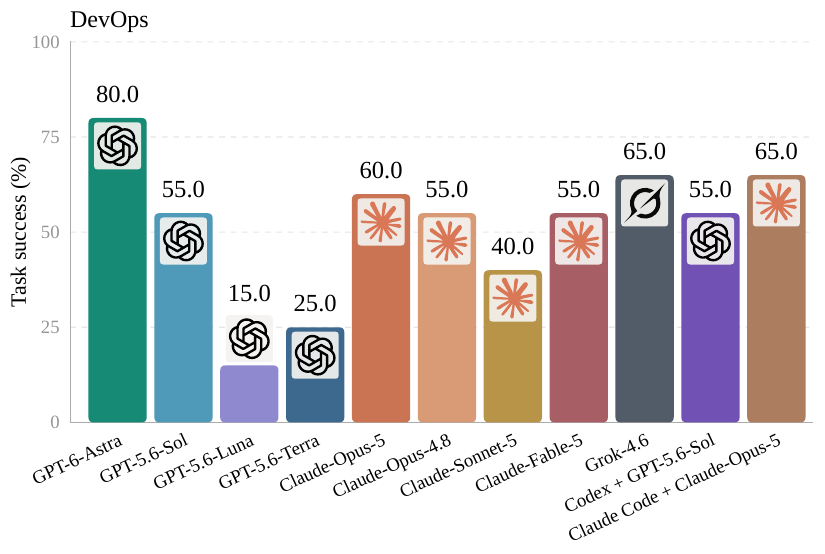}
\par\vspace{3pt}
\includegraphics[width=\linewidth]{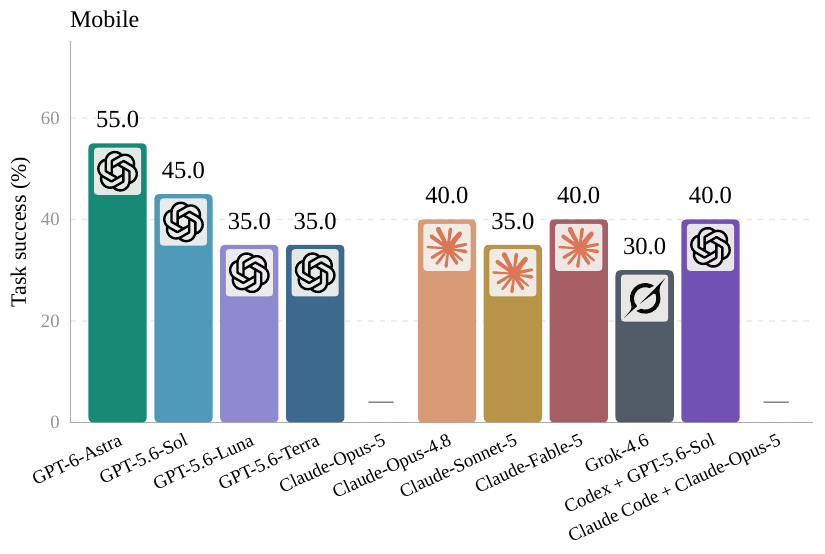}
\caption{\textbf{CUA task success on DevOps (top) and Mobile (bottom).}
Bars report CUA task success rates for the evaluated models and agent
systems. Colors follow Figure~\ref{fig:domain-web}.}
\label{fig:domain-devops}
\label{fig:domain-mobile}
\end{figure}

\clearpage
\begin{figure}[H]
\centering
\includegraphics[width=\linewidth]{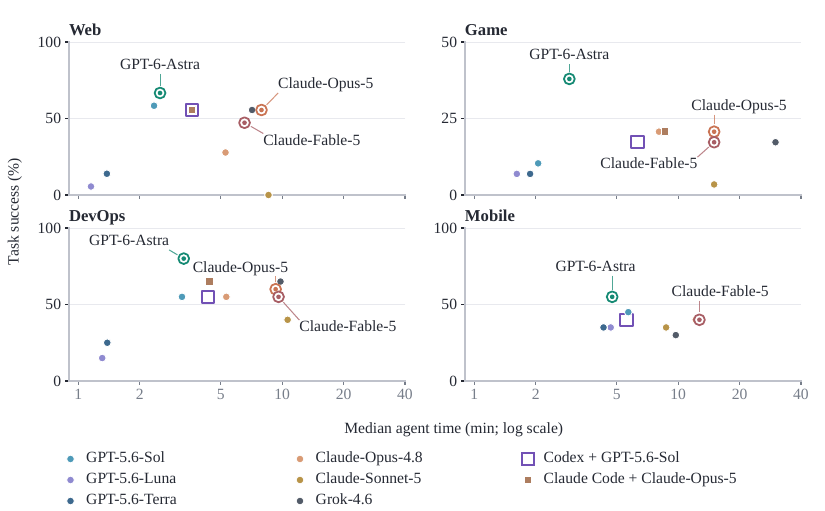}
\caption{\textbf{Task success and median development time within each domain.}
Points pair task success (vertical axis) with median
agent time (horizontal axis, logarithmic). Resource coverage and
aggregation are specified in Appendix~\ref{app:workload-accounting}. Colored rings and direct labels highlight GPT-6-Astra,
Claude-Opus-5, and Claude-Fable-5; the shared legend names the other entries.
Circles denote frontier models; open and filled squares denote Codex and Claude
Code, respectively, with nested squares for the nearly coincident Web
estimates. Success axes span 0--50\% for Game and 0--100\% elsewhere.}
\label{fig:domain-success-time}
\end{figure}

\clearpage
\section{Task-family performance}\label{app:task-families}

\clearpage
\section{Task-level outcomes and paired differences}\label{app:task-outcomes}
\begin{figure}[H]
\centering
\includegraphics[width=\linewidth]{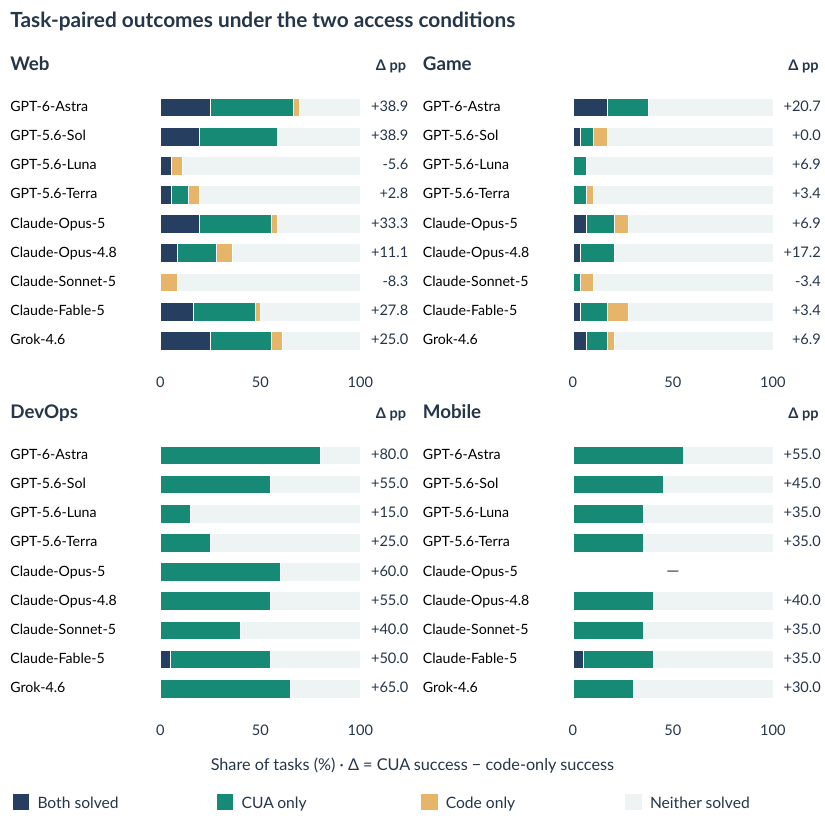}
\caption{\textbf{Task-paired outcomes under the two access conditions.} Each bar
partitions a model's matched tasks by the joint code-only and CUA outcome.
The four segments sum to 100\%. Values at right give the CUA minus code-only success difference
in percentage points, computed as the share solved only with CUA minus
the share solved only with code. Task membership is identical across
conditions within each model and domain. Both conditions include source-level execution and command feedback; hybrid CUA additionally provides application screenshots and graphical actions. Information-requirement strata are reported in Table~\ref{tab:information-strata}.}
\label{fig:paired-transitions}
\end{figure}

\begingroup
Figure~\ref{fig:computer-use-gain} compares development
with source-level execution against development that additionally uses
application screenshots and GUI actions. Both conditions provide feedback
from nonvisual commands; hybrid CUA also exposes the assembled
application's visual state and materials and supports interactive checks
of the effects of code changes.

\paragraph{Connecting visual evidence to the repair.}
The additional successes reveal two ways application access supports
software development. It exposes material that determines the required
implementation, such as received drawings and displayed service behavior.
It also lets agents exercise a changed application and inspect the outcome:
GPT-6-Astra's additional Game successes include delayed terminal state,
checkpoint synchronization, and camera behavior after restart. In Web,
its nested-selection trajectory uses a drag after resizing to discover
and repair a second interaction defect. These cases connect the access
comparison to concrete changes in the development process.

\paragraph{Web and Game.}
On Web, GPT-5.6-Sol achieves 19.4\% task success with code-only access
and 58.3\% with hybrid access; Claude-Opus-5 achieves 22.2\% and 55.6\%,
respectively. On Game, the corresponding rates are 3.4\% and 20.7\% for
Claude-Opus-4.8, and 17.2\% and 37.9\% for GPT-6-Astra.

\paragraph{DevOps and Mobile.}
Agents inspect service interfaces and visual materials,
use graphical controls to explore their behavior, and translate
these observations into code changes. On DevOps, the nine frontier models average 50.0\% with
hybrid access and 0.6\% with code alone. On Mobile, GPT-5.6-Sol
achieves 0.0\% with code-only access and 45.0\% with hybrid access;
GPT-6-Astra achieves 0.0\% and 55.0\%, respectively. Figure~\ref{fig:paired-transitions}
and Appendix~\ref{app:task-outcomes} provide task-level comparisons.

\paragraph{Performance by task information requirements.}
Application-material-dependent tasks account for 14 of
36 Web tasks, 9 of 29 Game tasks, and all 20 tasks in each of DevOps and
Mobile. Across the eight frontier models evaluated in both conditions in all
four domains, the mean hybrid advantage on these tasks is 42.0, 23.6,
48.1, and 38.8 percentage points, respectively. On source-specified tasks,
the corresponding mean differences are 0.0 points on Web and $-0.6$ on
Game. This distribution connects the aggregate comparison to the tasks'
information requirements: the largest gains occur when agents must turn
application materials into working software. Appendix~\ref{app:task-outcomes}
reports all model-level comparisons on matched task subsets.
For source-specified tasks, code-only agents can check behavior through
program execution, as the 2048 trajectory in
Appendix~\ref{app:code-execution-example} illustrates. GUI access adds
application-level evidence; completing the repair requires the agent to
interpret that evidence and act on it.
\par\endgroup
\clearpage
\begin{figure}[H]
\centering
\includegraphics[width=\linewidth]{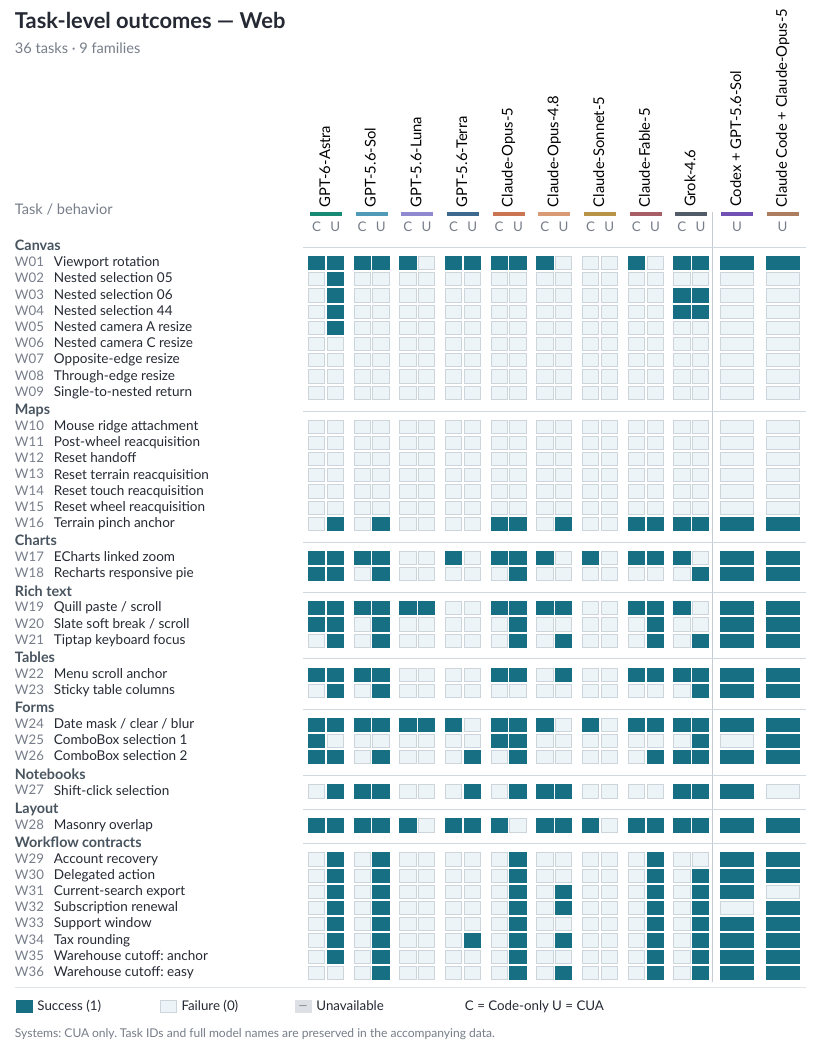}
\caption{\textbf{Task-level Web outcomes.} Rows cover all 36 Web tasks,
grouped by task family. Each frontier model has adjacent code-only (C) and CUA
(U) cells; each agent system has one CUA cell. Colors encode the recorded
outcome according to the key, and gray em dashes indicate unavailable
outcomes.}
\label{fig:outcomes-web}
\end{figure}

\clearpage
\begin{figure}[H]
\centering
\includegraphics[width=\linewidth]{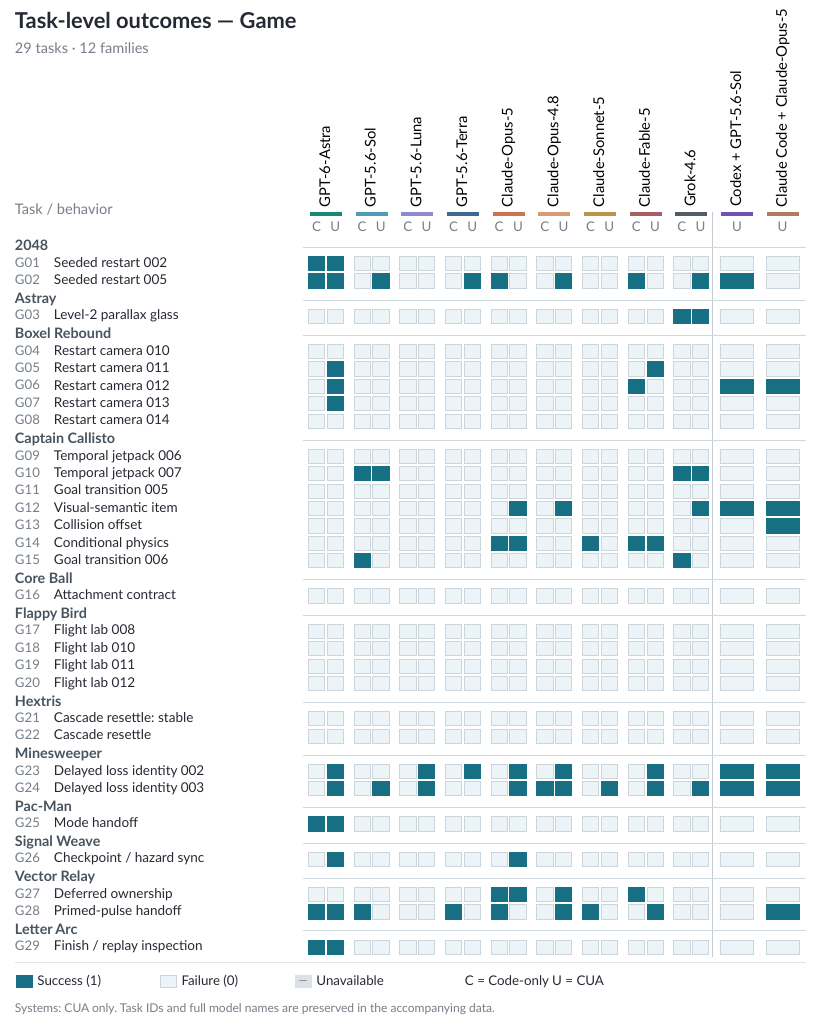}
\caption{\textbf{Task-level Game outcomes.} Rows cover all 29 Game tasks,
grouped by source game. Adjacent C and U cells show code-only and CUA
outcomes for frontier models; the two system columns show CUA outcomes. The
encoding follows Figure~\ref{fig:outcomes-web}.}
\label{fig:outcomes-game}
\end{figure}

\clearpage
\begin{figure}[H]
\centering
\includegraphics[width=\linewidth]{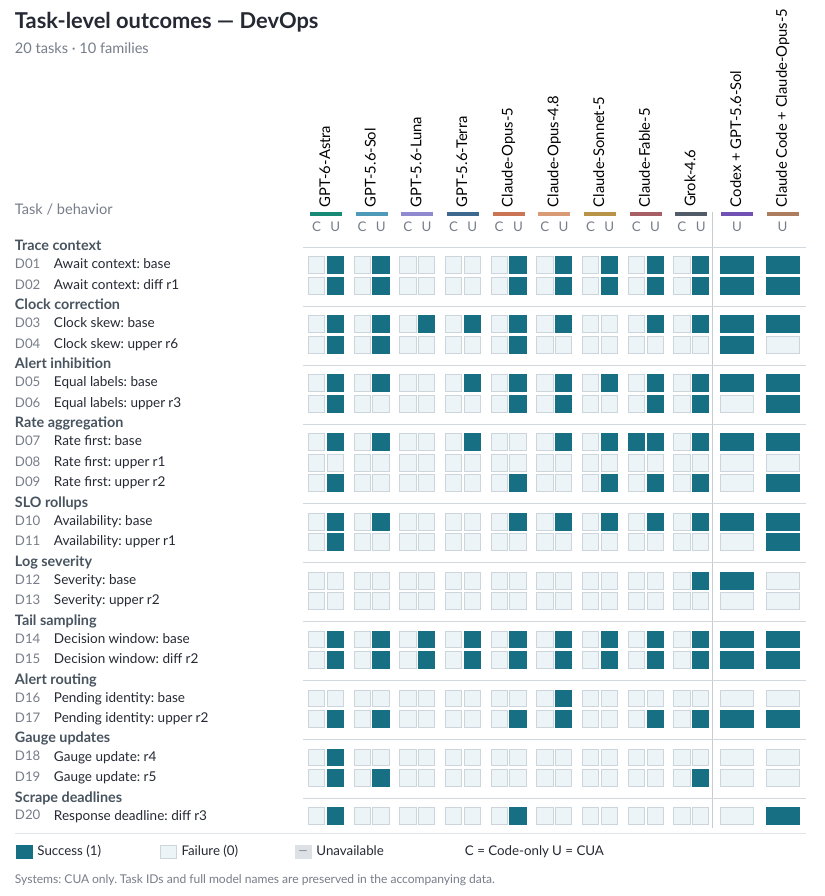}
\caption{\textbf{Task-level DevOps outcomes.} Rows cover all 20 DevOps
tasks, grouped by operational contract. Adjacent C and U cells show
code-only and CUA outcomes for frontier models; the two system columns show
CUA outcomes. The encoding follows Figure~\ref{fig:outcomes-web}.}
\label{fig:outcomes-devops}
\end{figure}

\clearpage
\begin{figure}[H]
\centering
\includegraphics[width=\linewidth]{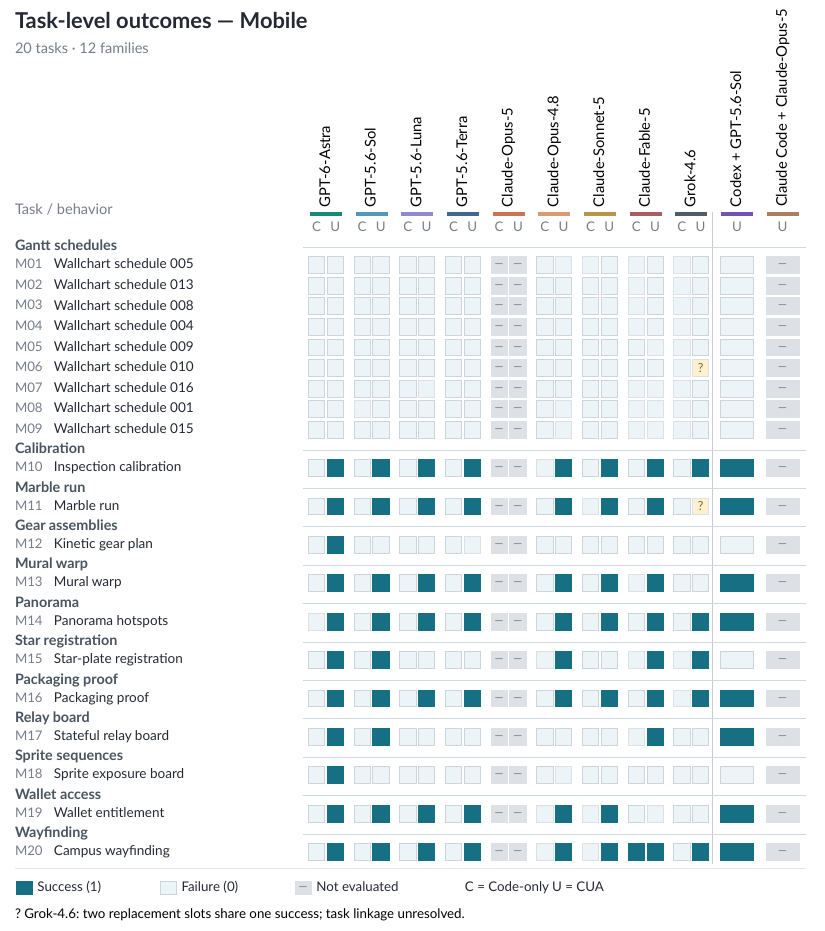}
\caption{\textbf{Task-level Mobile outcomes.} Rows cover all 20 Mobile
tasks, grouped by application contract. Adjacent C and U cells show
code-only and CUA outcomes for frontier models; the two system columns show
CUA outcomes. Gray em dashes denote unevaluated models.
Grok-4.6's two question marks identify tasks with one success between them
and unresolved individual outcomes (Appendix~\ref{app:cohort-reconciliation}).
Other colors follow
Figure~\ref{fig:outcomes-web}.}
\label{fig:outcomes-mobile}
\end{figure}

\clearpage

\clearpage
\paragraph{Paired outcomes by task information requirements.}
Tables~\ref{tab:information-summary}
and~\ref{tab:information-strata} stratify the original single-attempt
comparison by specification provenance. The aggregate summary uses the
same eight frontier models in every domain; the detailed table includes every
frontier model evaluated in both conditions in the relevant domain. All rates
within a row use exactly the same tasks. The M group includes five Game
tasks with bundled graphical reference cards; its other 58 tasks use
runtime-supplied materials. These comparisons describe performance across
task requirements under the two access conditions.
\input{tables/information-summary}
\input{tables/information-strata}

\clearpage
\section{Interaction workloads}\label{app:interaction-workloads}
\begingroup
\paragraph{Time and workload measures.}
Time is measured overthe agent's development episode. Agent time per
success divides mean episode time by task-success probability, using
equal domain weights for cross-domain summaries.
Appendix~\ref{app:results} describes the evaluation protocol, condition
comparisons, and workload accounting.

GPT-6-Astra achieves the highest four-domain task success rate, 59.9\%,
and the lowest agent time per success, 7.2 minutes, among the eight frontier
models (Figure~\ref{fig:hybrid-execution-efficiency}). Its Hybrid episodes average
46.3 model responses with 145.8 output tokens per response. GPT-5.6-Sol
uses 37.7 responses and 203.7 tokens per response, with 8.7 agent minutes
per success. These profiles characterize how each model distributes its
work across interaction rounds and response length.

\begin{figure}[H]
\centering
\includegraphics[width=0.62\linewidth]{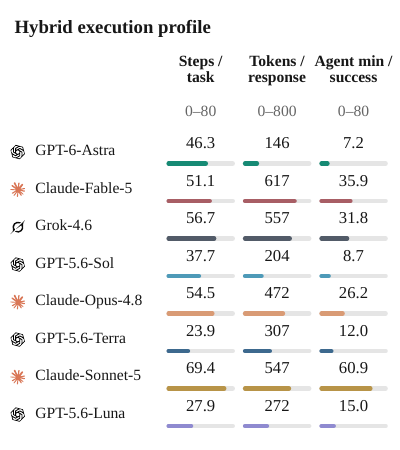}
\caption{\textbf{Hybrid execution efficiency.} Model responses per task,
output tokens per response, and agent minutes per success for eight frontier
models. Episode workloads are averaged within each domain and then weighted
equally across the four domains. Agent minutes per success divides that mean
cost by the full-inventory task-success probability.
Appendix~\ref{app:workload-accounting} specifies workload coverage.}
\label{fig:hybrid-execution-efficiency}
\end{figure}

\paragraph{Success and episode time.}
Figure~\ref{fig:success-effort}(a) summarizes task success and mean
development time across four equally weighted domains. GPT-6-Astra
achieves 59.9\% task success with a mean of 4.3 minutes per attempt;
GPT-5.6-Sol reaches 42.2\% at 3.7 minutes. The comparison covers nine
models and agent systems evaluated in all four domains.
Appendix~\ref{app:domain-performance} provides the domain-level profiles.

\paragraph{Effort on shared successful tasks.}
On Web tasks successfully repaired by both models, GPT-6-Astra uses fewer
responses on 83.3\% of tasks shared with Grok-4.6 and 72.2\% shared with
Claude-Opus-5 (Figure~\ref{fig:success-effort}(b)). Median response counts are
28.5 versus 43.5 and 21.5 versus 27.5, respectively. Each comparison uses
its own set of 18 shared successful tasks, linking interaction effort to
the same completed repairs. Appendix~\ref{app:shared-success} gives the
task-level pairs and resource comparisons across domains.
System-level results also
show different development profiles: Claude Code with Claude-Opus-5 achieves
65.0\% on DevOps at a median of 4.4 minutes, while the frontier model achieves
60.0\% at 9.3 minutes. Appendix~\ref{app:resource-profiles} gives the
response, token, and system-specific workload distributions.
\par\endgroup
\clearpage
\begin{figure}[H]
\centering
\includegraphics[width=\linewidth]{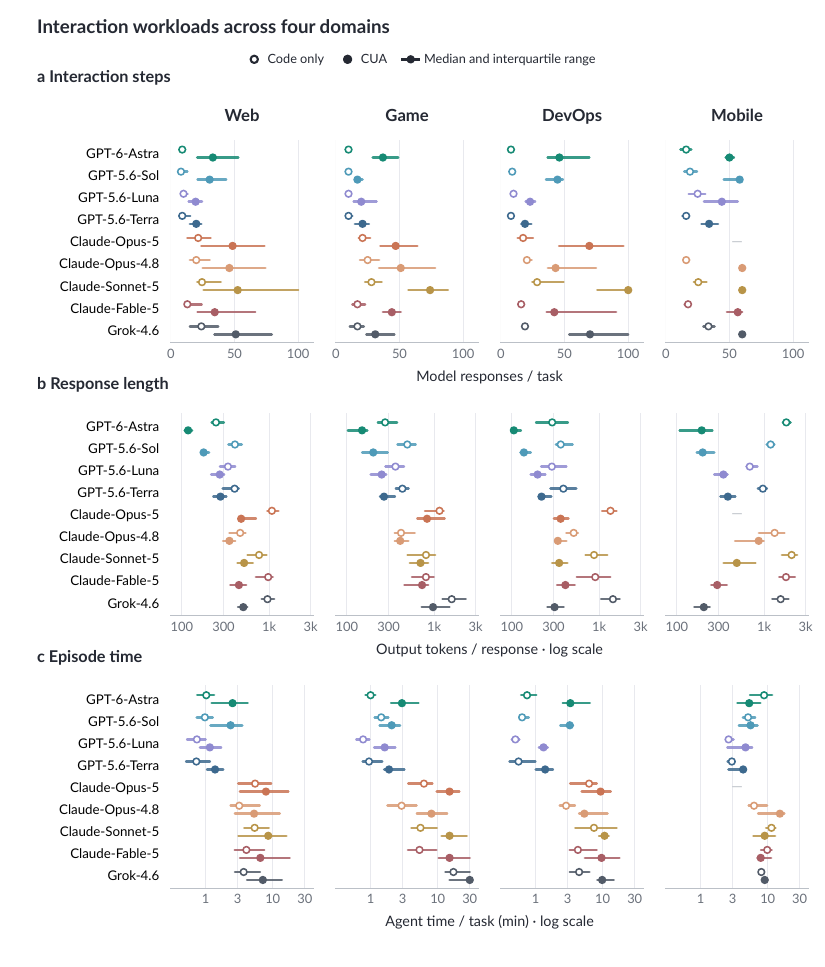}
\caption{\textbf{Interaction steps, response length, and development
time.} Hollow circles show code-only and filled circles show CUA.
Markers give task medians; intervals span the 25th--75th percentiles.
Steps count completed model responses. Response length is recorded output
tokens divided by completed responses for each episode, summarized
across tasks. Time covers the agent episode. Each metric uses the same
eligible tasks across conditions within a model and domain, retaining
the full range of task outcomes. Resource coverage is specified in
Appendix~\ref{app:workload-accounting}.}
\label{fig:interaction-workloads}
\end{figure}

\clearpage
\section{Effort on shared successful tasks}\label{app:shared-success}
\begin{figure}[H]
\centering
\includegraphics[width=\linewidth]{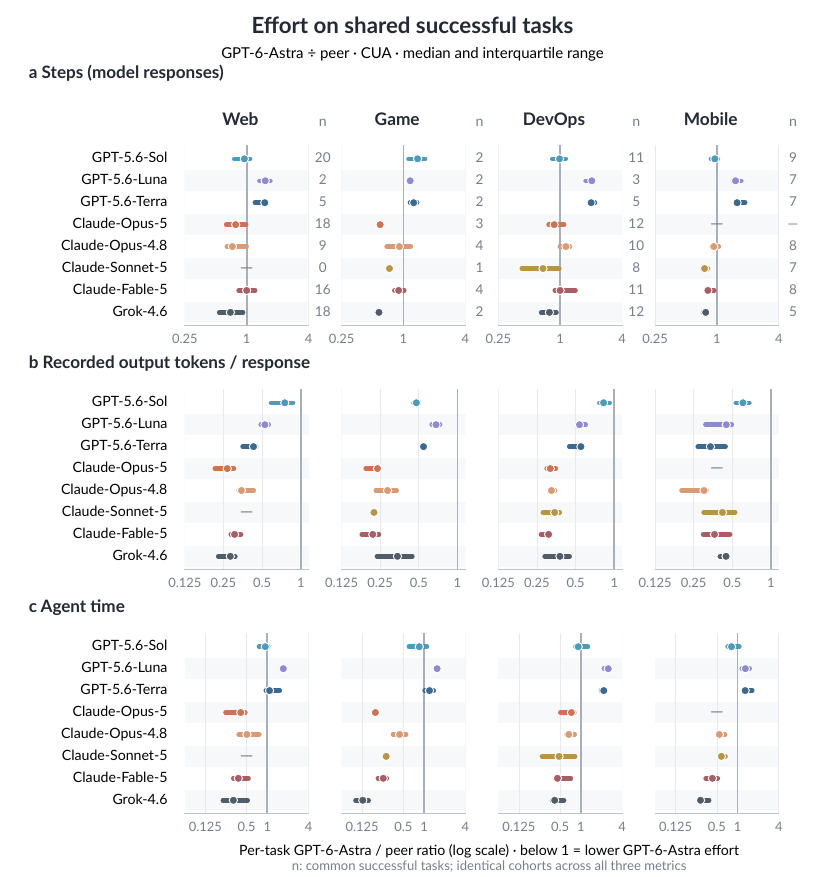}
\caption{\textbf{Effort ratios on shared successful tasks.} Each
comparison uses the tasks successfully completed by both GPT-6-Astra
and the corresponding frontier model with CUA. For each task, effort is divided
by the peer's effort; markers show median ratios and intervals span the
25th--75th percentiles. The line at one indicates equal effort.
Axes are logarithmic. Steps count model responses; response length uses
recorded output tokens per response; time covers the agent episode.
Each peer has its own shared-success cohort, whose size appears in the
top row. Pairs include executions with resource measurements for
both models (Appendix~\ref{app:workload-accounting}).}
\label{fig:shared-success-effort}
\end{figure}

\clearpage
\begin{figure}[H]
\centering
\includegraphics[width=\linewidth]{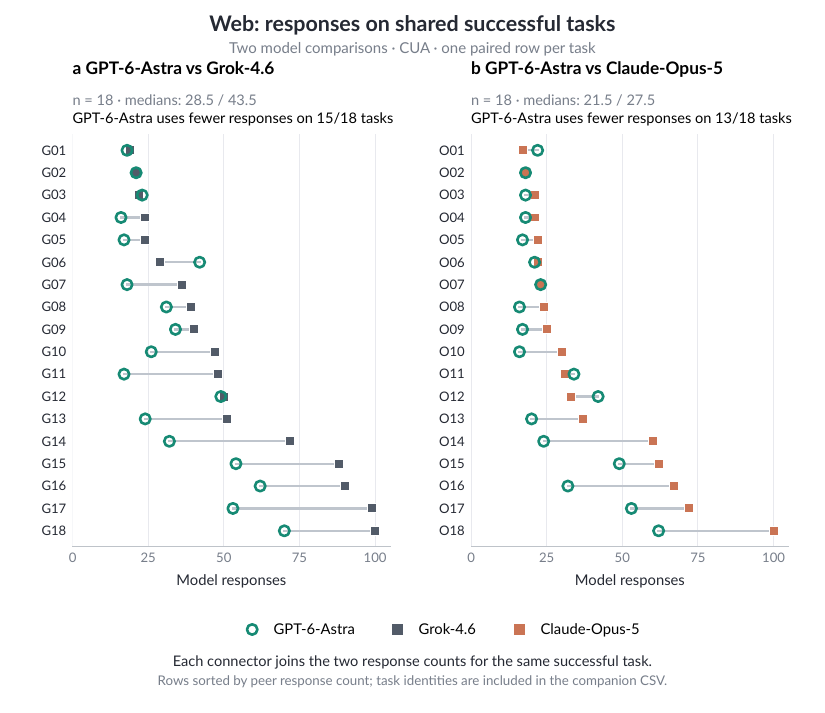}
\caption{\textbf{Response counts on shared successful Web tasks.}
Each connected pair gives GPT-6-Astra's and its peer's response counts
on one task. The GPT-6-Astra--Grok-4.6 comparison covers 18 shared successes,
with medians of 28.5 and 43.5 responses. The GPT-6-Astra--Claude-Opus-5 comparison
also covers 18 shared successes, with medians of 21.5 and 27.5 responses.
Each panel uses its corresponding pair's shared-success task set.}
\label{fig:web-shared-responses}
\end{figure}

\clearpage
\section{Resource profiles}\label{app:resource-profiles}
\begin{figure}[H]
\centering
\includegraphics[width=\linewidth]{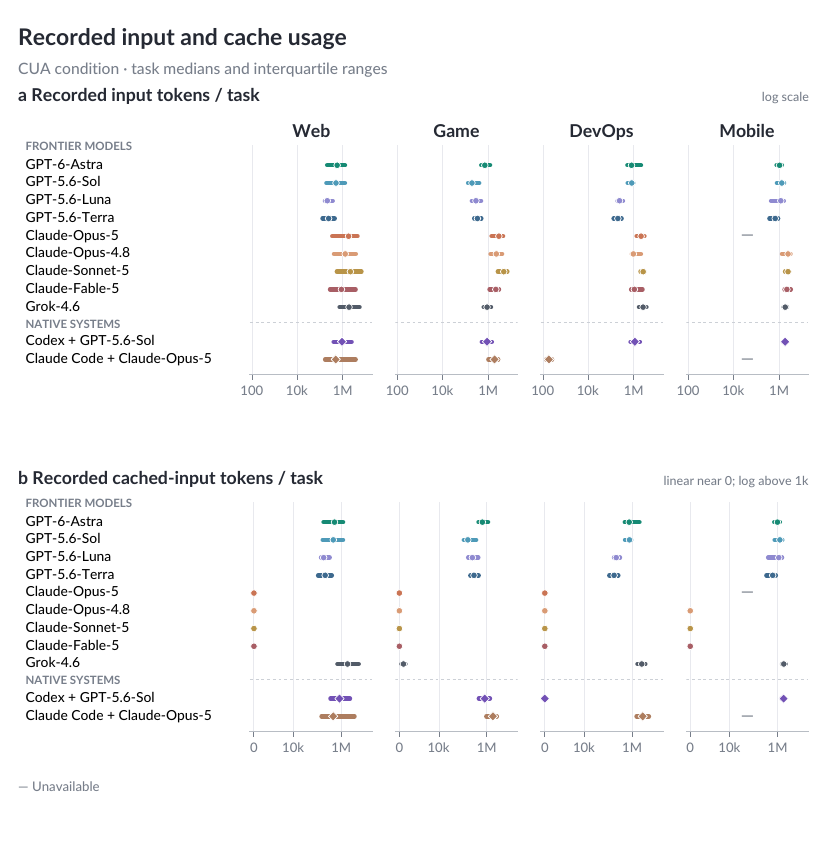}
\caption{\textbf{Recorded input and cached-input tokens per CUA task.}
Markers show medians and intervals span the 25th--75th percentiles
over each entry's measured CUA cohort. The two counters retain their
provider-specific definitions and are summarized separately. Input
tokens use a logarithmic axis. Cached-input tokens use a symmetric
logarithmic axis with a linear region from zero to 1,000 tokens;
tick labels give token counts. Token summaries use episodes with
complete recorded usage.}
\label{fig:input-cache-profiles}
\end{figure}

\clearpage
\begin{figure}[H]
\centering
\includegraphics[width=\linewidth]{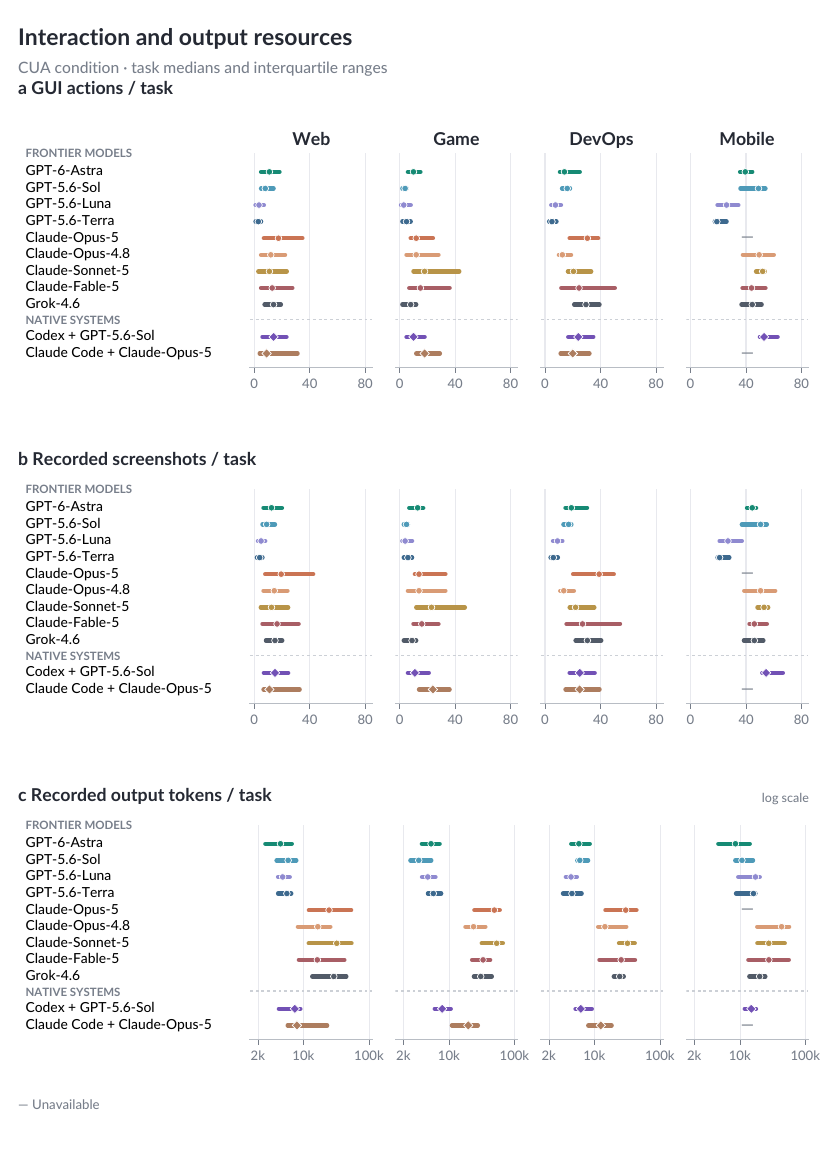}
\caption{\textbf{CUA interaction and output profiles.} Markers show
medians and intervals span the 25th--75th percentiles over each entry's
measured task cohort. Circles denote frontier models and diamonds denote
agent systems. GUI actions and recorded screenshots use linear axes;
output tokens use a logarithmic axis. Screenshot counts retain each
source's capture definition, and token summaries use complete recorded
usage.}
\label{fig:interaction-output-profiles}
\end{figure}

\clearpage
\begin{figure}[H]
\centering
\includegraphics[width=\linewidth]{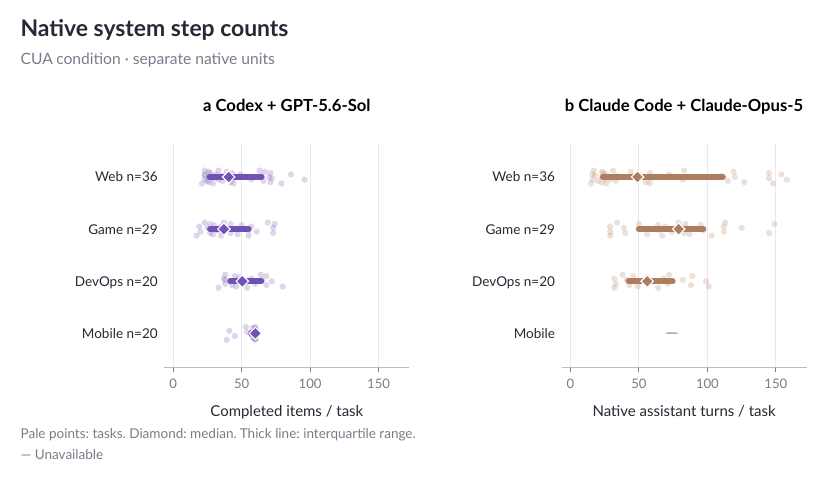}
\caption{\textbf{Native agent-system steps with CUA.} Codex with
GPT-5.6-Sol counts completed items; Claude Code with Claude-Opus-5 counts
native assistant turns. The two step units appear in separately
labeled panels. Pale points show individual tasks, diamonds give
medians, and thick intervals span the 25th--75th percentiles. Each panel
uses all available native step measurements for the system and domain.}
\label{fig:system-steps}
\end{figure}

\clearpage
\section{Successful repair trajectories}\label{app:repair-cases}
\begin{figure}[H]
\centering
\includegraphics[width=\linewidth]{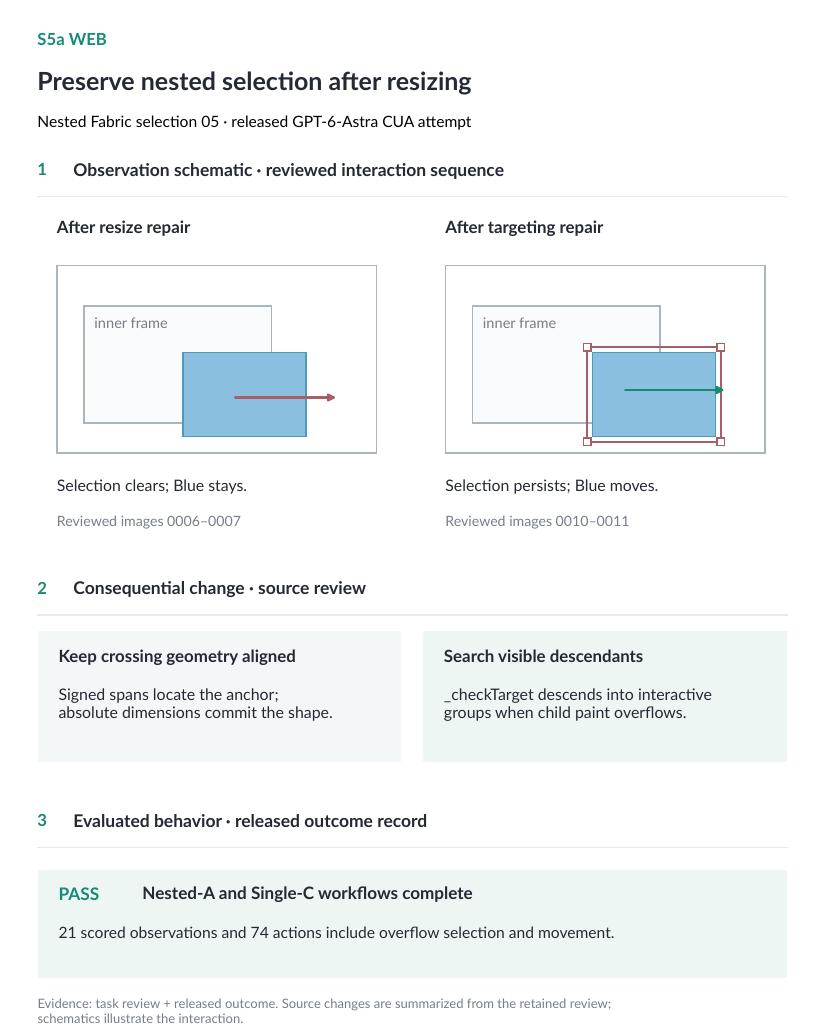}
\caption{\textbf{Successive interaction-guided edits on Web.}
GPT-6-Astra repairs resizing, tests selection beyond the containing
frame, and extends hit testing to selectable descendants. The scene
panels are conceptual schematics of the reviewed trajectory; the edit
description summarizes the source change. The resulting patch passes
the task's evaluator.}
\label{fig:case-web}
\end{figure}

These cases illustrate four repair mechanisms through
first-eligible trajectories from the single-attempt evaluation. Each
case follows the agent's observations, source edits, and verification
results.
\begingroup
\paragraph{Interaction guides successive edits.}

In the Web nested-selection task, GPT-6-Astra repairs resizing and then tests
a drag beyond the containing frame. The resulting selection behavior
motivates a second edit: the group's hit testing is extended to consider
selectable descendants outside its bounds. The agent repeats the interaction
and submits a passing patch, completing both geometry and selection behavior
(Figure~\ref{fig:case-web}).
\par\endgroup

\begin{figure}[H]
\centering
\includegraphics[width=\linewidth]{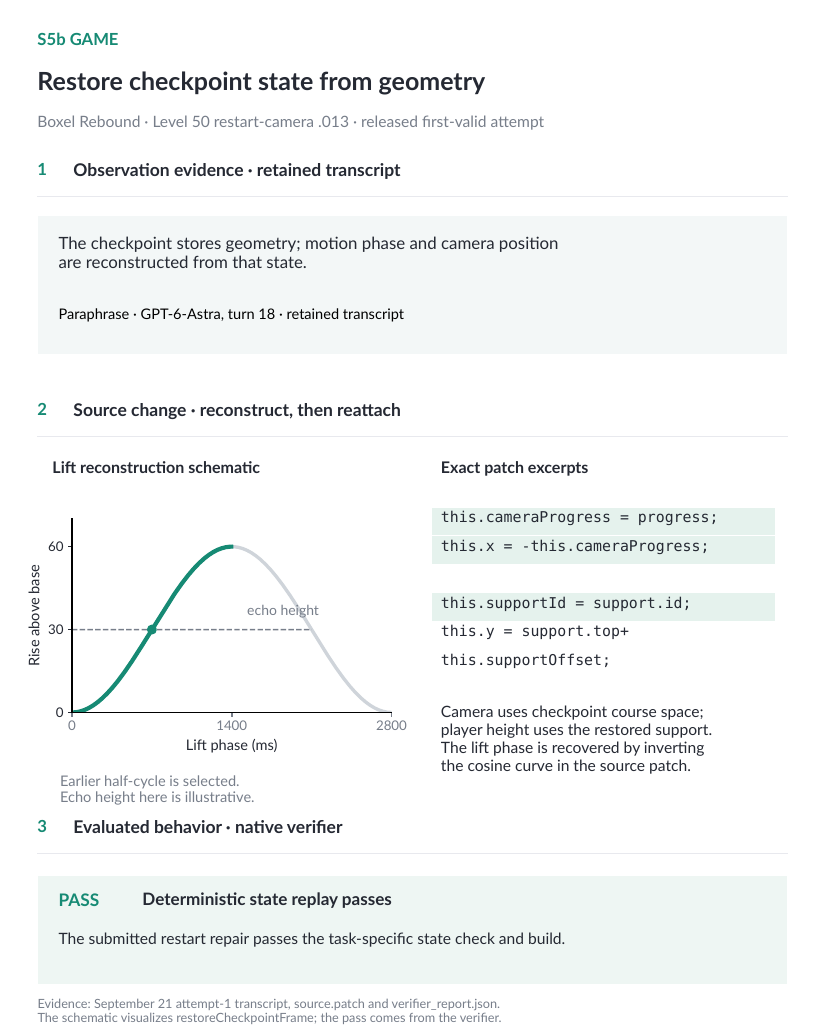}
\caption{\textbf{Coordinated restart restoration in Game.} GPT-6-Astra combines lift-phase recovery, course-space camera restoration,
and support-relative player attachment. The phase curve is a
source-based schematic with an illustrative echo height, and the
turn-18 text paraphrases the released first-valid trajectory. Deterministic
state replay verifies the resulting repair. This episode belongs to the
single-attempt evaluation in Table~\ref{tab:four-domain-results}.}
\label{fig:case-game}
\end{figure}

\clearpage
\begingroup
\paragraph{State updates jointly restore the scene.}

In the Game checkpoint-restart task, GPT-6-Astra reconstructs lift phase from a
faded reference frame, restores camera progress in course coordinates, and
reattaches the player at its saved support-relative height. The patch also
preserves the checkpoint against updates from the beacon's static map cell.
Independent build and state-replay checks pass, establishing the combined
repair (Figure~\ref{fig:case-game}).
\par\endgroup

\begin{figure}[H]
\centering
\includegraphics[width=\linewidth]{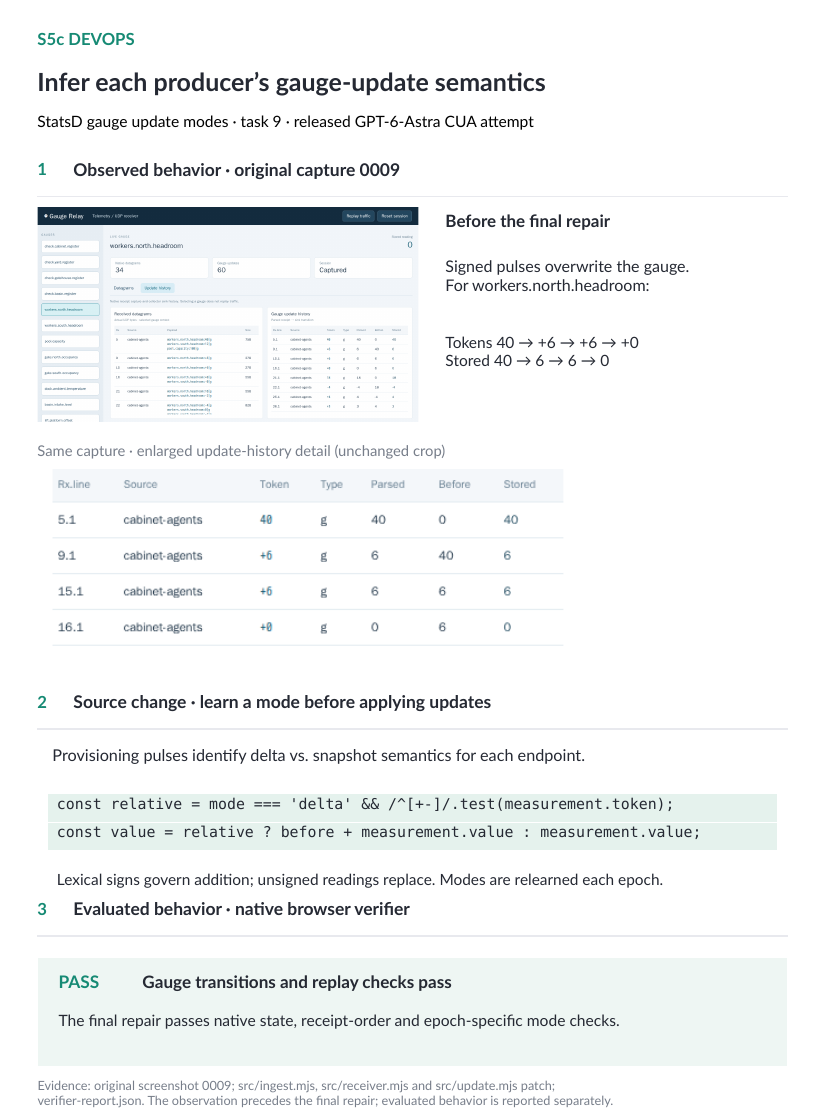}
\caption{\textbf{Recovering gauge-update semantics in DevOps.}
The recorded observation and its update-history crop show behavior
during development. Opening provisioning pulses reveal producer
update modes; the source excerpts preserve signed update tokens and
apply the inferred modes. The final browser check verifies GPT-6-Astra's repair.}
\label{fig:case-devops}
\end{figure}

\clearpage
\begingroup
\paragraph{Live behavior identifies the operative contract.}

In the DevOps gauge task, GPT-6-Astra uses a pulse--pulse--settle sequence to infer
each endpoint's update mode. The repair resets that mode at each epoch and
preserves receipt order across rollout transitions
(Figure~\ref{fig:case-devops}). This connects visual observations of the service's
responses to the semantics implemented in code.
\par\endgroup

\begin{figure}[H]
\centering
\includegraphics[width=\linewidth]{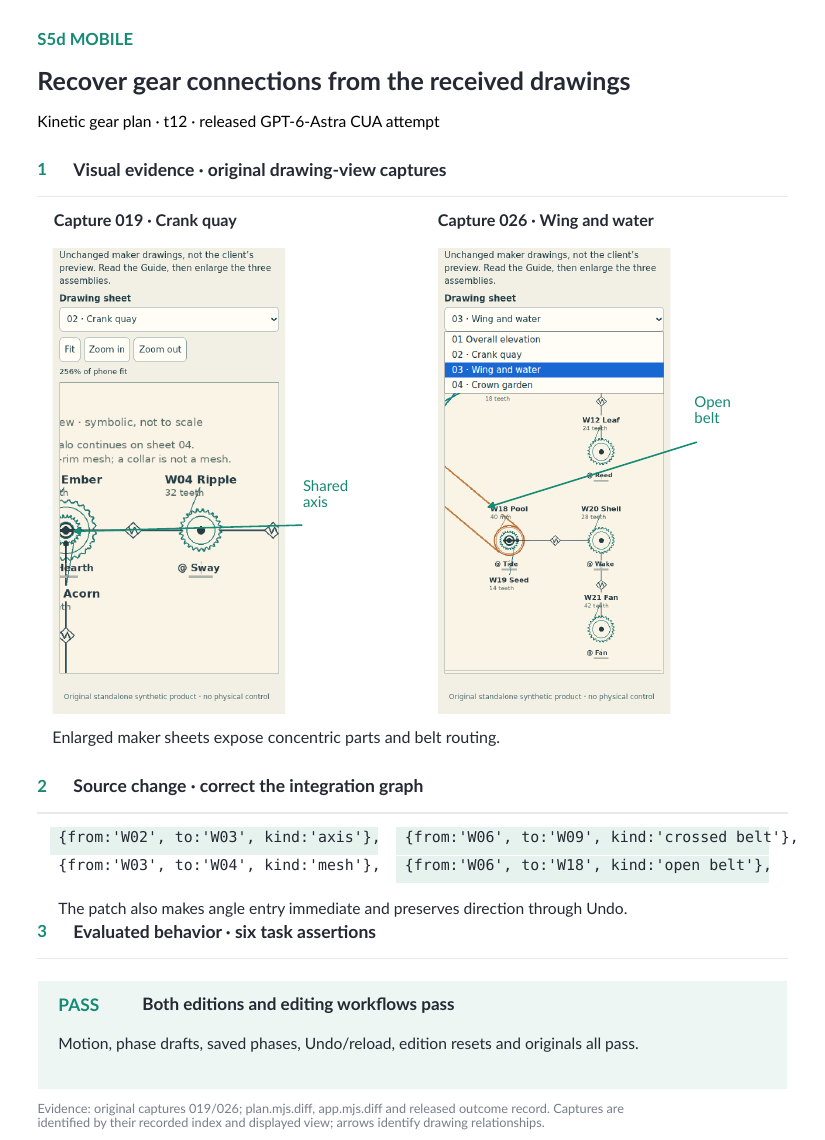}
\caption{\textbf{Translating a Mobile drawing into gear relationships.}
The two recorded drawing views guide GPT-6-Astra's graph repair.
The source excerpt shows four selected added edges from the patch.
The completed repair passes all six task assertions.}
\label{fig:case-mobile}
\end{figure}

\clearpage
\begingroup
\paragraph{Visual structure becomes executable behavior.}

On Mobile, GPT-6-Astra reconstructs all 23 gear relations and implements their
phase, angle, and editing behavior (Figure~\ref{fig:case-mobile}). In the
sprite task, it recovers all 28 approved pose assignments together with timing
and transport behavior. Both repairs carry precise visual relationships
through the application's interactive operations.
\par\endgroup

\begingroup
\subsection{Paired repairs and remaining capability demands}\label{app:capability-comparisons}
\paragraph{Selection after a geometry change.}
The nested-selection comparison uses the same task under hybrid access.
GPT-6-Astra repairs signed resizing and then extends descendant hit testing
after its own overflow drag loses selection. Its final patch passes both
the resize and selection workflows. GPT-5.6-Sol removes the resize clamps
but reverts its experimental hit-testing changes before submission;
Claude-Opus-5 also removes the clamps without repairing descendant discovery.
Both patches pass the reached resize prefix but lose selection when the
visible object is pressed outside the containing frame. This contrast
links their different outcomes to the second repair, which preserves the
interaction after the geometry change.

\paragraph{Exact visual relations and animation assignments.}
In the kinetic-gear task, GPT-6-Astra recovers the 23 driving relations.
GPT-5.6-Sol's reconstruction selects the wrong immediate driver for two
compound assemblies: it substitutes an outer rim for the connected inner
rim, changing downstream motion. In the sprite task, GPT-6-Astra recovers
all 28 approved pose assignments, while GPT-5.6-Terra swaps two crossing
poses and selects an incorrect late catching pose. GPT-5.6-Sol completes
the sprite task on its second independent attempt. These comparisons
distinguish the accuracy of the recovered relations from the ability to
implement the surrounding interaction and animation machinery.

\paragraph{Service semantics and complementary model strengths.}
In the gauge-calibration task, GPT-6-Astra is the only successful entry.
It infers update modes from the opening pulse--pulse--settle sequence,
resets them at each deployment epoch, and preserves receipt order.
Other service tasks favor different agents. Claude-Opus-4.8 uniquely
repairs the receiver-selector task by propagating route metadata into
rule compilation and retaining the required selector labels. GPT-6-Astra
instead separates literal and templated enrichments without recovering
the route's actual identity keys. In log normalization, Grok-4.6 and
Codex combine delivered format with producer provenance; GPT-6-Astra
changes the format selection but leaves relayed syslog severities
interpreted under the wrong convention. The final verifier outcomes
match these differences in the submitted implementations.

\paragraph{Coupled game state and dense schedule dependencies.}
No entry completes Core Ball in the original single-attempt evaluation.
GPT-6-Astra's patch corrects impact-angle ownership but leaves failures in
waiting-shot promotion, post-commit ordering, pace-change persistence,
completion, and rejected-collision occupancy. In Mobile, the original
hybrid attempts leave all nine schedule variants unsolved. The inspected
GPT-6-Astra schedule patch recovers starts and durations but misassigns
two dependency endpoints and one lag; its self-tests encode the same
incorrect transcription. The repeated evaluation adds one GPT-6-Astra
schedule success on the third attempt. These observations identify the
joint recovery of event ordering and relational constraints as a demand
that remains beyond the successful local repairs in these cases.
\par\endgroup

\clearpage
\section{Patch correctness and task success}\label{app:score-decomposition}
We report three complementary measures for each model and
access condition. Patch correctness averages $P_i$, task success averages
$S_i=P_iU_i$, and the third measure averages $P_i(1-U_i)$ over the same
$N$ task attempts. All three are expressed as percentages. The first
measure summarizes verified repairs; the latter two partition them by
trajectory validation. Table~\ref{tab:score-decomposition} gives the
results across the four domains.
\input{tables/score-decomposition}

%% file: tables/hybrid-pass3.tex
\begin{table}[htbp]
\centering\footnotesize
\caption{\textbf{Three-attempt hybrid CUA success (\%).} Each cell reports pass@1 / pass@3. Pass@1 uses the first attempt; pass@3 uses all three.}
\label{tab:hybrid-pass3}
\setlength{\tabcolsep}{3pt}
\begin{tabular}{@{}lccc@{}}
\toprule
Domain & \shortstack{GPT-6-\\Astra} & \shortstack{GPT-5.6-\\Sol} & \shortstack{Claude-\\Fable-5} \\
\midrule
Web & \textbf{66.7 / 75.0} & 58.3 / 58.3 & 47.2 / 61.1 \\
Game & \textbf{37.9 / 62.1} & 10.3 / 31.0 & 17.2 / 41.4 \\
DevOps & \textbf{80.0 / 90.0} & 55.0 / 60.0 & 55.0 / 70.0 \\
Mobile & \textbf{55.0 / 60.0} & 45.0 / 50.0 & 40.0 / 40.0 \\
\bottomrule
\end{tabular}
\end{table}

%% file: tables/repeated-passk.tex
\begin{table}[htbp]
\centering\small
\setlength{\tabcolsep}{3pt}
\caption{\textbf{Task success across attempt budgets (\%).} Mean of three is the average single-attempt success within this repeated cohort. Pass@2 averages all two-attempt subsets; pass@3 uses all three. The difference is pass@3 minus mean single-attempt success, in percentage points. Results cover Web and Game.}
\label{tab:repeated-passk}
\begin{tabular}{@{}llrrrr@{}}
\toprule
Domain & Model & Mean of three & Pass@2 & Pass@3 & Diff. (pp) \\
\midrule
Web & GPT-6-Astra & 67.6 & 72.2 & 75.0 & +7.4 \\
 & GPT-5.6-Sol & 52.8 & 57.4 & 58.3 & +5.6 \\
 & Claude-Fable-5 & 52.8 & 59.3 & 61.1 & +8.3 \\
\midrule
Game & GPT-6-Astra & 50.6 & 59.8 & 62.1 & +11.5 \\
 & GPT-5.6-Sol & 20.7 & 29.9 & 31.0 & +10.3 \\
 & Claude-Fable-5 & 27.6 & 36.8 & 41.4 & +13.8 \\
\bottomrule
\end{tabular}
\end{table}

%% file: tables/repeated-success-frequency.tex
\begin{table}[htbp]
\centering\small
\setlength{\tabcolsep}{3pt}
\caption{\textbf{Success frequency across three attempts.} Entries give the percentage of tasks with zero, one, two, or three successful attempts. Each row sums to 100\% before rounding. Results cover Web and Game.}
\label{tab:repeated-success-frequency}
\begin{tabular}{@{}llrrrr@{}}
\toprule
Domain & Model & 0 successes & 1 success & 2 successes & 3 successes \\
\midrule
Web & GPT-6-Astra & 25.0 & 8.3 & 5.6 & 61.1 \\
 & GPT-5.6-Sol & 41.7 & 2.8 & 11.1 & 44.4 \\
 & Claude-Fable-5 & 38.9 & 5.6 & 13.9 & 41.7 \\
\midrule
Game & GPT-6-Astra & 37.9 & 6.9 & 20.7 & 34.5 \\
 & GPT-5.6-Sol & 69.0 & 3.4 & 24.1 & 3.4 \\
 & Claude-Fable-5 & 58.6 & 13.8 & 13.8 & 13.8 \\
\bottomrule
\end{tabular}
\end{table}

%% file: tables/repeated-paired-differences.tex
\begin{table}[htbp]
\centering\footnotesize
\setlength{\tabcolsep}{3pt}
\caption{\textbf{Paired task-success differences for GPT-6-Astra (pp).} Intervals are 95\% percentile intervals from 20,000 paired resamples. Task resampling operates within each domain; family resampling selects whole task families and retains task weighting. All model outcomes and three attempts for each task remain together. Results cover Web and Game.}
\label{tab:repeated-paired-differences}
\begin{tabular}{@{}llcrrr@{}}
\toprule
Domain & Compared with & Measure & Difference & Task 95\% CI & Family 95\% CI \\
\midrule
Web & GPT-5.6-Sol & Mean of three & +14.8 & [3.7, 25.9] & [-1.3, 33.3] \\
 & GPT-5.6-Sol & Pass@3 & +16.7 & [2.8, 30.6] & [-6.7, 42.9] \\
 & Claude-Fable-5 & Mean of three & +14.8 & [3.7, 25.9] & [0.0, 32.5] \\
 & Claude-Fable-5 & Pass@3 & +13.9 & [0.0, 27.8] & [-4.3, 33.3] \\
\midrule
Game & GPT-5.6-Sol & Mean of three & +29.9 & [17.2, 43.7] & [15.3, 47.2] \\
 & GPT-5.6-Sol & Pass@3 & +31.0 & [13.8, 48.3] & [11.5, 50.0] \\
 & Claude-Fable-5 & Mean of three & +23.0 & [8.0, 39.1] & [10.0, 42.0] \\
 & Claude-Fable-5 & Pass@3 & +20.7 & [0.0, 41.4] & [0.0, 41.4] \\
\bottomrule
\end{tabular}
\end{table}

%% file: tables/four-domain-results.tex
\begin{table}[htbp]
\centering\small
\caption{\textbf{Single-attempt task success (\%) across four domains.} Code denotes code-only access; hybrid CUA adds screenshots and graphical actions. Teal marks column maxima; an em dash indicates no reported result.}
\label{tab:four-domain-results}
\setlength{\tabcolsep}{3pt}
\resizebox{0.84\linewidth}{!}{%
\begin{tabular}{@{}lrrrrrrrr@{}}
\toprule
& \multicolumn{2}{c}{Web} & \multicolumn{2}{c}{Game} & \multicolumn{2}{c}{DevOps} & \multicolumn{2}{c}{Mobile} \\
\cmidrule(lr){2-3}\cmidrule(lr){4-5}\cmidrule(lr){6-7}\cmidrule(lr){8-9}
Model / system & Code & \shortstack{Hybrid\\CUA} & Code & \shortstack{Hybrid\\CUA} & Code & \shortstack{Hybrid\\CUA} & Code & \shortstack{Hybrid\\CUA} \\
\midrule
GPT-6-Astra & 27.8 & \textcolor[HTML]{167D8D}{\textbf{66.7}} & \textcolor[HTML]{167D8D}{\textbf{17.2}} & \textcolor[HTML]{167D8D}{\textbf{37.9}} & 0.0 & \textcolor[HTML]{167D8D}{\textbf{80.0}} & 0.0 & \textcolor[HTML]{167D8D}{\textbf{55.0}} \\
GPT-5.6-Sol & 19.4 & 58.3 & 10.3 & 10.3 & 0.0 & 55.0 & 0.0 & 45.0 \\
GPT-5.6-Luna & 11.1 & 5.6 & 0.0 & 6.9 & 0.0 & 15.0 & 0.0 & 35.0 \\
GPT-5.6-Terra & 11.1 & 13.9 & 3.4 & 6.9 & 0.0 & 25.0 & 0.0 & 35.0 \\
Claude-Opus-5 & 22.2 & 55.6 & 13.8 & 20.7 & 0.0 & 60.0 & \textemdash & \textemdash \\
Claude-Opus-4.8 & 16.7 & 27.8 & 3.4 & 20.7 & 0.0 & 55.0 & 0.0 & 40.0 \\
Claude-Sonnet-5 & 8.3 & 0.0 & 6.9 & 3.4 & 0.0 & 40.0 & 0.0 & 35.0 \\
Claude-Fable-5 & 19.4 & 47.2 & 13.8 & 17.2 & \textcolor[HTML]{167D8D}{\textbf{5.0}} & 55.0 & \textcolor[HTML]{167D8D}{\textbf{5.0}} & 40.0 \\
Grok-4.6 & \textcolor[HTML]{167D8D}{\textbf{30.6}} & 55.6 & 10.3 & 17.2 & 0.0 & 65.0 & 0.0 & 30.0 \\
\midrule
Codex + GPT-5.6-Sol & \textemdash & 55.6 & \textemdash & 17.2 & \textemdash & 55.0 & \textemdash & 40.0 \\
Claude Code + Claude-Opus-5 & \textemdash & 55.6 & \textemdash & 20.7 & \textemdash & 65.0 & \textemdash & \textemdash \\
\bottomrule
\end{tabular}}
\end{table}

%% file: tables/information-summary.tex
\begingroup
\begin{table}[htbp]\centering\small
\caption{Success by task information requirement, averaged over the eight frontier models evaluated under both access conditions in all four domains. Each domain--type group uses the same tasks for both conditions. M: application-material dependent; S: source-specified. The labels identify specification provenance; both groups require implementation and verification. Full model results appear in Appendix~\ref{app:task-outcomes}.}\label{tab:information-summary}
\setlength{\tabcolsep}{6pt}
\begin{tabular}{@{}llrrrr@{}}\toprule
Domain & Type & Tasks & Code-only & Hybrid CUA & $\Delta$ (pp) \\ \midrule
Web & S & 22 & 23.9 & 23.9 & +0.0 \\
Web & M & 14 & 8.9 & 50.9 & +42.0 \\
Game & S & 20 & 10.6 & 10.0 & -0.6 \\
Game & M & 9 & 2.8 & 26.4 & +23.6 \\
DevOps & M & 20 & 0.6 & 48.8 & +48.1 \\
Mobile & M & 20 & 0.6 & 39.4 & +38.8 \\
\bottomrule\end{tabular}
\end{table}
\endgroup

%% file: tables/information-strata.tex
\begingroup\small
\setlength{\tabcolsep}{5pt}\setlength{\LTcapwidth}{\linewidth}
\begin{longtable}{@{}llrrrr@{}}
\caption{Task success by information requirement (\%). Each row compares the same tasks and original single attempts under code-only and hybrid CUA access. M and S follow Appendix~\ref{sec:runtime-information}; $\Delta$ is hybrid minus code-only in percentage points. DevOps and Mobile contain no S tasks; their S results are not applicable. Only models evaluated under both access conditions are included.}\label{tab:information-strata}\\
\toprule Type & Model & $N$ & Code-only & Hybrid CUA & $\Delta$ \\ \midrule\endfirsthead
\toprule Type & Model & $N$ & Code-only & Hybrid CUA & $\Delta$ \\ \midrule\endhead
\bottomrule\endfoot
\multicolumn{6}{l}{\textbf{Web}}\\*
S & GPT-6-Astra & 22 & 27.3 & 54.5 & +27.3 \\
S & GPT-5.6-Sol & 22 & 27.3 & 36.4 & +9.1 \\
S & GPT-5.6-Luna & 22 & 13.6 & 4.5 & -9.1 \\
S & GPT-5.6-Terra & 22 & 18.2 & 13.6 & -4.5 \\
S & Claude-Opus-5 & 22 & 27.3 & 31.8 & +4.5 \\
S & Claude-Opus-4.8 & 22 & 22.7 & 18.2 & -4.5 \\
S & Claude-Sonnet-5 & 22 & 13.6 & 0.0 & -13.6 \\
S & Claude-Fable-5 & 22 & 27.3 & 27.3 & +0.0 \\
S & Grok-4.6 & 22 & 40.9 & 36.4 & -4.5 \\
M & GPT-6-Astra & 14 & 28.6 & 85.7 & +57.1 \\
M & GPT-5.6-Sol & 14 & 7.1 & 92.9 & +85.7 \\
M & GPT-5.6-Luna & 14 & 7.1 & 7.1 & +0.0 \\
M & GPT-5.6-Terra & 14 & 0.0 & 14.3 & +14.3 \\
M & Claude-Opus-5 & 14 & 14.3 & 92.9 & +78.6 \\
M & Claude-Opus-4.8 & 14 & 7.1 & 42.9 & +35.7 \\
M & Claude-Sonnet-5 & 14 & 0.0 & 0.0 & +0.0 \\
M & Claude-Fable-5 & 14 & 7.1 & 78.6 & +71.4 \\
M & Grok-4.6 & 14 & 14.3 & 85.7 & +71.4 \\
\midrule
\multicolumn{6}{l}{\textbf{Game}}\\*
S & GPT-6-Astra & 20 & 25.0 & 25.0 & +0.0 \\
S & GPT-5.6-Sol & 20 & 15.0 & 10.0 & -5.0 \\
S & GPT-5.6-Luna & 20 & 0.0 & 0.0 & +0.0 \\
S & GPT-5.6-Terra & 20 & 5.0 & 5.0 & +0.0 \\
S & Claude-Opus-5 & 20 & 20.0 & 10.0 & -10.0 \\
S & Claude-Opus-4.8 & 20 & 0.0 & 15.0 & +15.0 \\
S & Claude-Sonnet-5 & 20 & 10.0 & 0.0 & -10.0 \\
S & Claude-Fable-5 & 20 & 15.0 & 10.0 & -5.0 \\
S & Grok-4.6 & 20 & 15.0 & 15.0 & +0.0 \\
M & GPT-6-Astra & 9 & 0.0 & 66.7 & +66.7 \\
M & GPT-5.6-Sol & 9 & 0.0 & 11.1 & +11.1 \\
M & GPT-5.6-Luna & 9 & 0.0 & 22.2 & +22.2 \\
M & GPT-5.6-Terra & 9 & 0.0 & 11.1 & +11.1 \\
M & Claude-Opus-5 & 9 & 0.0 & 44.4 & +44.4 \\
M & Claude-Opus-4.8 & 9 & 11.1 & 33.3 & +22.2 \\
M & Claude-Sonnet-5 & 9 & 0.0 & 11.1 & +11.1 \\
M & Claude-Fable-5 & 9 & 11.1 & 33.3 & +22.2 \\
M & Grok-4.6 & 9 & 0.0 & 22.2 & +22.2 \\
\midrule
\multicolumn{6}{l}{\textbf{DevOps}}\\*
M & GPT-6-Astra & 20 & 0.0 & 80.0 & +80.0 \\
M & GPT-5.6-Sol & 20 & 0.0 & 55.0 & +55.0 \\
M & GPT-5.6-Luna & 20 & 0.0 & 15.0 & +15.0 \\
M & GPT-5.6-Terra & 20 & 0.0 & 25.0 & +25.0 \\
M & Claude-Opus-5 & 20 & 0.0 & 60.0 & +60.0 \\
M & Claude-Opus-4.8 & 20 & 0.0 & 55.0 & +55.0 \\
M & Claude-Sonnet-5 & 20 & 0.0 & 40.0 & +40.0 \\
M & Claude-Fable-5 & 20 & 5.0 & 55.0 & +50.0 \\
M & Grok-4.6 & 20 & 0.0 & 65.0 & +65.0 \\
\midrule
\multicolumn{6}{l}{\textbf{Mobile}}\\*
M & GPT-6-Astra & 20 & 0.0 & 55.0 & +55.0 \\
M & GPT-5.6-Sol & 20 & 0.0 & 45.0 & +45.0 \\
M & GPT-5.6-Luna & 20 & 0.0 & 35.0 & +35.0 \\
M & GPT-5.6-Terra & 20 & 0.0 & 35.0 & +35.0 \\
M & Claude-Opus-4.8 & 20 & 0.0 & 40.0 & +40.0 \\
M & Claude-Sonnet-5 & 20 & 0.0 & 35.0 & +35.0 \\
M & Claude-Fable-5 & 20 & 5.0 & 40.0 & +35.0 \\
M & Grok-4.6 & 20 & 0.0 & 30.0 & +30.0 \\
\end{longtable}
\endgroup

%% file: tables/score-decomposition.tex
\begingroup
\small\setlength{\tabcolsep}{4pt}\setlength{\LTcapwidth}{\linewidth}\renewcommand{\arraystretch}{0.97}
\begin{longtable}{@{}lrrrrrr@{}}
\caption{\textbf{Patch correctness and task success (\%).} Patch reports $P=1$; Task reports $PU=1$; the third column reports $P=1,U=0$. Each evaluated entry uses the same $N$ tasks and single attempts as Table~\ref{tab:four-domain-results}. An em dash denotes an unevaluated condition.}\label{tab:score-decomposition}\\
\toprule
& \multicolumn{3}{c}{Code-only} & \multicolumn{3}{c}{Hybrid CUA} \\
\cmidrule(lr){2-4}\cmidrule(lr){5-7}
Model / system & Patch & Task & $P{=}1,U{=}0$ & Patch & Task & $P{=}1,U{=}0$ \\
\midrule
\endfirsthead
\multicolumn{7}{l}{\small Table~\ref{tab:score-decomposition} (continued)}\\
\toprule
& \multicolumn{3}{c}{Code-only} & \multicolumn{3}{c}{Hybrid CUA} \\
\cmidrule(lr){2-4}\cmidrule(lr){5-7}
Model / system & Patch & Task & $P{=}1,U{=}0$ & Patch & Task & $P{=}1,U{=}0$ \\
\midrule
\endhead
\bottomrule
\endfoot
\multicolumn{7}{l}{\textbf{Web} ($N=36$)}\\*
GPT-6-Astra & 27.8 & 27.8 & 0.0 & 66.7 & 66.7 & 0.0 \\
GPT-5.6-Sol & 19.4 & 19.4 & 0.0 & 58.3 & 58.3 & 0.0 \\
GPT-5.6-Luna & 11.1 & 11.1 & 0.0 & 13.9 & 5.6 & 8.3 \\
GPT-5.6-Terra & 11.1 & 11.1 & 0.0 & 13.9 & 13.9 & 0.0 \\
Claude-Opus-5 & 22.2 & 22.2 & 0.0 & 55.6 & 55.6 & 0.0 \\
Claude-Opus-4.8 & 16.7 & 16.7 & 0.0 & 55.6 & 27.8 & 27.8 \\
Claude-Sonnet-5 & 8.3 & 8.3 & 0.0 & 50.0 & 0.0 & 50.0 \\
Claude-Fable-5 & 22.2 & 19.4 & 2.8 & 55.6 & 47.2 & 8.3 \\
Grok-4.6 & 30.6 & 30.6 & 0.0 & 58.3 & 55.6 & 2.8 \\
Codex + GPT-5.6-Sol & \textemdash & \textemdash & \textemdash & 55.6 & 55.6 & 0.0 \\
Claude Code + Claude-Opus-5 & \textemdash & \textemdash & \textemdash & 55.6 & 55.6 & 0.0 \\
\midrule
\multicolumn{7}{l}{\textbf{Game} ($N=29$)}\\*
GPT-6-Astra & 17.2 & 17.2 & 0.0 & 37.9 & 37.9 & 0.0 \\
GPT-5.6-Sol & 10.3 & 10.3 & 0.0 & 10.3 & 10.3 & 0.0 \\
GPT-5.6-Luna & 0.0 & 0.0 & 0.0 & 6.9 & 6.9 & 0.0 \\
GPT-5.6-Terra & 3.4 & 3.4 & 0.0 & 10.3 & 6.9 & 3.4 \\
Claude-Opus-5 & 13.8 & 13.8 & 0.0 & 20.7 & 20.7 & 0.0 \\
Claude-Opus-4.8 & 3.4 & 3.4 & 0.0 & 20.7 & 20.7 & 0.0 \\
Claude-Sonnet-5 & 6.9 & 6.9 & 0.0 & 10.3 & 3.4 & 6.9 \\
Claude-Fable-5 & 13.8 & 13.8 & 0.0 & 20.7 & 17.2 & 3.4 \\
Grok-4.6 & 10.3 & 10.3 & 0.0 & 20.7 & 17.2 & 3.4 \\
Codex + GPT-5.6-Sol & \textemdash & \textemdash & \textemdash & 17.2 & 17.2 & 0.0 \\
Claude Code + Claude-Opus-5 & \textemdash & \textemdash & \textemdash & 20.7 & 20.7 & 0.0 \\
\midrule
\multicolumn{7}{l}{\textbf{DevOps} ($N=20$)}\\*
GPT-6-Astra & 0.0 & 0.0 & 0.0 & 80.0 & 80.0 & 0.0 \\
GPT-5.6-Sol & 0.0 & 0.0 & 0.0 & 55.0 & 55.0 & 0.0 \\
GPT-5.6-Luna & 0.0 & 0.0 & 0.0 & 15.0 & 15.0 & 0.0 \\
GPT-5.6-Terra & 0.0 & 0.0 & 0.0 & 25.0 & 25.0 & 0.0 \\
Claude-Opus-5 & 0.0 & 0.0 & 0.0 & 60.0 & 60.0 & 0.0 \\
Claude-Opus-4.8 & 0.0 & 0.0 & 0.0 & 55.0 & 55.0 & 0.0 \\
Claude-Sonnet-5 & 0.0 & 0.0 & 0.0 & 40.0 & 40.0 & 0.0 \\
Claude-Fable-5 & 5.0 & 5.0 & 0.0 & 55.0 & 55.0 & 0.0 \\
Grok-4.6 & 0.0 & 0.0 & 0.0 & 65.0 & 65.0 & 0.0 \\
Codex + GPT-5.6-Sol & \textemdash & \textemdash & \textemdash & 55.0 & 55.0 & 0.0 \\
Claude Code + Claude-Opus-5 & \textemdash & \textemdash & \textemdash & 65.0 & 65.0 & 0.0 \\
\midrule
\multicolumn{7}{l}{\textbf{Mobile} ($N=20$)}\\*
GPT-6-Astra & 0.0 & 0.0 & 0.0 & 55.0 & 55.0 & 0.0 \\
GPT-5.6-Sol & 0.0 & 0.0 & 0.0 & 45.0 & 45.0 & 0.0 \\
GPT-5.6-Luna & 0.0 & 0.0 & 0.0 & 35.0 & 35.0 & 0.0 \\
GPT-5.6-Terra & 0.0 & 0.0 & 0.0 & 35.0 & 35.0 & 0.0 \\
Claude-Opus-5 & \textemdash & \textemdash & \textemdash & \textemdash & \textemdash & \textemdash \\
Claude-Opus-4.8 & 0.0 & 0.0 & 0.0 & 40.0 & 40.0 & 0.0 \\
Claude-Sonnet-5 & 0.0 & 0.0 & 0.0 & 35.0 & 35.0 & 0.0 \\
Claude-Fable-5 & 5.0 & 5.0 & 0.0 & 40.0 & 40.0 & 0.0 \\
Grok-4.6 & 0.0 & 0.0 & 0.0 & 30.0 & 30.0 & 0.0 \\
Codex + GPT-5.6-Sol & \textemdash & \textemdash & \textemdash & 40.0 & 40.0 & 0.0 \\
Claude Code + Claude-Opus-5 & \textemdash & \textemdash & \textemdash & \textemdash & \textemdash & \textemdash \\
\end{longtable}
\endgroup

%% file: sections_preprint/11-appendix-related-work.tex
\section{Extended related work}\label{app:extended-related-work}
\subsection{Coding and software engineering benchmarks}
HumanEval measures whether programs synthesized from docstrings satisfy
functional tests~\citep{chen2021code}. LiveCodeBench uses newly collected
programming problems to evaluate generation, self-repair, execution, and
test-output prediction~\citep{jain2024livecodebench}. BigCodeBench broadens
function-level evaluation to complex instructions and compositions of
library calls~\citep{zhuo2025bigcodebench}. These benchmarks make functional
correctness measurable while varying the program understanding and tool
knowledge required to produce a solution.

Repository-level benchmarks introduce dependencies between changes across
an existing project. SWE-bench evaluates issue resolution in real
repositories~\citep{jimenez2024swebench}; SWE-bench Pro emphasizes longer
engineering tasks involving substantial, often multi-file
changes~\citep{deng2025swebenchpro}. SWE-Lancer includes both independent
engineering work, from bug fixes to feature implementations, and decisions
among implementation proposals. Its independent engineering tasks use
end-to-end tests~\citep{miserendino2025swelancer}. Terminal-Bench evaluates
multi-step work in command-line environments, pairing tasks with dedicated
environments, reference solutions, and verification
tests~\citep{merrill2026terminalbench}. These settings establish that SWE
evaluation extends beyond isolated code generation. CUA-SWE adds direct
interaction with the running software to this development process, so that
agents can connect implementation choices with observed application behavior.

\subsection{Computer-use benchmarks and visual grounding}
Mind2Web provides demonstrations of tasks on real websites and studies
generalization across websites and domains~\citep{deng2023mind2web}.
WebArena supplies reproducible, functional websites and evaluates task
completion~\citep{zhou2023webarena}. VisualWebArena adds tasks for which
visual information is needed to identify the appropriate objects and
actions~\citep{koh2024visualwebarena}. WebVoyager studies end-to-end
interaction with live websites and uses a multimodal model to judge task
completion~\citep{he2024webvoyager}. These evaluation settings differ in
whether they use recorded demonstrations or online execution and in how
they assess the final outcome.

OSWorld supports tasks across desktop applications with execution-based
evaluation~\citep{xie2024osworld}. Windows Agent Arena adapts this approach
to a real Windows environment~\citep{bonatti2024windowsarena}, while
AndroidWorld defines mobile tasks with programmatic initialization,
success checks, and teardown~\citep{rawles2024androidworld}.
ScreenSpot-Pro isolates a constituent capability: locating interface targets
in high-resolution screenshots of professional
applications~\citep{li2025screenspotpro}. Grounding accuracy, successful
navigation, and correct software changes measure different capabilities.
CUA-SWE requires agents to coordinate interface interaction with source or
configuration changes, and checks whether the resulting software satisfies
the task requirements.

\subsection{Visual and interactive software development}
SWE-bench Multimodal introduces images into issues and tests involving
user-facing JavaScript software~\citep{yang2024multimodal}. Design2Code
studies the reconstruction of webpages from reference screenshots, with
visual metrics and human evaluation~\citep{si2025design2code}.
WebGen-Bench evaluates websites generated from natural-language
requirements by using a navigation agent to exercise and check their
functionality~\citep{lu2025webgenbench}. GameDevBench studies development
in a game engine and investigates image- and video-based feedback during
implementation~\citep{chi2026gamedevbench}. Together, these works show how
visual information enters development as a specification, an observation,
or an evaluation signal. CUA-SWE emphasizes observations acquired while
using the software: the agent chooses what to inspect and how to interact,
then relates the observed behavior to its changes. Its task-specific tests
check explicit requirements across web development, game development,
mobile app development, and DevOps.

\subsection{Agent interfaces and post-training}
SWE-agent studies interfaces for repository navigation, editing, and
execution~\citep{yang2024sweagent}. OpenHands provides a platform for
software agents with code, command-line, and browser
tools~\citep{wang2024openhands}. On the computer-use side, Agent S2
combines generalist planning with specialist components for
interaction~\citep{agashe2025agents2}. UI-TARS learns screenshot-grounded
interaction policies~\citep{qin2025uitars}, and OpenCUA provides
computer-use demonstrations, data-processing infrastructure, and trained
agent models~\citep{wang2025opencua}. AndroidControl studies how training
data scale and instruction granularity affect mobile-agent
performance~\citep{li2024androidcontrol}. These approaches inform the
interfaces and supervision through which an agent learns to act.
CUA-SWE provides a task environment in which coding actions and computer
use belong to the same development episode, with executable feedback on
the software produced by that episode.

\subsection{Hybrid coding and computer-use environments}
Programming with Pixels provides a visual IDE for diverse SWE tasks and
evaluates both visual interaction and access to file-editing, shell, and
IDE APIs. Its programmatic verifiers can check compilation and test
outcomes independently of the agent's interaction
channel~\citep{aggarwal2025pixels}. This directly connects software
engineering with computer use and executable evaluation.
WeaveBench studies long-horizon work across graphical interfaces,
command-line tools, and code in multiple domains. Its trajectory-aware
agentic judge examines artifacts and interaction evidence against task
requirements~\citep{li2026weavebench}.

CUA-SWE organizes this combination around software engineering tasks with
deterministic tests of the resulting implementation and behavior. The
benchmark, evaluation pipeline, and post-training environment share these
executable requirements. This supports the study of how coding agents
choose when to use software, acquire visual feedback, revise code, and
verify the consequences of their changes.

%% file: sections_preprint/12-appendix-trajectories.tex
\clearpage
\section{Quantitative analysis of development trajectories}\label{app:trajectory-analysis}
We analyze interaction counts from 2,040 evaluation attempts and detailed
annotations of 214 trajectories, complementing the examples in
Appendix~\ref{app:repair-cases} with measurements of observation,
implementation, and verification behavior.

\subsection{Evaluation sample and annotation procedure}
\paragraph{Coverage and sampling.}
The records comprise 720 Web, 580 Game, 400 DevOps, and 340 Mobile attempts,
with the task-success outcomes of Table~\ref{tab:four-domain-results}.
Interaction counts use the full collection. Detailed review selects up to
two failures and one success per model--domain--condition combination,
using a fixed hash ranking of task identities within each outcome stratum.
The sample contains 59 Web, 59 Game, 52 DevOps, and 44 Mobile trajectories:
154 failures and 60 successes across 77 evaluated combinations. Review
examines instructions, interactions, authored patches, verifier findings,
and relevant screenshots. Annotation counts describe this stratified sample;
outcome associations weight each attempt by $N_h/n_h$, where $N_h$ and $n_h$
are its stratum's population and sample sizes.

\paragraph{Annotations and outcomes.}
Annotations distinguish relevant runtime information, relevant GUI behavior,
substantive edits, and GUI checks of an edited revision. Runtime information
includes execution of product code in tests or simulations; static source
inspection alone does not qualify. Fields with insufficient evidence remain unknown.
We report task success $S_i=P_iU_i$ and patch correctness $P_i$ separately.
Reviewers annotated the sample; a second reviewer annotated 32 overlapping
trajectories before seeing the primary labels, followed by adjudication.
Section~\ref{app:trajectory-reliability} reports original agreement and scope.

\subsection{Failure stages across models and domains}
\paragraph{A taxonomy of development behavior.}
We annotate five overlapping stages: \emph{observation-acquisition gaps} (O),
where relevant evidence has not been acquired; \emph{incorrect interpretation}
(Perc) of observed evidence; \emph{mistaken diagnosis} (D) of the underlying
mechanism; \emph{implementation errors or incomplete repairs} (I); and
\emph{verification-coverage gaps} (V), including unexamined interactions or
checks of stale revisions. Labels describe evidenced behavior, including
subsequently corrected errors, rather than mutually exclusive terminal causes.

\paragraph{Domain profiles.}
Among 84 sampled hybrid failures, implementation errors or incomplete repairs
occur in 61 trajectories (21 of 22 in Game), observation gaps in 59, and
verification gaps in 82. Verification gaps also occur in 36 of 41 sampled
hybrid successes, reflecting incomplete coverage beyond a repaired interaction.
Code-only O/V labels also reflect the condition's exclusion of GUI access.
Tables~\ref{tab:trajectory-stages} and~\ref{tab:trajectory-stages-model}
retain separate conditions and unknown counts.
\input{tables/trajectory-stages-domain}
\clearpage
\input{tables/trajectory-stages-model}

\subsection{Observation, editing, and GUI re-verification}
\paragraph{When agents observe relevant behavior.}
Of the 214 reviewed trajectories, 201 contain substantive edits and 13 contain
none. Table~\ref{tab:trajectory-observation} separates observation before the
first edit from observation at any point in the attempt. Among the 116 edited
hybrid trajectories, 83 establish relevant observation before editing, 32 do
not, and one has unknown ordering. All 116 acquire relevant runtime information
at some point, while six never observe the relevant GUI behavior. Among the
85 edited code-only trajectories, 47 obtain relevant runtime information
through non-GUI channels. These counts distinguish an early edit made before
observation from an attempt that never acquires relevant evidence.
\input{tables/trajectory-observation}

\paragraph{Checking the changed application.}
Post-edit GUI re-verification requires that an edited revision is active,
that the relevant behavior is exercised, and that the resulting GUI observation
is delivered before submission. Final-edit re-verification additionally
requires this sequence after the last substantive edit. The check may reveal
success or a remaining defect; its label records whether the agent checked
the behavior. Source-level tests and the evaluator's subsequent checks are
recorded separately from this agent-side GUI activity.

Table~\ref{tab:trajectory-reverification} reports weighted outcome rates among
the 116 edited hybrid trajectories. Observed final-edit re-verification is
associated with higher task success in each domain: 45.8\% versus 5.7\% on Web,
16.6\% versus 12.8\% on Game, 68.1\% versus 5.9\% on DevOps, and 83.7\% versus
10.6\% on Mobile. Patch-correctness rates show the same ordering; for Web,
they are 61.3\% versus 5.7\%. The table retains unknown check labels, including
13 of the 31 edited Game trajectories for the final-edit measure. These are
descriptive associations within the sampled attempts; task difficulty, model,
and the trajectory-validation component of task success can also contribute.
The small outcome strata support descriptive estimates rather than
significance tests or design-based confidence intervals.
\input{tables/trajectory-reverification}

\subsection{Interaction counts for successful and unsuccessful repairs}
\paragraph{Image deliveries and GUI actions.}
Table~\ref{tab:trajectory-workload} reports all 1,115 hybrid attempts in the
execution collection, separating task successes and failures within each
model and domain. We report the median and first and third quartiles, retaining
observed zeros. Image counts measure delivered screenshot content: successful
image-view results for Web, Game, and DevOps frontier models; pending-image
deliveries for Mobile frontier models; and returned image blocks for CLI agents.
Repeated transmission of a Mobile image is a separate transport operation and
is not added to its delivery count. \clearpage
\input{tables/trajectory-workload}

\paragraph{Action units.}
GUI actions use the domain's recorded
action unit: completed browser operations on Web, scaffold actions on Game,
non-observation protocol actions on DevOps, and completed non-screenshot
GUI actions on Mobile. These units are kept separate by model and interface.

\paragraph{Interaction quantity and repair outcome.}
More interaction does not uniformly accompany a successful repair. For
GPT-6-Astra on Web, successful attempts have a median of 7.5 image deliveries
and 6 GUI actions, compared with 19 deliveries and 17.5 actions for failures.
The complete distributions in Table~\ref{tab:trajectory-workload} situate such
patterns across all models, alongside the behavior-specific re-verification
measure above. Counts describe the recorded execution: 80 streams in the full
collection lack a recognized terminal marker (4 Web, 71 Game, 4 DevOps, and
1 Mobile), so their observed image counts are lower bounds. Quartiles use
linear interpolation at position $(n-1)p$ in the sorted observations.

\subsection{Patch correctness and paired task outcomes}
\paragraph{Separating repair quality from trajectory validation.}
We compare the same tasks under code-only and hybrid access in 925 frontier model
pairs across 35 model--domain groups. Three groups have lower hybrid task
success: GPT-5.6-Luna on Web and Claude-Sonnet-5 on Web and Game.
Table~\ref{tab:trajectory-paired} separates their passing patches from task
success. Across these 101 paired tasks, passing patches increase from 9 to 26,
while task successes change from 9 to 3; 23 hybrid passing patches do not
satisfy trajectory validation. The seven task-level losses comprise three
incorrect patches and four passing patches with invalid tool use; one hybrid
win offsets part of these losses. The passing-patch losses involve prohibited
HTTP or structured observations. The three incorrect-patch losses involve
two masonry-layout repairs and one conditional-physics repair, and also have
invalid tool use. This decomposition locates the lower task-success rates in
specific repair and tool-use outcomes rather than treating every loss as an
implementation failure. It uses the complete paired outcome records and
retained patch evidence for the loss cases.
\input{tables/trajectory-paired}

\Needspace{29\baselineskip}
\subsection{Annotation consistency}\label{app:trajectory-reliability}
\paragraph{Independent overlap and adjudication.}
Table~\ref{tab:trajectory-agreement} reports agreement on the 32 doubly annotated
trajectories before adjudication. Agreement is 100\% for acquiring relevant
runtime information and observing relevant GUI behavior, 90.6\% for observation
before the first edit, and 87.5\% for both GUI re-verification measures.
Stage annotations have lower agreement, especially interpretation and
diagnosis; many disagreements concern the distinction between not observed
and unknown. We therefore report stage labels as exploratory descriptive
annotations, alongside unknown counts and the original agreement statistics.
Adjudication examined 110 differing fields and changed 17 primary annotations.
The second reviewer had prior exposure to earlier population summaries
and selected calibration examples, but not to the new primary annotations.
The overlap measures reviewer consistency. Substantive-edit occurrence was not separately annotated by the
second reviewer, so agreement for that field is unavailable.
\input{tables/trajectory-agreement}

%% file: tables/trajectory-stages-domain.tex
\begin{table}[H]
\centering\fontsize{8.5}{10.1}\selectfont\setlength{\tabcolsep}{3pt}\renewcommand{\arraystretch}{1.06}
\caption{\textbf{Failure-stage annotations by domain and access condition.} Each cell gives supported/not observed/unknown counts among $n$ sampled task failures. O: observation; Perc: interpretation; D: diagnosis; I: implementation; V: verification coverage. Stages may overlap.}\label{tab:trajectory-stages}
\begin{tabular}{@{}llrccccc@{}}
\toprule
Domain & Condition & $n$ & O & Perc & D & I & V \\
\midrule
Web & Code-only & 18 & 18/0/0 & 0/18/0 & 2/14/2 & 5/0/13 & 18/0/0 \\
Web & Hybrid & 22 & 16/6/0 & 6/13/3 & 6/8/8 & 12/3/7 & 21/1/0 \\
Game & Code-only & 18 & 18/0/0 & 0/16/2 & 3/13/2 & 16/0/2 & 18/0/0 \\
Game & Hybrid & 22 & 15/7/0 & 6/12/4 & 7/9/6 & 21/0/1 & 22/0/0 \\
DevOps & Code-only & 18 & 18/0/0 & 1/16/1 & 14/0/4 & 16/0/2 & 18/0/0 \\
DevOps & Hybrid & 22 & 13/8/1 & 6/12/4 & 1/17/4 & 15/0/7 & 22/0/0 \\
Mobile & Code-only & 16 & 16/0/0 & 1/15/0 & 9/7/0 & 14/1/1 & 16/0/0 \\
Mobile & Hybrid & 18 & 15/1/2 & 2/13/3 & 0/16/2 & 13/3/2 & 17/1/0 \\
\bottomrule
\end{tabular}
\end{table}

%% file: tables/trajectory-stages-model.tex
\begingroup
\fontsize{8.3}{9.9}\selectfont\setlength{\tabcolsep}{3pt}\setlength{\LTcapwidth}{\linewidth}\renewcommand{\arraystretch}{1.06}
\begin{longtable}{@{}p{133pt}lrccccc@{}}
\caption{\textbf{Failure-stage counts for each model and domain.} Cells report supported/unknown counts; the remaining annotations are not observed. Counts describe the outcome-stratified review sample, with at most two failed attempts per model, domain, and condition. Stage definitions are shared with Table~\ref{tab:trajectory-stages}.}\label{tab:trajectory-stages-model}\\
\toprule
Model / system & Condition & $n$ & O & Perc & D & I & V \\
\midrule
\endfirsthead
\multicolumn{8}{l}{Table~\ref{tab:trajectory-stages-model} (continued)}\\
\toprule
Model / system & Condition & $n$ & O & Perc & D & I & V \\
\midrule
\endhead
\bottomrule
\endfoot
\midrule
\multicolumn{8}{@{}l}{\textbf{Web}}\\*
GPT-6-Astra & Code-only & 2 & 2/0 & 0/0 & 0/0 & 1/1 & 2/0 \\
GPT-6-Astra & Hybrid & 2 & 2/0 & 0/0 & 0/1 & 1/1 & 2/0 \\
GPT-5.6-Sol & Code-only & 2 & 2/0 & 0/0 & 0/0 & 0/2 & 2/0 \\
GPT-5.6-Sol & Hybrid & 2 & 2/0 & 0/0 & 1/0 & 0/2 & 2/0 \\
GPT-5.6-Luna & Code-only & 2 & 2/0 & 0/0 & 0/1 & 0/2 & 2/0 \\
GPT-5.6-Luna & Hybrid & 2 & 2/0 & 0/0 & 1/0 & 2/0 & 2/0 \\
GPT-5.6-Terra & Code-only & 2 & 2/0 & 0/0 & 1/0 & 1/1 & 2/0 \\
GPT-5.6-Terra & Hybrid & 2 & 2/0 & 1/0 & 1/1 & 2/0 & 2/0 \\
Claude-Opus-5 & Code-only & 2 & 2/0 & 0/0 & 0/0 & 1/1 & 2/0 \\
Claude-Opus-5 & Hybrid & 2 & 0/0 & 0/2 & 0/2 & 2/0 & 2/0 \\
Claude-Opus-4.8 & Code-only & 2 & 2/0 & 0/0 & 0/1 & 1/1 & 2/0 \\
Claude-Opus-4.8 & Hybrid & 2 & 1/0 & 1/0 & 1/0 & 0/0 & 2/0 \\
Claude-Sonnet-5 & Code-only & 2 & 2/0 & 0/0 & 1/0 & 1/1 & 2/0 \\
Claude-Sonnet-5 & Hybrid & 2 & 0/0 & 0/1 & 0/0 & 1/0 & 1/0 \\
Claude-Fable-5 & Code-only & 2 & 2/0 & 0/0 & 0/0 & 0/2 & 2/0 \\
Claude-Fable-5 & Hybrid & 2 & 2/0 & 0/0 & 1/0 & 1/1 & 2/0 \\
Grok-4.6 & Code-only & 2 & 2/0 & 0/0 & 0/0 & 0/2 & 2/0 \\
Grok-4.6 & Hybrid & 2 & 2/0 & 0/0 & 0/1 & 1/1 & 2/0 \\
Codex +\newline GPT-5.6-Sol & Hybrid & 2 & 2/0 & 2/0 & 0/2 & 0/2 & 2/0 \\
Claude Code +\newline Claude-Opus-5 & Hybrid & 2 & 1/0 & 2/0 & 1/1 & 2/0 & 2/0 \\
\midrule
\multicolumn{8}{@{}l}{\textbf{Game}}\\*
GPT-6-Astra & Code-only & 2 & 2/0 & 0/0 & 0/0 & 2/0 & 2/0 \\
GPT-6-Astra & Hybrid & 2 & 2/0 & 0/0 & 1/0 & 2/0 & 2/0 \\
GPT-5.6-Sol & Code-only & 2 & 2/0 & 0/0 & 1/0 & 1/1 & 2/0 \\
GPT-5.6-Sol & Hybrid & 2 & 2/0 & 0/0 & 0/0 & 2/0 & 2/0 \\
GPT-5.6-Luna & Code-only & 2 & 2/0 & 0/1 & 0/1 & 2/0 & 2/0 \\
GPT-5.6-Luna & Hybrid & 2 & 0/0 & 0/0 & 1/0 & 2/0 & 2/0 \\
GPT-5.6-Terra & Code-only & 2 & 2/0 & 0/0 & 0/0 & 2/0 & 2/0 \\
GPT-5.6-Terra & Hybrid & 2 & 2/0 & 0/0 & 1/0 & 2/0 & 2/0 \\
Claude-Opus-5 & Code-only & 2 & 2/0 & 0/0 & 1/0 & 1/1 & 2/0 \\
Claude-Opus-5 & Hybrid & 2 & 1/0 & 1/0 & 1/1 & 2/0 & 2/0 \\
Claude-Opus-4.8 & Code-only & 2 & 2/0 & 0/0 & 1/0 & 2/0 & 2/0 \\
Claude-Opus-4.8 & Hybrid & 2 & 2/0 & 2/0 & 0/2 & 2/0 & 2/0 \\
Claude-Sonnet-5 & Code-only & 2 & 2/0 & 0/0 & 0/0 & 2/0 & 2/0 \\
Claude-Sonnet-5 & Hybrid & 2 & 2/0 & 2/0 & 2/0 & 2/0 & 2/0 \\
Claude-Fable-5 & Code-only & 2 & 2/0 & 0/1 & 0/1 & 2/0 & 2/0 \\
Claude-Fable-5 & Hybrid & 2 & 0/0 & 0/0 & 0/0 & 2/0 & 2/0 \\
Grok-4.6 & Code-only & 2 & 2/0 & 0/0 & 0/0 & 2/0 & 2/0 \\
Grok-4.6 & Hybrid & 2 & 2/0 & 0/2 & 0/2 & 1/1 & 2/0 \\
Codex +\newline GPT-5.6-Sol & Hybrid & 2 & 0/0 & 1/0 & 0/0 & 2/0 & 2/0 \\
Claude Code +\newline Claude-Opus-5 & Hybrid & 2 & 2/0 & 0/2 & 1/1 & 2/0 & 2/0 \\
\midrule
\multicolumn{8}{@{}l}{\textbf{DevOps}}\\*
GPT-6-Astra & Code-only & 2 & 2/0 & 0/0 & 2/0 & 2/0 & 2/0 \\
GPT-6-Astra & Hybrid & 2 & 0/1 & 0/0 & 0/0 & 1/1 & 2/0 \\
GPT-5.6-Sol & Code-only & 2 & 2/0 & 0/0 & 2/0 & 2/0 & 2/0 \\
GPT-5.6-Sol & Hybrid & 2 & 2/0 & 0/1 & 0/1 & 1/1 & 2/0 \\
GPT-5.6-Luna & Code-only & 2 & 2/0 & 0/0 & 2/0 & 2/0 & 2/0 \\
GPT-5.6-Luna & Hybrid & 2 & 2/0 & 1/0 & 1/1 & 2/0 & 2/0 \\
GPT-5.6-Terra & Code-only & 2 & 2/0 & 0/0 & 2/0 & 2/0 & 2/0 \\
GPT-5.6-Terra & Hybrid & 2 & 2/0 & 0/1 & 0/1 & 0/2 & 2/0 \\
Claude-Opus-5 & Code-only & 2 & 2/0 & 0/0 & 1/1 & 2/0 & 2/0 \\
Claude-Opus-5 & Hybrid & 2 & 1/0 & 2/0 & 0/0 & 2/0 & 2/0 \\
Claude-Opus-4.8 & Code-only & 2 & 2/0 & 0/0 & 1/1 & 1/1 & 2/0 \\
Claude-Opus-4.8 & Hybrid & 2 & 2/0 & 1/0 & 0/0 & 1/1 & 2/0 \\
Claude-Sonnet-5 & Code-only & 2 & 2/0 & 1/1 & 1/1 & 2/0 & 2/0 \\
Claude-Sonnet-5 & Hybrid & 2 & 2/0 & 0/1 & 0/1 & 2/0 & 2/0 \\
Claude-Fable-5 & Code-only & 2 & 2/0 & 0/0 & 2/0 & 2/0 & 2/0 \\
Claude-Fable-5 & Hybrid & 2 & 1/0 & 0/1 & 0/0 & 1/1 & 2/0 \\
Grok-4.6 & Code-only & 2 & 2/0 & 0/0 & 1/1 & 1/1 & 2/0 \\
Grok-4.6 & Hybrid & 2 & 1/0 & 2/0 & 0/0 & 2/0 & 2/0 \\
Codex +\newline GPT-5.6-Sol & Hybrid & 2 & 0/0 & 0/0 & 0/0 & 2/0 & 2/0 \\
Claude Code +\newline Claude-Opus-5 & Hybrid & 2 & 0/0 & 0/0 & 0/0 & 1/1 & 2/0 \\
\midrule
\multicolumn{8}{@{}l}{\textbf{Mobile}}\\*
GPT-6-Astra & Code-only & 2 & 2/0 & 0/0 & 0/0 & 1/1 & 2/0 \\
GPT-6-Astra & Hybrid & 2 & 0/1 & 0/1 & 0/0 & 1/1 & 2/0 \\
GPT-5.6-Sol & Code-only & 2 & 2/0 & 0/0 & 2/0 & 2/0 & 2/0 \\
GPT-5.6-Sol & Hybrid & 2 & 2/0 & 0/0 & 0/0 & 1/0 & 2/0 \\
GPT-5.6-Luna & Code-only & 2 & 2/0 & 0/0 & 2/0 & 2/0 & 2/0 \\
GPT-5.6-Luna & Hybrid & 2 & 2/0 & 0/0 & 0/0 & 2/0 & 2/0 \\
GPT-5.6-Terra & Code-only & 2 & 2/0 & 0/0 & 1/0 & 2/0 & 2/0 \\
GPT-5.6-Terra & Hybrid & 2 & 2/0 & 0/0 & 0/0 & 1/1 & 2/0 \\
Claude-Opus-4.8 & Code-only & 2 & 2/0 & 0/0 & 2/0 & 2/0 & 2/0 \\
Claude-Opus-4.8 & Hybrid & 2 & 2/0 & 0/0 & 0/0 & 2/0 & 2/0 \\
Claude-Sonnet-5 & Code-only & 2 & 2/0 & 1/0 & 1/0 & 2/0 & 2/0 \\
Claude-Sonnet-5 & Hybrid & 2 & 2/0 & 0/0 & 0/0 & 1/0 & 1/0 \\
Claude-Fable-5 & Code-only & 2 & 2/0 & 0/0 & 0/0 & 2/0 & 2/0 \\
Claude-Fable-5 & Hybrid & 2 & 2/0 & 1/0 & 0/0 & 2/0 & 2/0 \\
Grok-4.6 & Code-only & 2 & 2/0 & 0/0 & 1/0 & 1/0 & 2/0 \\
Grok-4.6 & Hybrid & 2 & 2/0 & 1/0 & 0/0 & 1/0 & 2/0 \\
Codex +\newline GPT-5.6-Sol & Hybrid & 2 & 1/1 & 0/2 & 0/2 & 2/0 & 2/0 \\
\end{longtable}
\endgroup

%% file: tables/trajectory-observation.tex
\begin{table}[H]
\centering\fontsize{8.5}{10.1}\selectfont\setlength{\tabcolsep}{3pt}\renewcommand{\arraystretch}{1.06}
\caption{\textbf{Observation and editing in the reviewed trajectories.} $N/E$ gives all/edited attempts. Before-edit observation reports yes/no/unknown among the $E$ edited attempts; no-edit attempts are excluded. The last two columns count edited attempts with no relevant runtime information or no relevant GUI observation at any point.}\label{tab:trajectory-observation}
\begin{tabular}{@{}llrccc@{}}
\toprule
Domain & Condition & $N/E$ & \shortstack{Before edit\\Y/N/U} & \shortstack{No runtime\\information} & \shortstack{No GUI\\observation} \\
\midrule
Web & Code-only & 27/27 & 0/27/0 & 21 & 27 \\
Web & Hybrid & 32/32 & 19/13/0 & 0 & 4 \\
Game & Code-only & 26/24 & 0/24/0 & 13 & 24 \\
Game & Hybrid & 33/31 & 14/16/1 & 0 & 2 \\
DevOps & Code-only & 19/18 & 1/17/0 & 3 & 18 \\
DevOps & Hybrid & 33/32 & 32/0/0 & 0 & 0 \\
Mobile & Code-only & 17/16 & 1/15/0 & 1 & 16 \\
Mobile & Hybrid & 27/21 & 18/3/0 & 0 & 0 \\
\bottomrule
\end{tabular}
\end{table}

%% file: tables/trajectory-reverification.tex
\begin{table}[H]
\centering\fontsize{8.5}{10.1}\selectfont\setlength{\tabcolsep}{3pt}\renewcommand{\arraystretch}{1.06}
\caption{\textbf{GUI re-verification and repair outcomes among edited hybrid attempts.} Y/N/U are raw counts with an observed/absent/unknown GUI check. Task success $PU$ and patch correctness $P$ are inverse-inclusion-weighted percentages, shown separately for Y and N. Unknown check labels are retained in U and excluded from the Y--N comparison. All protocol statuses are included.}\label{tab:trajectory-reverification}
\begin{tabular}{@{}llcrrrr@{}}
\toprule
 &  &  & \multicolumn{2}{c}{Task success (\%)} & \multicolumn{2}{c}{Patch correctness (\%)} \\
Domain & GUI check & Y/N/U & Y & N & Y & N \\
\midrule
Web & Post-edit & 27/5/0 & 47.6 & 6.9 & 61.2 & 6.9 \\
Web & Final-edit & 24/7/1 & 45.8 & 5.7 & 61.3 & 5.7 \\
Game & Post-edit & 10/9/12 & 16.6 & 9.3 & 16.6 & 9.3 \\
Game & Final-edit & 7/11/13 & 16.6 & 12.8 & 16.6 & 12.8 \\
DevOps & Post-edit & 23/9/0 & 68.1 & 5.9 & 68.1 & 5.9 \\
DevOps & Final-edit & 23/9/0 & 68.1 & 5.9 & 68.1 & 5.9 \\
Mobile & Post-edit & 19/2/0 & 54.8 & 0.0 & 54.8 & 0.0 \\
Mobile & Final-edit & 10/11/0 & 83.7 & 10.6 & 83.7 & 10.6 \\
\bottomrule
\end{tabular}
\end{table}

%% file: tables/trajectory-workload.tex
\begingroup
\fontsize{7.7}{9.3}\selectfont\setlength{\tabcolsep}{3pt}\setlength{\LTcapwidth}{\linewidth}\renewcommand{\arraystretch}{1.06}
\begin{longtable}{@{}p{121pt}rcccc@{}}
\caption{\textbf{Observed image deliveries and GUI actions by repair outcome.} Every entry is a median [first quartile, third quartile] over hybrid task successes (S) or failures (F); $n$ gives S/F attempts. An em dash marks an empty success group. Counts are retained-log observations with interface-specific units, defined in the text.}\label{tab:trajectory-workload}\\
\toprule
 &  & \multicolumn{2}{c}{Image deliveries} & \multicolumn{2}{c}{GUI actions} \\
Model / system & $n$ (S/F) & S & F & S & F \\
\midrule
\endfirsthead
\multicolumn{6}{l}{Table~\ref{tab:trajectory-workload} (continued)}\\
\toprule
 &  & \multicolumn{2}{c}{Image deliveries} & \multicolumn{2}{c}{GUI actions} \\
Model / system & $n$ (S/F) & S & F & S & F \\
\midrule
\endhead
\bottomrule
\endfoot
\midrule
\multicolumn{6}{@{}l}{\textbf{Web}}\\*
GPT-6-Astra & 24/12 & 7.5 [6, 16.8] & 19 [14, 26.5] & 6 [4, 15.5] & 17.5 [12.2, 25.2] \\
GPT-5.6-Sol & 21/15 & 7 [5, 10] & 13 [9.5, 18] & 6 [4, 8] & 12 [8.5, 17] \\
GPT-5.6-Luna & 2/34 & 4 [3.5, 4.5] & 2.5 [1, 3] & 3.5 [3.2, 3.8] & 3.5 [1, 7] \\
GPT-5.6-Terra & 5/31 & 4 [2, 4] & 2 [1, 4.5] & 3 [2, 4] & 3 [1, 5] \\
Claude-Opus-5 & 20/16 & 6.5 [5, 11] & 20 [15.8, 24] & 9.5 [5, 13.2] & 36.5 [31.8, 50] \\
Claude-Opus-4.8 & 10/26 & 7.5 [5.2, 8.8] & 10.5 [4, 15.8] & 7 [5, 18.2] & 16 [6, 22.8] \\
Claude-Sonnet-5 & 0/36 & -- & 9.5 [3, 19] & -- & 11 [3, 23.5] \\
Claude-Fable-5 & 17/19 & 6 [5, 6] & 16 [12, 21] & 5 [5, 11] & 22 [17.5, 37] \\
Grok-4.6 & 20/16 & 12.5 [6.8, 15.2] & 18 [12, 21] & 11.5 [5.8, 14.2] & 17 [11.8, 22.2] \\
Codex +\newline GPT-5.6-Sol & 20/16 & 9 [6, 14.2] & 29 [19, 32.5] & 8 [5, 13.2] & 27.5 [18, 30.8] \\
Claude Code +\newline Claude-Opus-5 & 20/16 & 7 [5, 9.2] & 32 [25.2, 46] & 4.5 [3.8, 8] & 30 [21.2, 42] \\
\midrule
\multicolumn{6}{@{}l}{\textbf{Game}}\\*
GPT-6-Astra & 11/18 & 8 [6, 14] & 14.5 [9.5, 27.8] & 6 [5, 12] & 12 [8.5, 26.8] \\
GPT-5.6-Sol & 3/26 & 1 [1, 2] & 2.5 [1, 4.8] & 4 [3, 6] & 4 [2, 5] \\
GPT-5.6-Luna & 2/27 & 3.5 [2.8, 4.2] & 2 [1, 3.5] & 13 [10.5, 15.5] & 3 [0.5, 7] \\
GPT-5.6-Terra & 2/27 & 4.5 [3.8, 5.2] & 3 [1, 4] & 5 [3.5, 6.5] & 5 [1.5, 8.5] \\
Claude-Opus-5 & 6/23 & 19 [12.8, 25.2] & 10 [7, 12.5] & 17 [14, 22.2] & 10 [7, 23.5] \\
Claude-Opus-4.8 & 6/23 & 5.5 [4.2, 18.8] & 11 [6, 18.5] & 5 [4.2, 28.2] & 13 [5, 25] \\
Claude-Sonnet-5 & 1/28 & 12 [12, 12] & 14.5 [9.8, 19.2] & 8 [8, 8] & 19 [10.8, 44] \\
Claude-Fable-5 & 5/24 & 12 [11, 14] & 9 [6.8, 14.5] & 15 [10, 27] & 14 [6.8, 36] \\
Grok-4.6 & 5/24 & 12 [10, 25] & 8 [2.8, 11.2] & 12 [10, 25] & 7 [1, 10] \\
Codex +\newline GPT-5.6-Sol & 5/24 & 11 [9, 13] & 11 [6, 25.8] & 10 [8, 11] & 10 [5, 23.8] \\
Claude Code +\newline Claude-Opus-5 & 6/23 & 24 [18.8, 33.8] & 24 [11.5, 34] & 21 [14, 28.8] & 18 [10, 28] \\
\midrule
\multicolumn{6}{@{}l}{\textbf{DevOps}}\\*
GPT-6-Astra & 16/4 & 18 [13.8, 27] & 31 [18, 44.2] & 13.5 [9.8, 21] & 24.5 [13.5, 35.8] \\
GPT-5.6-Sol & 11/9 & 18 [16, 18.5] & 15 [11, 19] & 17 [15, 17.5] & 14 [10, 18] \\
GPT-5.6-Luna & 3/17 & 3 [3, 4] & 2 [1, 3] & 9 [8, 11] & 7 [3, 11] \\
GPT-5.6-Terra & 5/15 & 6 [4, 7] & 5 [2.5, 5.5] & 6 [5, 7] & 4 [3, 8.5] \\
Claude-Opus-5 & 12/8 & 20.5 [13.8, 26] & 30 [26, 36.2] & 25 [13.8, 35.5] & 33.5 [26.5, 64] \\
Claude-Opus-4.8 & 11/9 & 10 [9.5, 14.5] & 12 [9, 27] & 12 [10.5, 16] & 13 [9, 27] \\
Claude-Sonnet-5 & 8/12 & 19.5 [15, 24.8] & 28 [24.5, 33] & 18.5 [14, 23.5] & 28 [20, 34.8] \\
Claude-Fable-5 & 11/9 & 13 [10.5, 18.5] & 30 [11, 35] & 16 [11.5, 29] & 50 [12, 60] \\
Grok-4.6 & 13/7 & 24 [18, 32] & 41 [32, 44.5] & 26 [18, 31] & 34 [31, 40] \\
Codex +\newline GPT-5.6-Sol & 11/9 & 33 [20, 38.5] & 24 [17, 29] & 32 [19, 37.5] & 23 [16, 28] \\
Claude Code +\newline Claude-Opus-5 & 13/7 & 21 [14, 38] & 35 [23, 43.5] & 17 [9, 26] & 32 [20, 33] \\
\midrule
\multicolumn{6}{@{}l}{\textbf{Mobile}}\\*
GPT-6-Astra & 11/9 & 41 [35.5, 44.5] & 48 [45, 50] & 36 [33.5, 38] & 46 [44, 47] \\
GPT-5.6-Sol & 9/11 & 37 [33, 39] & 54 [52, 55] & 36 [32, 38] & 54 [53, 55] \\
GPT-5.6-Luna & 7/13 & 21 [19, 21.5] & 35 [25, 42] & 19 [17.5, 20] & 33 [23, 39] \\
GPT-5.6-Terra & 7/13 & 19 [18.5, 20] & 27 [23, 32] & 17 [16.5, 18] & 25 [21, 30] \\
Claude-Opus-4.8 & 8/12 & 38.5 [31.5, 45.5] & 58 [48.8, 65.5] & 37.5 [29.8, 44.8] & 58 [48, 65.2] \\
Claude-Sonnet-5 & 7/13 & 46 [39.5, 49.5] & 54 [52, 55] & 45 [39, 49] & 52 [52, 55] \\
Claude-Fable-5 & 8/12 & 47 [42.5, 53.5] & 43.5 [40.2, 52.5] & 46 [41.5, 53.5] & 41 [28.8, 52.5] \\
Grok-4.6 & 6/14 & 49 [48.2, 52.8] & 39 [37, 51.5] & 48 [46.2, 52] & 39 [36, 51.2] \\
Codex +\newline GPT-5.6-Sol & 8/12 & 53.5 [47.5, 66.5] & 54.5 [53.8, 64.8] & 51.5 [46.5, 61.2] & 53.5 [52.8, 63.8] \\
\end{longtable}
\endgroup

%% file: tables/trajectory-paired.tex
\begin{table}[H]
\centering\fontsize{8.1}{9.7}\selectfont\setlength{\tabcolsep}{3pt}\renewcommand{\arraystretch}{1.06}
\caption{\textbf{Matched tasks for model--domain pairs with lower hybrid task success.} C/H denotes code-only/hybrid. Wins and losses compare task success on the same task. The final column counts all hybrid passing patches with $U=0$, including tasks where code-only also fails.}\label{tab:trajectory-paired}
\begin{tabular}{@{}lp{100pt}rcccc@{}}
\toprule
Domain & Model & $N$ & \shortstack{Task\\C/H} & \shortstack{Patch\\C/H} & \shortstack{Hybrid\\wins/losses} & $P{=}1,U{=}0$ \\
\midrule
Web & GPT-5.6-Luna & 36 & 4 / 2 & 4 / 5 & 0 / 2 & 3 \\
Web & Claude-Sonnet-5 & 36 & 3 / 0 & 3 / 18 & 0 / 3 & 18 \\
Game & Claude-Sonnet-5 & 29 & 2 / 1 & 2 / 3 & 1 / 2 & 2 \\
\bottomrule
\end{tabular}
\end{table}

%% file: tables/trajectory-agreement.tex
\begin{table}[H]
\centering\fontsize{8.5}{10.1}\selectfont\setlength{\tabcolsep}{3pt}\renewcommand{\arraystretch}{1.06}
\caption{\textbf{Reviewer agreement before adjudication.} The same 32 trajectories receive both annotations. Unknown is an explicit category in nominal Cohen\textquotesingle s $\kappa$.}\label{tab:trajectory-agreement}
\begin{tabular}{@{}lrrr@{}}
\toprule
Annotation & Agreement & (\%) & $\kappa$ \\
\midrule
Editing before relevant observation & 31/32 & 96.9 & 0.939 \\
Relevant GUI behavior observed at any point & 32/32 & 100.0 & 1.000 \\
Final-edit GUI re-verification & 28/32 & 87.5 & 0.771 \\
Post-edit GUI re-verification & 28/32 & 87.5 & 0.775 \\
Relevant observation before first edit & 29/32 & 90.6 & 0.819 \\
Relevant runtime information observed & 32/32 & 100.0 & 1.000 \\
O: observation-acquisition gap & 16/32 & 50.0 & 0.238 \\
Perc: incorrect interpretation & 1/32 & 3.1 & 0.028 \\
D: mistaken diagnosis & 2/32 & 6.2 & 0.000 \\
I: implementation error or incomplete repair & 15/32 & 46.9 & 0.240 \\
V: verification-coverage gap & 28/32 & 87.5 & 0.301 \\
\bottomrule
\end{tabular}
\end{table}